\PassOptionsToPackage{table,xcdraw}{xcolor}
\documentclass[format=acmsmall, authorversion=true, nonacm=true]{acmart}

\AtBeginDocument{%
  \providecommand\BibTeX{{%
    \normalfont B\kern-0.5em{\scshape i\kern-0.25em b}\kern-0.8em\TeX}}}

\setcopyright{acmcopyright}

\acmJournal{CSUR}

\usepackage{amsfonts}
\usepackage{amsmath}
  
\usepackage{amssymb}
\usepackage{graphicx}  
\usepackage{float}  
\usepackage{subfigure}
\usepackage{bm}
\usepackage{courier}
\newcommand{\tabincell}[2]{\begin{tabular}{@{}#1@{}}#2\end{tabular}}
\usepackage{multirow,rotating}
\usepackage{booktabs}
\usepackage[table]{xcolor}
\usepackage{color}

\begin{document}

\title{A Survey of Orthogonal Moments for Image Representation: Theory, Implementation, and Evaluation} 

\titlenote{ACM Computing Surveys, Volume 55, Issue 1, January 2023, Article No. 1, pp 1–35, https://doi.org/10.1145/3479428.}
\author{Shuren Qi}
\email{shurenqi@nuaa.edu.cn}

\author{Yushu Zhang}
\email{yushu@nuaa.edu.cn}

\author{Chao Wang}
\email{c.wang@nuaa.edu.cn}

\affiliation{%
  \institution{Nanjing University of Aeronautics and Astronautics}
  \country{China}
}

\author{Jiantao Zhou}
\email{jtzhou@umac.mo}
\affiliation{%
	\institution{University of Macau}
	\country{China}
}

\author{Xiaochun Cao}
\email{caoxch5@sysu.edu.cn}
\affiliation{%
	\institution{Sun Yat-sen University}
	\country{China}
}

\thanks{This work was supported in part by the National Key R\&D Program of China under Grant 2018AAA0100600, in part by the National Natural Science Foundation of China under Grants 62072237 and 61971476, in part by the Research Fund of Guangxi Key Lab of Multi-source Information Mining \& Security under Grant MIMS20-02, in part by the Guangxi Key Laboratory of Trusted Software under Grant KX202027, in part by the Basic Research Program of Jiangsu Province under Grant BK20201290, and in part by the Macau Science and Technology Development Fund under Grant 077/2018/A2.}

\authorsaddresses{%
Authors’ addresses: S. Qi, Y. Zhang (corresponding author), and C. Wang are with the College of Computer Science and Technology, Nanjing University of Aeronautics and Astronautics, Nanjing, China, and also with the Collaborative Innovation Center of Novel Software Technology and Industrialization, Nanjing, China (e-mail: shurenqi@nuaa.edu.cn, yushu@nuaa.edu.cn, and c.wang@nuaa.edu.cn).\\
J. Zhou is with the State Key Laboratory of Internet of Things for Smart City, University of Macau, Macau, China, and also with the Department of Computer and Information Science, University of Macau, Macau, China (e-mail: jtzhou@umac.mo).\\
X. Cao is with the School of Cyber Science and Technology, Sun Yat-sen University, Shenzhen, China  (e-mail: caoxch5@sysu.edu.cn).
}

\renewcommand{\shortauthors}{S. Qi et al.}

\setcopyright{none}

\settopmatter{printacmref=false}


\begin{abstract}
  Image representation is an important topic in computer vision and pattern recognition. It plays a fundamental role in a range of applications towards understanding visual contents. Moment-based image representation has been reported to be effective in satisfying the core conditions of semantic description due to its beneficial mathematical properties, especially geometric invariance and independence. This paper presents a comprehensive survey of the orthogonal moments for image representation, covering recent advances in fast/accurate calculation, robustness/invariance optimization, definition extension, and application. We also create a software package for a variety of widely-used orthogonal moments and evaluate such methods in a same base. The presented theory analysis, software implementation, and evaluation results can support the community, particularly in developing novel techniques and promoting real-world applications.
\end{abstract}



\maketitle

\section{Introduction}
In the mid-twentieth century, American mathematician Claude Elwood Shannon published his paper \emph{A mathematical theory of communication} [1], which marked the creation of information theory. As a landmark contribution, information theory is the theoretical foundation of information storage, processing, and transmission in modern computer systems. On the other hand, it also dictates that the raw signal (i.e., digital data) of multimedia (e.g., image) is not semantic in nature. Therefore, one of the main requirements in many computer vision and pattern recognition applications is to have a “meaningful representation” in which semantic characteristics of digital image are readily apparent, as shown in Figure 1. This process is often termed as \emph{image representation}. For example, for recognition, the representation should highlight the most salient semantics; for denoising, it should efficiently distinguish between signal (semantically relevant) and noise (semantically irrelevant); and for compression, it should capture the most semantic information using the least coefficients.

\begin{figure}[!t]
	\centering
	\includegraphics[scale=0.4]{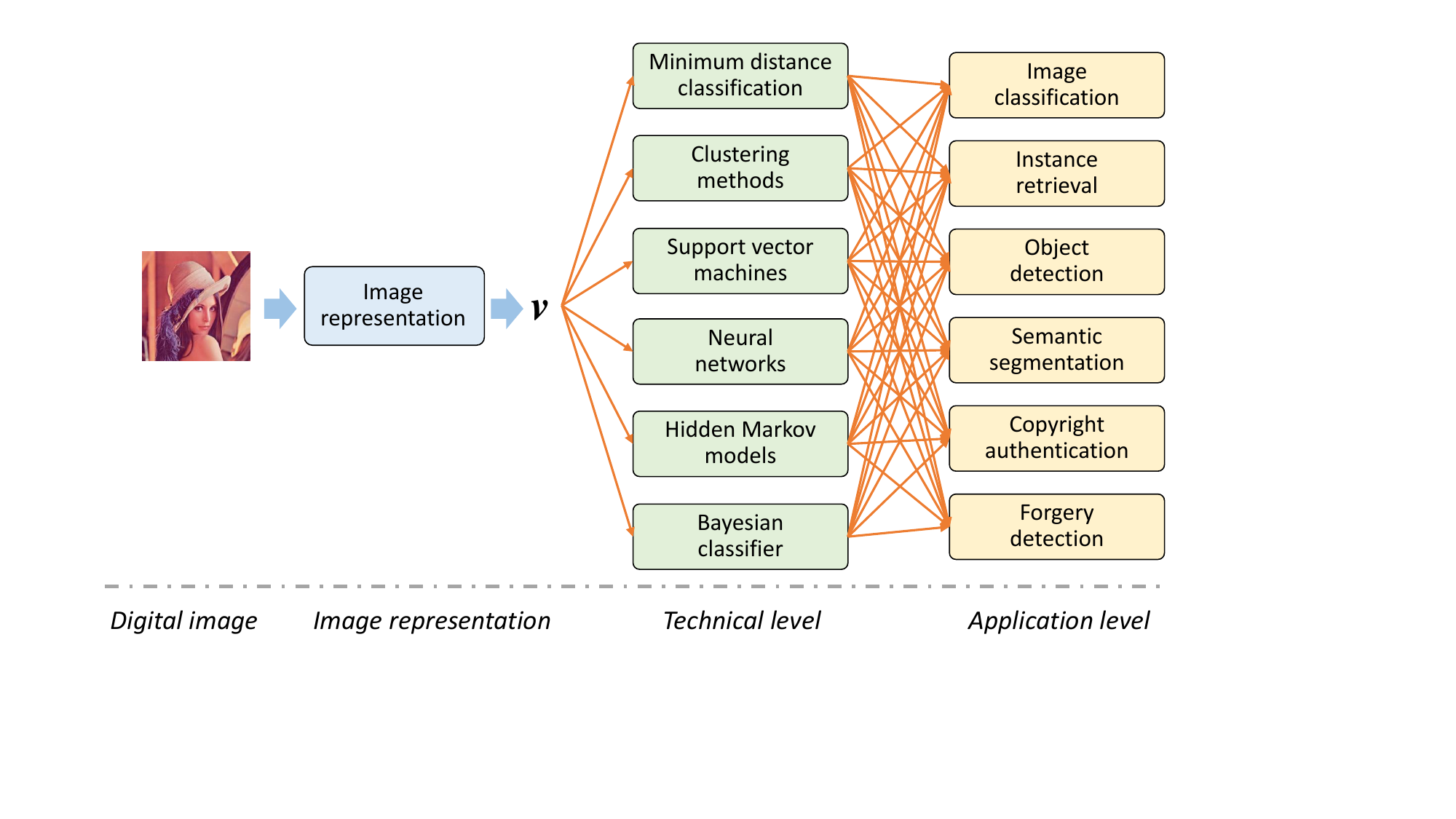}
	\centering
	\caption{The fundamental role of image representation in computer vision and pattern recognition applications. Typical visual system is generally formed on the basis of a "meaningful representation" of digital image (image representation), then followed by knowledge extraction techniques (technical level), and finally achieved the high-level visual understanding (application level).}
\end{figure}

Mathematically, a basic idea of image representation is to project the original image function onto a space formed by a set of specially designed basis functions and obtain the corresponding feature vector. For many years, dictionary (i.e., the set of basis functions) design has been pursued by many researchers in roughly two different paths: \emph{hand-crafted} and \emph{deep learning} [2].

Recently, deep learning techniques, represented by Convolutional Neural Networks (CNN), have led to very good performance on a variety problems of computer vision and pattern recognition. Deep learning based image representations formed by the composition of multiple nonlinear transformations, mapping raw image data directly into abstract semantic representations without manual intervention (i.e., end-to-end paradigm). For such representation learning methods, the dictionary can be considered as a composite function and is trained/learned by back-propagating error. Deep learning based image representations offer great flexibility and the ability to adapt to specific signal data. Due to the data-driven nature, this line of approaches is strongly influenced by the latest advances in optimization algorithms, computing equipment, and training data. As a result, the deep learning approaches exhibit limitations in the following three aspects [3]: 1) The quality of the representation depends heavily on the completeness of the training data, i.e., a large and diverse training set is required. 2) The time/space cost of these approaches is often very high, which prevents them from being used in time-critical applications. 3) The robustness to geometric transformation is limited, requiring the data augmentation to enhance the geometric invariance, but at the cost of time/space complexity. In contrast, hand-crafted image representations are still competitive in the above three aspects. It is also worth mentioning that the successful experiences behind hand-crafted features are instructive for the design of deep learning methods such as Principal Component Analysis Network (PCANet) [4] and Spatial Pyramid Pooling (SPP) [5].

In the pre-CNN era, hand-crafted representations and feature engineering had made important contributions to the development of computer vision and pattern recognition. At current stage, hand-crafted features still cannot be completely replaced, considering that some limitations of deep learning are just the characteristics of hand-crafted representations [6]. The existing hand-crafted image representation methods can be roughly divided into four categories [7]:

\begin{itemize}
	\item \emph{Frequency transform} – such as Fourier Transform, Walsh Hadamard Transform, and Wavelet Transform;
	\item \emph{Texture} – such as Scale Invariant Feature Transform (SIFT), Gradient Location and Orientation Histogram (GLOH), and Local Binary Patterns (LBP);
	\item \emph{Dimensionality reduction} – such as Principal Component Analysis (PCA), Singular Value Decomposition (SVD), and Locally Linear Embedding (LLE);
	\item \emph{Moments and moment invariants} – such as Zernike Moments (ZM), Legendre Moments, and Polar Harmonic Transforms (PHT).
\end{itemize}

Starting from the semantic nature of the representation task, it is clear that image representation should satisfy the following basic conditions [8]:

\begin{itemize}
	\item \emph{Discriminability} – the representation reflects inter-class variations, i.e., two objects from two different classes have different features;
	\item \emph{Robustness} – the representation is not influenced by intra-class variations, i.e., two objects from one class have the same features.
\end{itemize}

Hand-crafted representations based on frequency transform, texture, and dimensionality reduction have been widely used in the real-world applications. However, due to the inherent \emph{semantic gap} between low-level descriptors and high-level visual concepts, these methods have flaws in one or both of robustness and discriminability. One example is that SIFT features exhibit \emph{synonymy} and \emph{polysemy} [9], caused by poor discriminability and robustness. In contrast, moments and moment invariants perform better in overcoming the semantic gap due to their beneficial mathematical properties:

\begin{itemize}
	\item \emph{Independence} – the orthogonality of basis functions ensures no information redundancy in moment set, which in turn leads to better discriminability in image representation;
	\item \emph{Geometric invariance} – the invariants w.r.t. geometric transformation, e.g., rotation, scaling, translation, and flipping, can be derived from the moment set, meaning better robustness in image representation.
\end{itemize}

Moments and moment invariants were introduced to the pattern recognition communities in 1962 by Hu [10]. Since then, after almost 60 years of research, numerous moment based techniques have been developed for image representation with varying degrees of success. In 1998, Mukundan et al. [11] surveyed the main publications proposed until then and summarized the theoretical aspects of several classical moment functions. In 2006, Pawlak [12] gave a comprehensive survey on the reconstruction and calculation aspects of the moments with great emphasis to the accuracy/error analysis. In 2007, Shu et al. [13–15] provided a brief literature review for the mathematical definitions, invariants, and fast/accurate calculations of the classical moments, respectively. In 2009, Flusser et al. [8] presented a unique overview of moment based pattern recognition methods with significant contribution to the theory of moment invariants. The substantial expansion [16] of this book includes more detailed analysis of the 3D object invariant representation. In 2011, Hoang [17] reviewed unit disk-based orthogonal moments in his doctoral dissertation, covering theoretical analysis, mathematical properties, and specific implementation. For most of the above reviews, state-of-the-art methods in the past 10 years are not covered. In 2014, Papakostas et al. [18] gave a global overview of the milestones in the 50 years research and highlighted all recent rising topics in this field. However, the theoretical basis for these latest research directions is rarely introduced. More recently, in 2019, Kaur et al. [19] provided a comparative review for many classical and new moments. Although this paper covers almost all the main literatures, it still lacks the overall analysis of the current research progress in various directions. A common weakness of all the works is that there are almost no available software packages, restricting the further development of the community.

The significant contribution of this paper is to give a systematic investigation for orthogonal moments based image representation along with an open-source implementation, which we believe would be a useful complement to [17–19]. For completeness, this paper starts with some basic theories and classical methods in the area of orthogonal moments. Different from the mentioned reviews, we pay special attention to the motivation and successful experiences behind these traditional works. Furthermore, we organize a discussions for the recent advances of orthogonal moments on different research directions, including fast/accurate calculation, robustness/invariance optimization, definition extension, and application. Such overall theoretical analysis of the state-of-the-art research progress is mostly ignored in previous studies. In addition, we show the performance evaluation of widely-used orthogonal moments in terms of moment calculation, image reconstruction, and pattern recognition. To embrace the concept of reproducible research, the corresponding software package is available online. In the end, several promising directions for future research are given along with some initial discussions/suggestions.

The rest of this paper is organized as follows. In the following Section 2, we first give the basic idea of image moments and categorize existing methods into different categories. In Section 3, we further review in detail the unit disk-based orthogonal moments that are most relevant to image representation. Then, the recent advances of orthogonal moments within each research direction are reviewed and analyzed in Section 4. Furthermore, Section 5 reports the comparative results of state-of-the-art orthogonal moments along with an open-source implementation. Section 6 gives a conclusion and highlights some promising directions in this field.

\section{Overview}
Mathematically, the image moment is generally defined as the inner product $ \left< f,{V_{nm}} \right> $  of the image function $f$ and the basis function ${V_{nm}}$ of $(n + m)$ order on the domain $D$ [8]:
\begin{equation}
	\left<f,{V_{nm}} \right> = \iint\limits_D {V_{nm}^*(x,y)f(x,y)dxdy},
\end{equation}
where the asterisk $*$ denotes the complex conjugate. The direct geometric interpretation of image moment set $ \left< f,{V_{nm}} \right> $ is that it is the projection of $f$ onto a subspace formed by a set of basis functions $\{V_{nm}:(n,m)\in \vmathbb{Z}^2\} $ [17]. Because there are an infinite number of basis function sets, it is often necessary to manually design a set of special basis functions ${V_{nm}}$ with beneficial properties in $\left< f,{V_{nm}} \right> $  that meet the semantic requirements. According to the mathematical properties of basis functions, the family of image moments can be divided into different categories, as shown in Figure 2.

\begin{figure}[!t]
	\centering
	\includegraphics[scale=0.32]{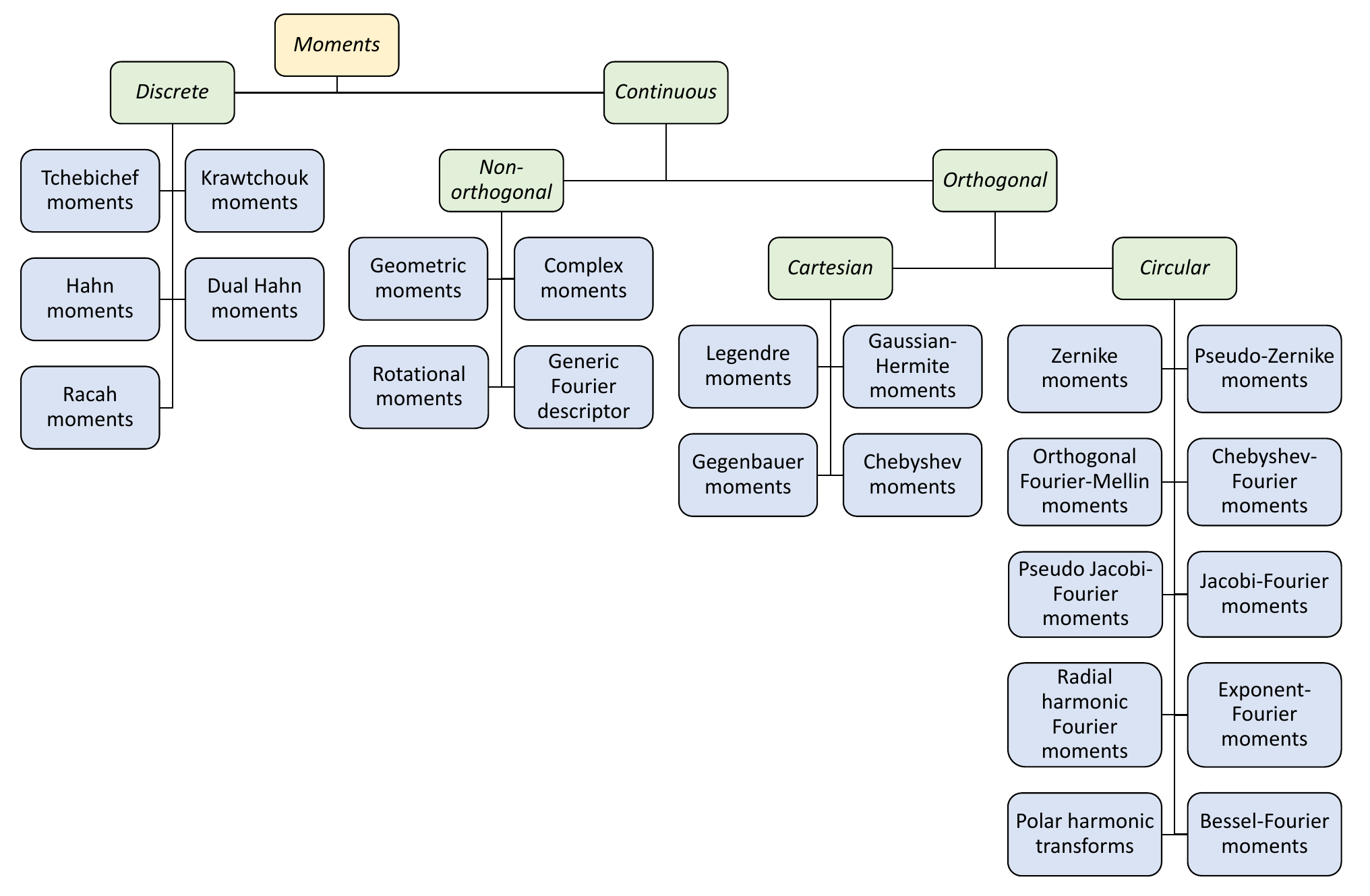}
	\centering
	\caption{A classification of image moments.}
\end{figure}

Firstly, depending on whether the basis functions satisfy \emph{orthogonality}, the image moments can be classified into orthogonal moments and non-orthogonal moments. The orthogonality means any two different basis functions ${V_{nm}}$ and ${V_{n'm'}}$ from the basis function set are uncorrelated or they are “perpendicular” in geometric term, leading to no redundancy in the moment set. Mathematically,  ${V_{nm}}$ and ${V_{n'm'}}$ are orthogonal when the following condition is satisfied
\begin{equation}
	\begin{split}
		\left<{V_{nm}},{V_{n'm'}}\right> = \iint\limits_D {{V_{nm}}(x,y)V_{n'm'}^*(x,y)dxdy}
		= {\delta _{nn'}}{\delta _{mm'}},
	\end{split}
\end{equation}
where ${\delta _{ij}}$  is the Kronecker delta function defined as
\begin{equation}
	{\delta _{ij}} = \left\{ {\begin{array}{*{20}{c}}
			0&{i \ne j}\\
			1&{i = j}
	\end{array}} \right..
\end{equation}

Some of the most popular non-orthogonal moments are geometric moments [8], rotational moments [20], complex moments [21], and generic Fourier descriptor [22]. Due to the non-orthogonality of the basis functions, high information redundancy exists in such moments. This further leads to difficulties in image reconstruction and poor discriminability/robustness in image representation. Therefore, it is a natural requirement to satisfy orthogonality when designing basis functions.

Secondly, according to whether the basis function is \emph{continuous}, the image moments can be divided into continuous moments and discrete moments. In the case of two-dimensional (2D) images, the basis functions for continuous and discrete moments are generally defined in the 2D real-valued space and the 2D digital image space, i.e., the domains $D \in \vmathbb{R}^2$ and $D \in \vmathbb{Z}^2$ , respectively. When it is necessary to calculate the continuous moments of a digital image, a suitable discretization approximation of the continuous integral is often introduced with corresponding computational errors. In Section 4.1, we will further describe the causes and solutions to such errors. On the contrary, discrete moments, such as Tchebichef moments [23], Krawtchouk moments [24], Hahn moments [25], dual Hahn moments [26], and Racah moments [27], do not involve any approximation errors. Thus they are more suitable for the high-precision image processing tasks, e.g., image reconstruction, compression, and denoising.

Finally, depending on the \emph{coordinate system} that defines the basis functions, image moments can be grouped into Cartesian moments and circular moments. In the case of continuous moments, the basis functions of Cartesian moments are defined in $D = \{ (x,y):x \in ( - \infty , + \infty ),y \in ( - \infty , + \infty )\} $ or $D = \{ (x,y):x \in [ - 1,1],y \in [ - 1,1]\} $, while the domain for the circular moments is $D = \{ (r,\theta ):r \in [0, + \infty ),\theta  \in [0,2\pi )\} $ or $D = \{ (r,\theta ):r \in [0,1],\theta  \in [0,2\pi )\} $ (i.e., the unit disk). According to the proof of Bhatia et al. [28], the basis function ${V_{nm}}$ will be \emph{invariant in form} w.r.t. rotations of axes about the origin $(x,y) = (0,0)$ if and only if, when expressed in polar coordinates $(r,\theta )$, it is of the form:
\begin{equation}
	{V_{nm}}(r\cos \theta ,r\sin \theta ) \equiv {V_{nm}}(r,\theta ) = {R_n}(r){A_m}(\theta ),
\end{equation}
with \emph{angular basis function} ${A_m}(\theta ) = \exp (\bm{j}m\theta )$ ($\bm{j} = \sqrt { - 1} $) and \emph{radial basis function} ${R_n}(r)$ could be of any form [17]. Let ${f^{\rm{rot}}}$ be the rotated version of the original image $f$. When ${V_{nm}}$  conforms to the form of Equation (4), there must be a function ${\mathcal{I}}$  such that
\begin{equation}
	{\mathcal{I}}(\{ \left< f,{V_{nm}} \right> \} ) \equiv {\mathcal{I}}(\{  \left< {f^{\rm{rot}}},{V_{nm}} \right> \} ),
\end{equation}
i.e., satisfying the \emph{rotation invariance}. Therefore, the Cartesian moments, such as Legendre moments [29], Gaussian-Hermite moments [30], Gegenbauer moments [31], Chebyshev moments [32], and the discrete moments listed above, have difficulties in achieving the rotation invariance. As for the calculation of circular moments, an appropriate coordinate transformation is often introduced since digital images are generally defined in a Cartesian coordinate system with corresponding computational errors. In Section 4.1, we will further describe the causes and solutions to such errors.

From the above theoretical analysis, it is clear that the circular orthogonal moments are generally better than other kind of moments as far as image representation tasks are concerned. Therefore, great scientific interest has been given to the circular orthogonal moments, mainly the unit disk-based orthogonal moments. As the most relevant works, existing unit disk-based orthogonal moments will be reviewed in the next section.

\section{Classical Orthogonal Moments}
It can be checked from Equation (4) that the basis functions of unit disk-based orthogonal moments are separable, i.e., decomposed into the product of the radial basis functions and the angular basis functions. Therefore, Equation (2) can be rewritten as
\begin{equation}
	\begin{split}
		\left<  {V_{nm}},{V_{n'm'}}\right> &= \int\limits_0^{2\pi } {\int\limits_0^1 {{R_n}(r){A_m}(\theta )R_{n'}^*(r)A_{m'}^*(\theta )rdrd\theta } } 
		= \int\limits_0^{2\pi } {{A_m}(\theta )A_{m'}^*(\theta )d\theta } \int\limits_0^1 {{R_n}(r)R_{n'}^*(r)rdr} \\
		&= 2\pi {\delta _{mm'}}\int\limits_0^1 {{R_n}(r)R_{n'}^*(r)rdr} .
	\end{split}
\end{equation}

Since $ \left<  {V_{nm}},{V_{n'm'}} \right>  = {\delta _{nn'}}{\delta _{mm'}}$ , the radial basis function ${R_n}(r)$  should satisfy the following weighted orthogonality condition:
\begin{equation}
	\int\limits_0^1 {{R_n}(r)R_{n'}^*(r)rdr}  = \frac{1}{{2\pi }}{\delta _{nn'}}.
\end{equation}

Equation (7) is a general requirement that must be considered when designing unit disk-based orthogonal moments, which ensures that the designed basis function set has the beneficial orthogonal properties. The angular basis functions ${A_m}(\theta )$  have a fixed form $\exp (\bm{j}m\theta )$  due to the proof of Bhatia et al. [28], which means that the difference in existing methods is only in the definition of the radial basis functions. In this regard, there are mainly three types of orthogonal functions used as the definition, including \emph{Jacobi polynomials}, \emph{harmonic functions}, and \emph{eigenfunctions}. Next, we will briefly introduce the specific methods in these three groups and give their radial basis function definitions in a unified form, i.e., normalized version as Equation (7).

\subsection{Jacobi Polynomials}

In this group, the famous methods mainly include ZM [29], Pseudo-Zernike Moments (PZM) [20], Orthogonal Fourier-Mellin Moments (OFMM) [33], Chebyshev-Fourier Moments (CHFM) [34], Pseudo Jacobi-Fourier Moments (PJFM) [35], and Jacobi-Fourier Moments (JFM) [36]. Their radial basis function definitions are summarized in Table 1, which directly satisfy the orthogonality condition in Equation (7).

\begin{table*}[!t]
	\caption{Definitions of Radial Basis Functions of Unit Disk-based Orthogonal Moments}
	\centering
	\begin{tabular}{cc}
		\toprule
		Method & Radial Basis Function \\ \midrule
		ZM &  $R_{nm}^{(\rm{ZM})}(r) = \sqrt {\frac{{n + 1}}{\pi }}  \sum\limits_{k = 0}^{\frac{{n - |m|}}{2}} {\frac{{{{( - 1)}^k}(n - k)!{r^{n - 2k}}}}{{k!(\frac{{n + |m|}}{2} - k)!(\frac{{n - |m|}}{2} - k)!}}}$\\
		PZM & $R_{nm}^{(\rm{PZM})}(r) = \sqrt {\frac{{n + 1}}{\pi }} \sum\limits_{k = 0}^{n - |m|} {\frac{{{{( - 1)}^k}(2n + 1 - k)!{r^{n - k}}}}{{k!(n + |m| + 1 - k)!(n - |m| - k)!}}}$ \\ 
		OFMM & $R_n^{(\rm{OFMM})}(r) = \sqrt {\frac{{n + 1}}{\pi }}\sum\limits_{k = 0}^n {\frac{{{{( - 1)}^{n + k}}(n + k + 1)!{r^k}}}{{k!(n - k)!(k + 1)!}}}$ \\ 
		CHFM & $R_n^{(\rm{CHFM})}(r) = \frac{2}{\pi }{\left( {\frac{{1 - r}}{r}} \right)^{\frac{1}{4}}} \sum\limits_{k = 0}^{\left\lfloor {\frac{n}{2}} \right\rfloor } {\frac{{{{( - 1)}^k}(n - k)!{{(4r - 2)}^{n - 2k}}}}{{k!(n - 2k)!}}}$ \\ 
		PJFM & $R_n^{(\rm{PJFM})}(r) =\sqrt {\frac{{(n + 2)(r - {r^2})}}{{\pi (n + 3)(n + 1)}}}\sum\limits_{k = 0}^n {\frac{{{{( - 1)}^{n + k}}(n + k + 3)!{r^k}}}{{k!(n - k)!(k + 2)!}}}$ \\ 
		JFM & $R_n^{(\rm{JFM})}(p,q,r) = \sqrt {\frac{{{r^{q - 2}}{{(1 - r)}^{p - q}}(p + 2n) \cdot \Gamma (q + n) \cdot n!}}{{2\pi \Gamma (p + n) \cdot \Gamma (p - q + n + 1)}}}\sum\limits_{k = 0}^n {\frac{{{{( - 1)}^k}\Gamma (p + n + k){r^k}}}{{k!(n - k)!\Gamma (q + k)}}}$ \\ 
		RHFM & $	R_n^{(\rm{RHFM})}(r) = \left\{ {\begin{array}{*{20}{c}}
				{\frac{1}{{\sqrt {2\pi r} }}}&{n = 0}\\
				{\sqrt {\frac{1}{{\pi r}}} \sin (\pi (n + 1)r)}&{n > 0\;\& \;n\;{\rm{odd}}}\\
				{\sqrt {\frac{1}{{\pi r}}} \cos (\pi nr)}&{n > 0\;\& \;n\;{\rm{even}}}
		\end{array}} \right.$ \\ 
		EFM & $R_n^{(\rm{EFM})}(r) = \frac{1}{{\sqrt {2\pi r} }}\exp (\bm{j}2n\pi r)$ \\ 
		PCET & $R_n^{(\rm{PCET})}(r) = \frac{1}{{\sqrt \pi  }}\exp (\bm{j}2n\pi {r^2})$ \\
		PCT & $R_n^{(\rm{PCT})}(r) = \left\{ {\begin{array}{*{20}{c}}
				{\frac{1}{{\sqrt \pi  }}}&{n = 0}\\
				{\sqrt {\frac{2}{\pi }} \cos (n\pi {r^2})}&{n > 0}
		\end{array}} \right.$ \\ 
		PST & $R_n^{(\rm{PST})}(r) = \sqrt {\frac{2}{\pi }} \sin (n\pi {r^2})$ \\ 
		BFM & $	R_n^{(\rm{BFM})}(r) = \frac{1}{{\sqrt {\pi} {{{{J_{v + 1}}({\lambda _n})}}} }}{J_v}({\lambda _n}r)$, ${J_v}(x) = \sum\limits_{k = 0}^\infty  {\frac{{{{( - 1)}^k}}}{{k!\Gamma (v + k + 1)}}} {\left( {\frac{x}{2}} \right)^{v + 2k}}$ \\ \bottomrule
	\end{tabular}
\end{table*}

It is worth mentioning that the radial basis function of JFM is constructed directly from the original Jacobi polynomials [28], while the radial basis functions of ZM, PZM, OFMM, CHFM, and PJFM are all special cases of the Jacobi polynomials. Thus, JFM is, in fact, a generic expression of the above famous methods. By properly setting the values of the parameters $p$ and $q$, $R_{nm}^{(\rm{OFMM})}(r)$, $R_{nm}^{(\rm{CHFM})}(r)$ and $R_{nm}^{(\rm{PJFM})}(r)$ can be directly obtained from $R_{nm}^{(\rm{JFM})}(p,q,r)$. The relationship between $R_{nm}^{(\rm{JFM})}(p,q,r)$ and $R_{nm}^{(\rm{ZM})}(r)$/$R_{nm}^{(\rm{PZM})}(r)$ is more complicated, and we refer readers to the work of Hoang et al. [37] for more details. Here, the parameter setting is listed below:
\begin{itemize}
	\item ZM – $p = |m| + 1$ and $q = |m| + 1$;
	\item PZM – $p = 2|m| + 2$ and $q = 2|m| + 2$;
	\item OFMM – $p = 2$ and $q = 2$;
	\item CHFM – $p = 2$ and $q = 1.5$;
	\item PJFM – $p = 4$ and $q = 3$;
\end{itemize}

The unit disk-based orthogonal moments using Jacobi polynomials, especially the landmark ZM, have a long-history in optical physics, digital image processing, and pattern recognition. As one can note from the formulas listed above, however, the definitions of these radial basis functions rely on factorial/gamma terms and summations, which leads to high computational complexity. In addition, the factorial and gamma functions tend to cause the calculation errors, mainly \emph{numerical instability}. In Sections 4.1 and 4.2, we will further describe the causes and solutions to above issues.

\subsection{Harmonic Functions}
In this group, the famous methods mainly include Radial Harmonic Fourier Moments (RHFM) [38], Exponent-Fourier Moments (EFM) [39], and PHT [40]. Here, PHT consists of three different transformations: Polar Complex Exponential Transform (PCET), Polar Cosine Transform (PCT), and Polar Sine Transform (PST). Their radial basis function definitions are summarized in Table 1, which directly satisfy the orthogonality condition in Equation (7).

It can be seen that the radial basis functions of RHFM, EFM, and PHT are all based on the harmonic functions commonly used in Fourier analysis, i.e., complex exponential functions $\{ \exp (\bm{j}2n\pi r):n \in \vmathbb{Z} \} $  and trigonometric functions $\{ 1,\cos (2n\pi r),\sin (2n\pi r):n \in \vmathbb{Z}^{+}\} $ . Therefore, the definitions of the above methods are closely related, and the radial basis functions can be transformed into each other via Euler's formula $\exp (\bm{j}\alpha ) = \cos (\alpha ) + \bm{j}\sin (\alpha )$  and variable substitution $dr = \frac{1}{2}d{r^2}$ . More details on this will be given in Section 4.4. As for the calculation, compared to Jacobi polynomials, the orthogonal moments using only harmonic functions do not involve any complicated factorial/gamma terms and long summations, meaning better time complexity and numerical stability.

\subsection{Eigenfunctions}
At current stage, there are still relatively few orthogonal moments based on eigenfunctions, and the most representative one is Bessel-Fourier Moments (BFM) [41]. The radial basis function definition of BFM is included in Table 1, which directly satisfies the orthogonality condition in Equation (7).

It is observed that the radial basis function relies on infinite series and factorial/gamma terms, and also requires root-finding algorithm for Bessel functions of the first kind  ${J_v}(x)$  to determine the $n$-th zero ${\lambda _n}$. Therefore, compared with the Jacobi polynomials and harmonic functions, the orthogonal moments based on eigenfunctions have significantly higher complexity in theory.

\begin{figure*}[!t]
	\centering
	\subfigure[ZM]{\includegraphics[scale=0.18]{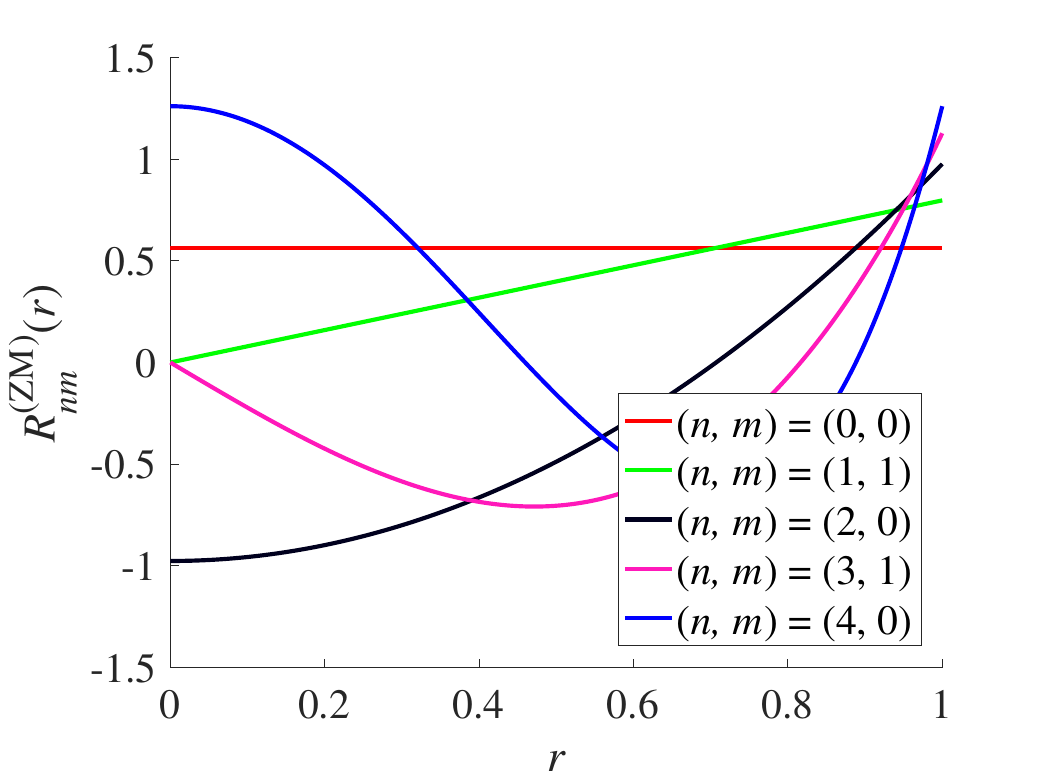}}
	\subfigure[PZM]{\includegraphics[scale=0.18]{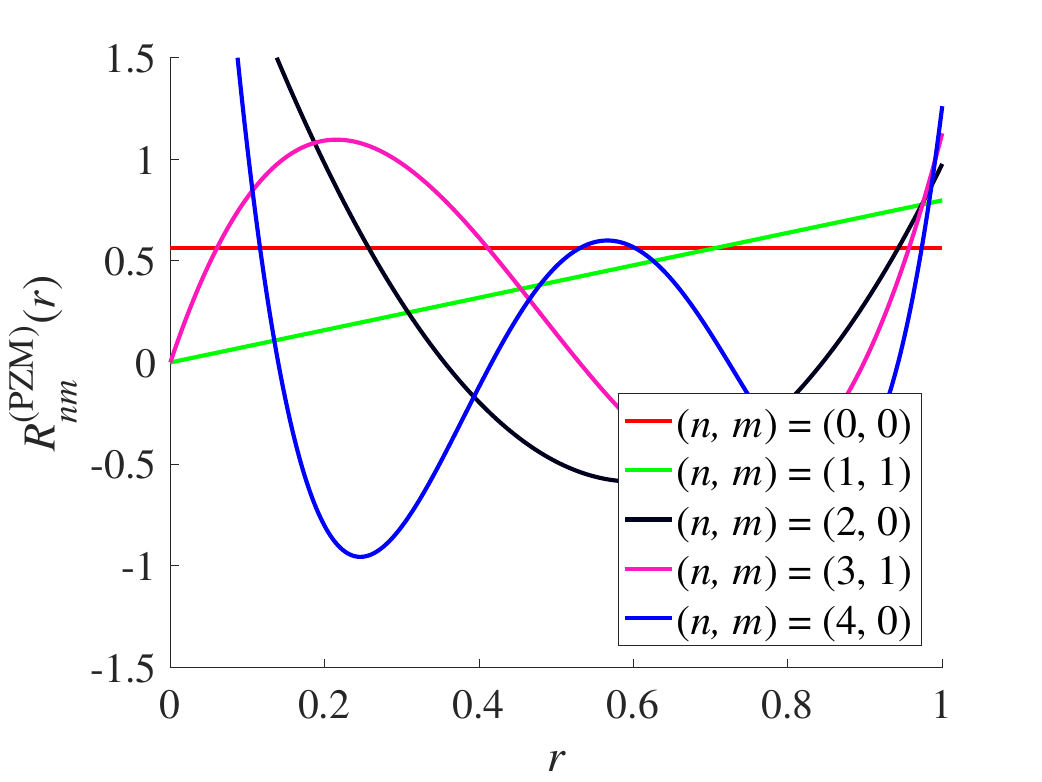}}
	\subfigure[OFMM]{\includegraphics[scale=0.18]{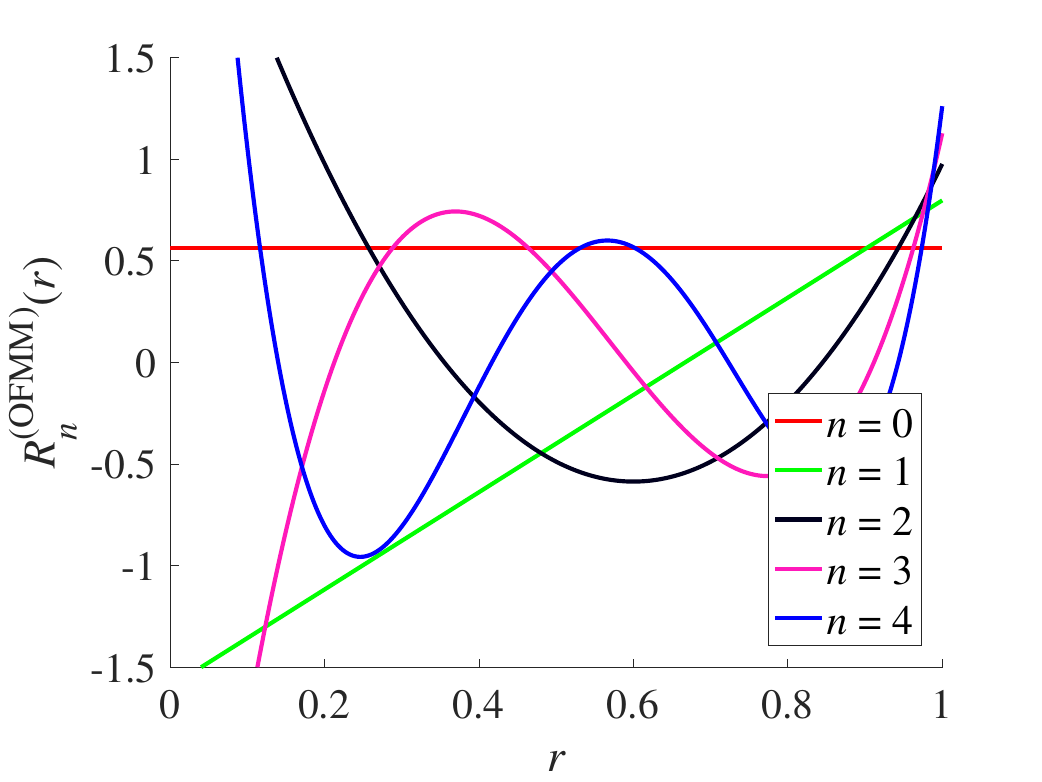}}
	\subfigure[CHFM]{\includegraphics[scale=0.18]{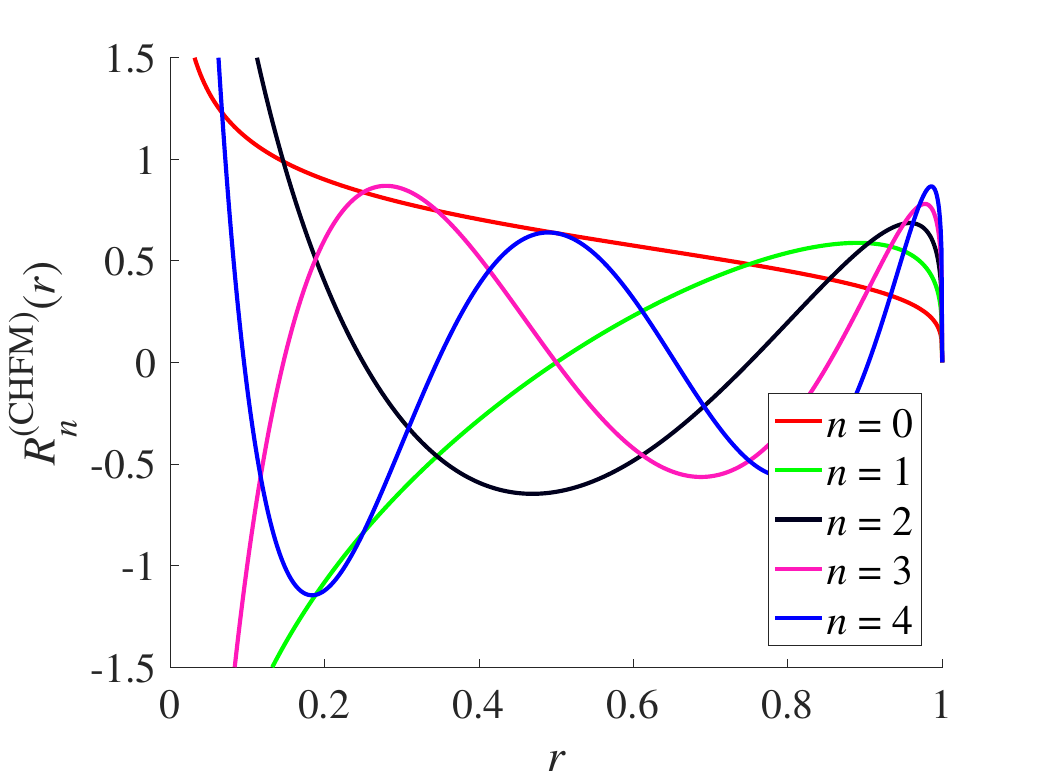}}
	\\	
	\subfigure[PJFM]{\includegraphics[scale=0.18]{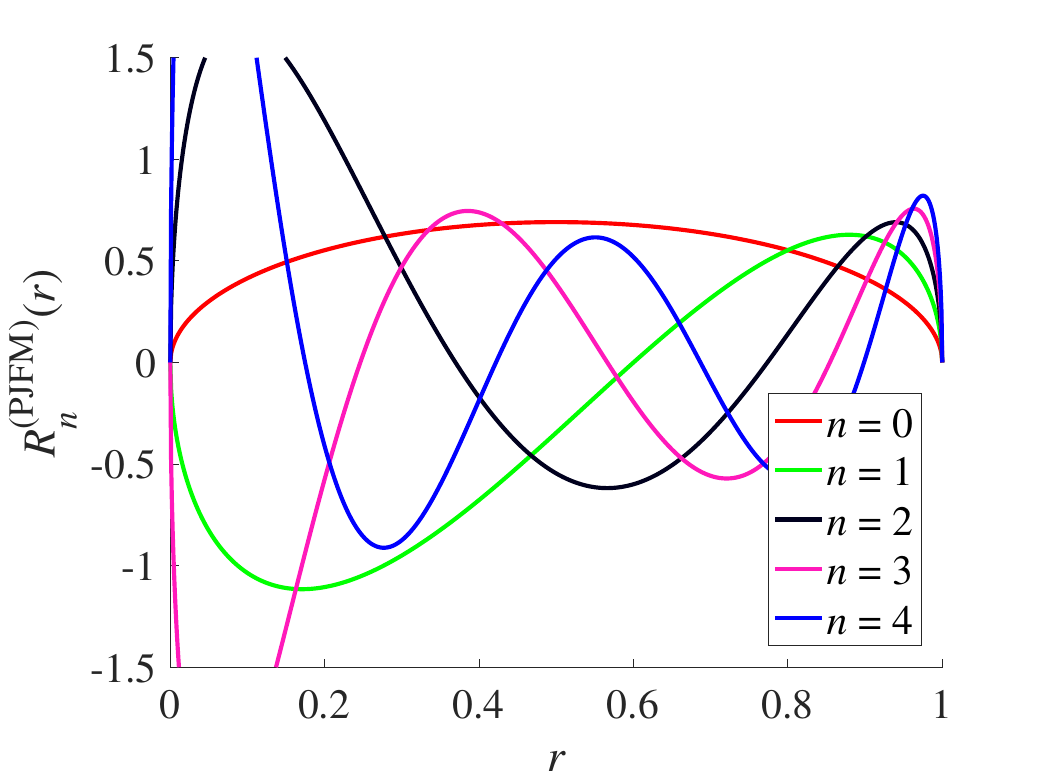}}
	\subfigure[JFM ($p,q = 5$)]{\includegraphics[scale=0.18]{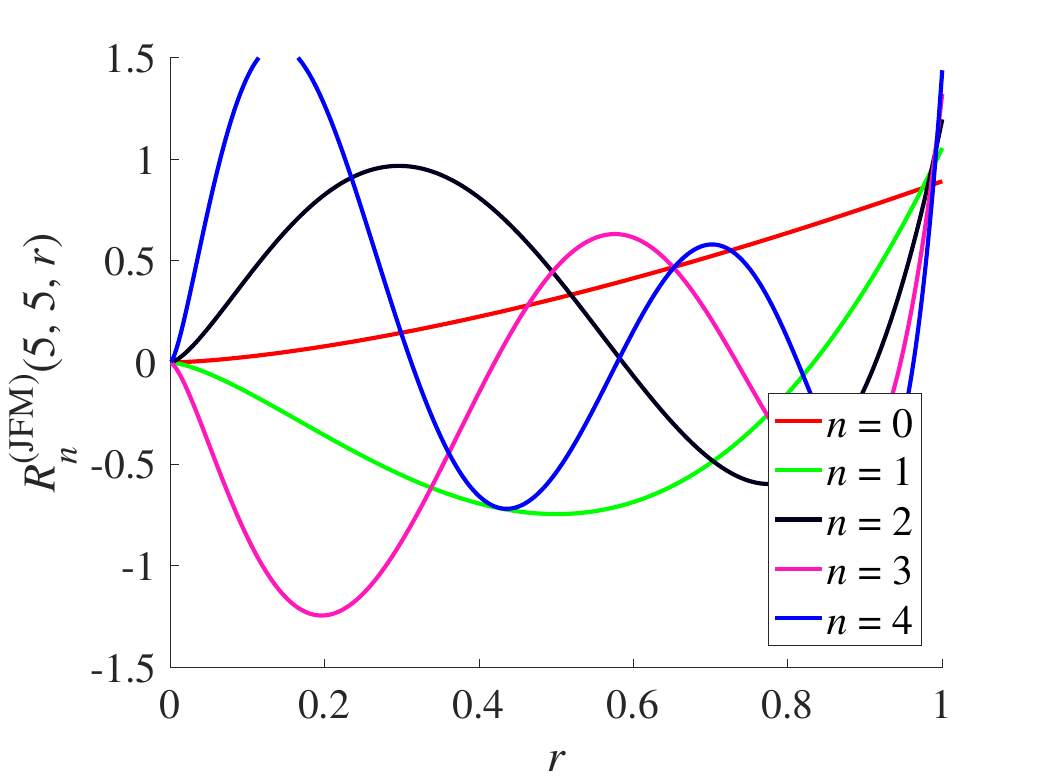}}
	\subfigure[RHFM]{\includegraphics[scale=0.18]{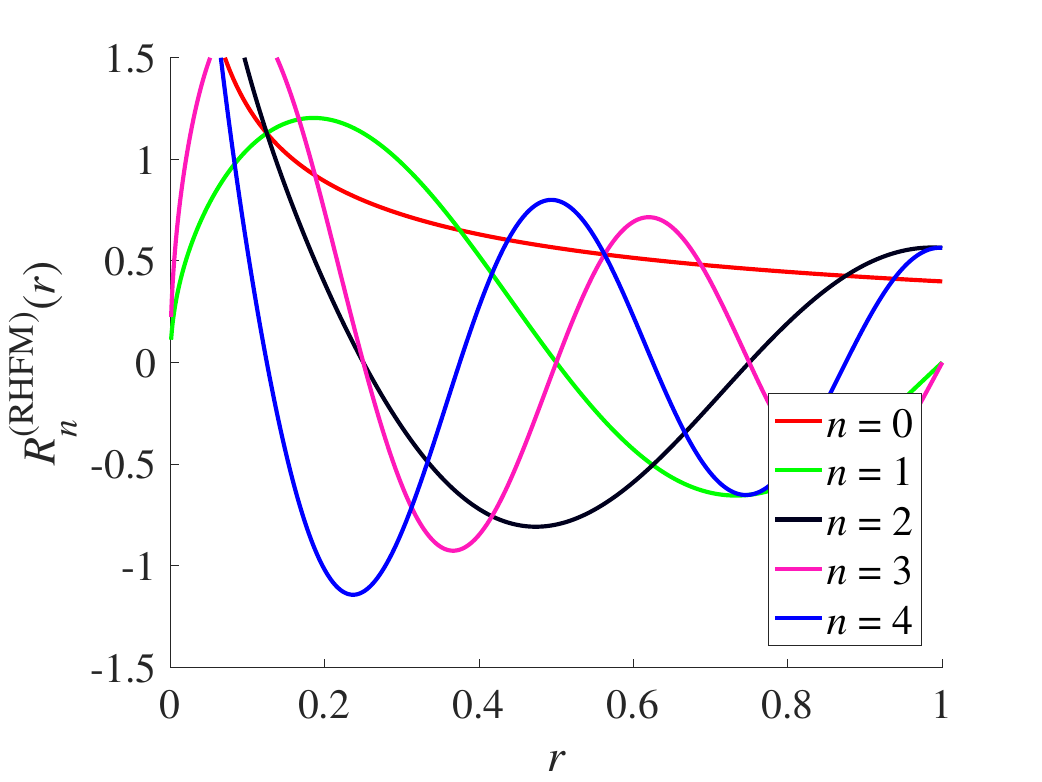}}
	\subfigure[EFM (real part)]{\includegraphics[scale=0.18]{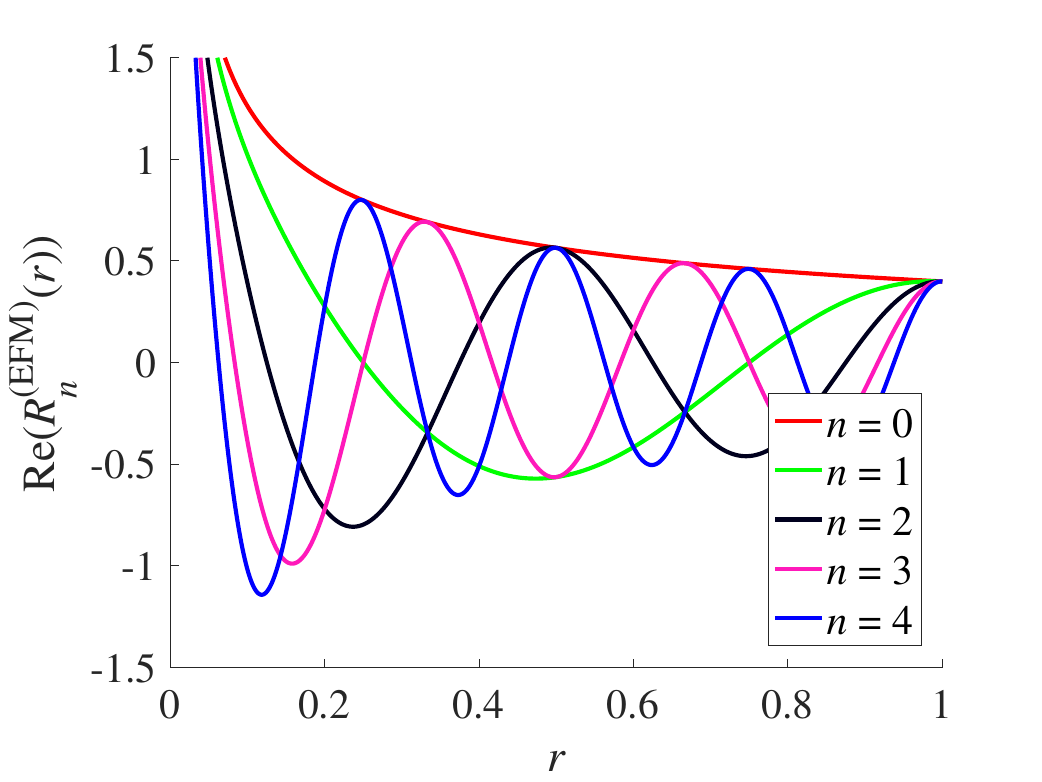}}
	\\
	\subfigure[PCET (real part)]{\includegraphics[scale=0.18]{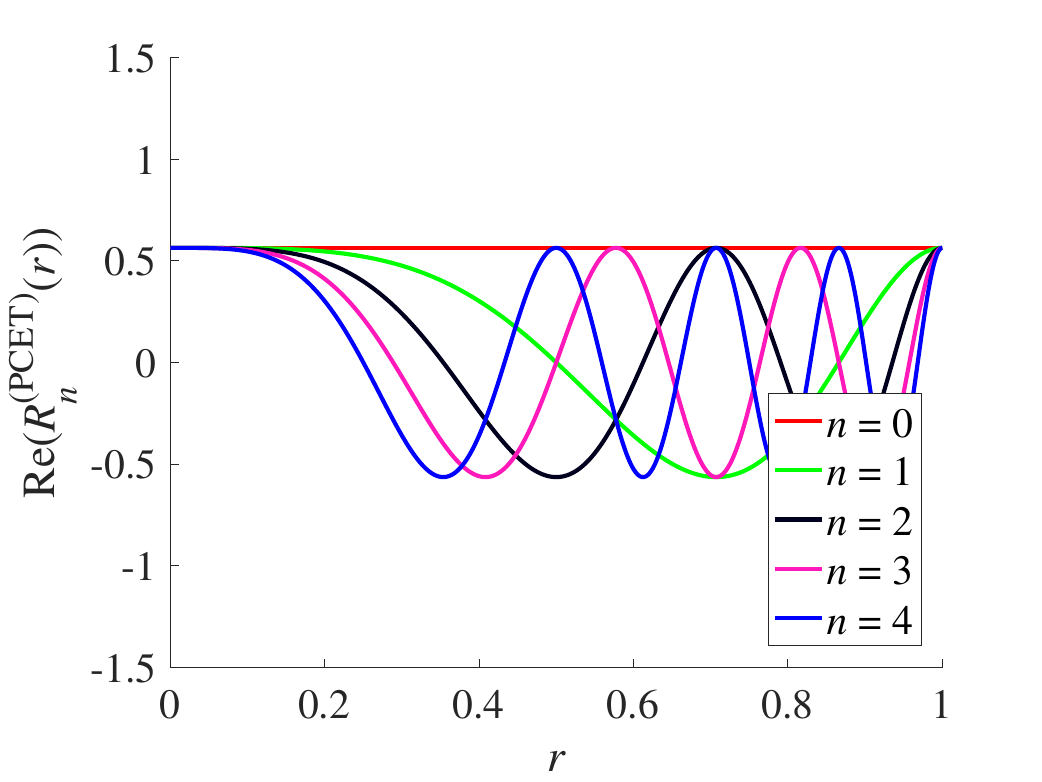}}
	\subfigure[PCT]{\includegraphics[scale=0.18]{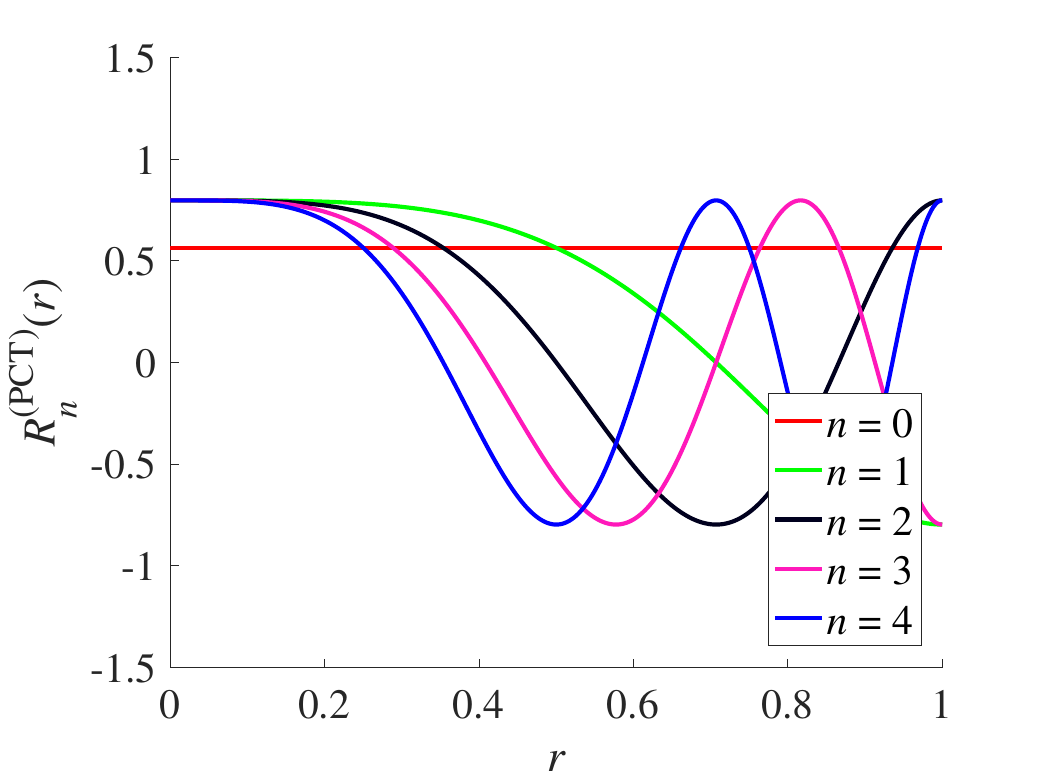}}
	\subfigure[PST]{\includegraphics[scale=0.18]{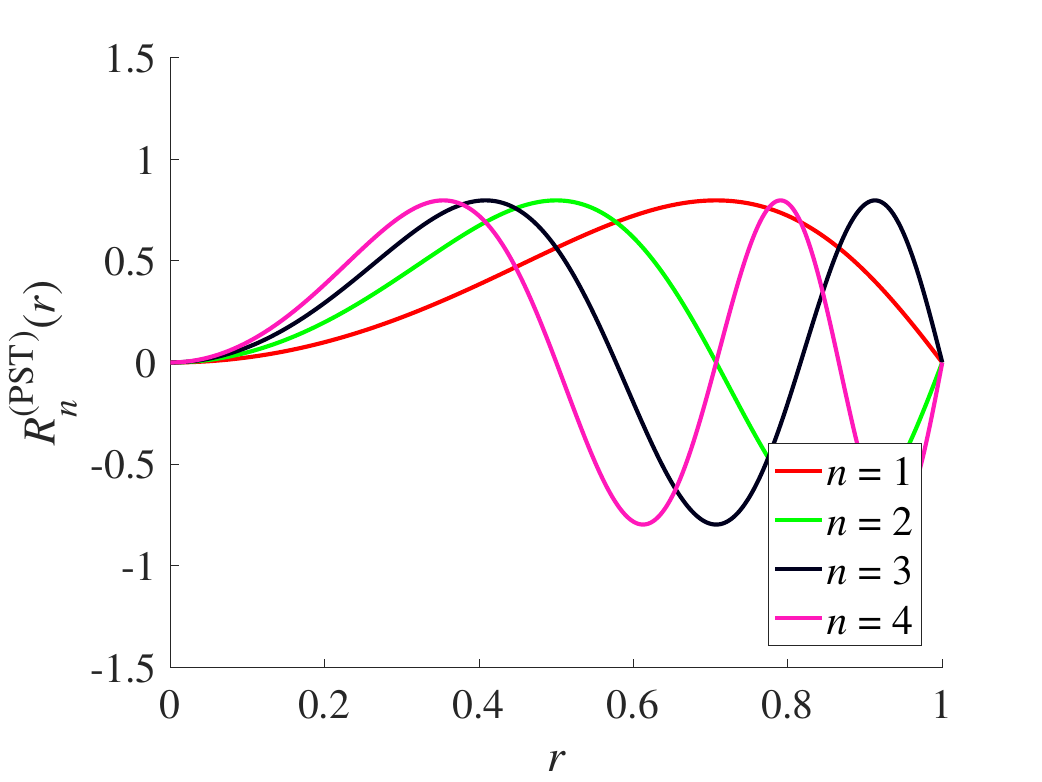}}
	\subfigure[BFM]{\includegraphics[scale=0.18]{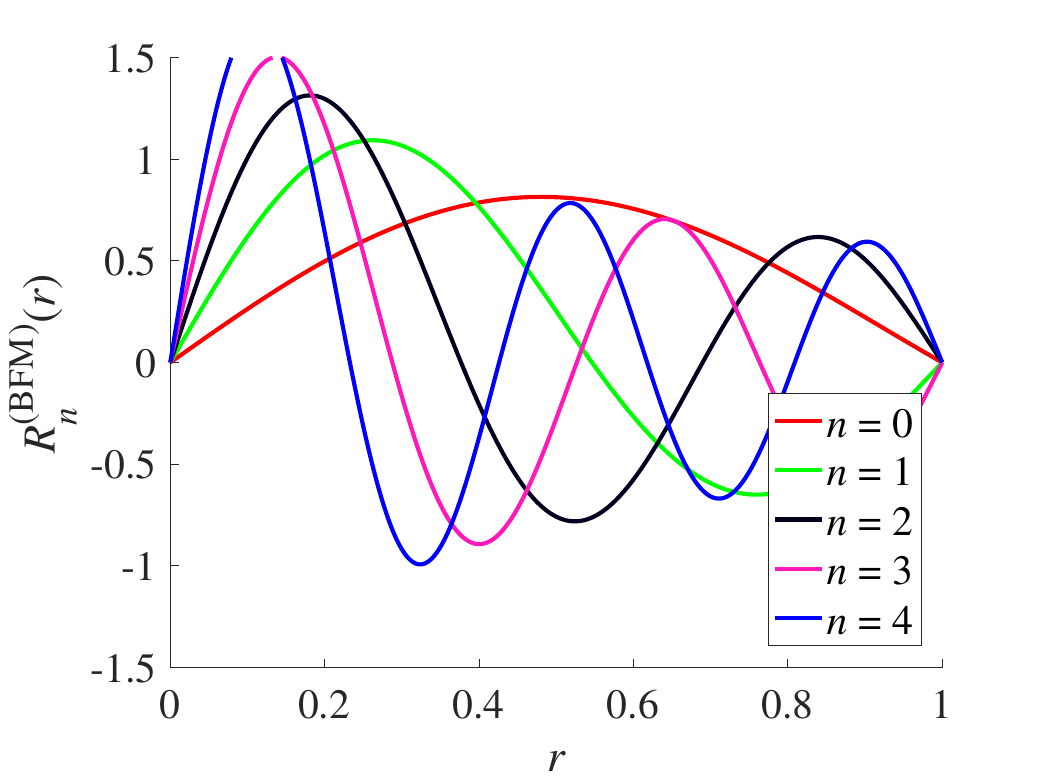}}
	\centering
	\caption{Illustrations of radial basis functions of unit disk-based orthogonal moments.}
\end{figure*}

\begin{table*}[!t]
	\caption{Mathematical Properties of Radial Basis Functions of Unit Disk-Based Orthogonal Moments}
	\centering
	\begin{tabular}{cccccc}
		\toprule
		Method & Parameter & Complexity & Stability & \tabincell{c}{Number \\ of Zeros} & \tabincell{c}{Distribution \\ of Zeros} \\ \midrule
		ZM &  \tabincell{c}{ $n \in\vmathbb{N} $, $m \in\vmathbb{Z} $, $|m| \le n$, \\ $n - |m| = {\rm{even}}$} & high & poor & $(n - |m|)/2$ & biased \\ 
		PZM & $n \in\vmathbb{N} $, $m \in\vmathbb{Z} $, $|m| \le n$ & high & poor & $n - |m|$ & biased \\ 
		OFMM & $n \in\vmathbb{N} $, $m \in\vmathbb{Z} $ & high & poor & $n$ & basically uniform \\ 
		CHFM & $n \in\vmathbb{N} $, $m \in\vmathbb{Z} $ & high & poor & $n$ & basically uniform \\ 
		PJFM & $n \in\vmathbb{N} $, $m \in\vmathbb{Z} $ & high & poor & $n$ & basically uniform \\ 
		JFM & \tabincell{c}{$n \in\vmathbb{N} $, $m \in\vmathbb{Z} $, $p,q\in\vmathbb{R}$, \\ $p - q>-1$, $q > 0$} & high & poor & $n$ & basically uniform \\ 
		RHFM & $n \in\vmathbb{N} $, $m \in\vmathbb{Z} $ & low & medium & $n$ & uniform \\ 
		EFM & $n \in\vmathbb{Z} $, $m \in\vmathbb{Z} $ & low & medium & $2n$ & uniform \\ 
		PCET & $n \in\vmathbb{Z} $, $m \in\vmathbb{Z} $ & low & good & $2n$  & biased \\ 
		PCT & $n \in\vmathbb{N} $, $m \in\vmathbb{Z} $ & low & good & $n$  & biased \\ 
		PST & $n \in\vmathbb{N}^{+} $, $m \in\vmathbb{Z} $ & low & good & $n-1$  & biased \\ 
		BFM & \tabincell{c}{ $n \in\vmathbb{N} $, $m \in\vmathbb{Z} $, $v\in\vmathbb{R}$, \\ ${\lambda _n} = n$-th zero of ${J_v}(x)$} & very high & medium & $n$  & basically uniform \\ \bottomrule
	\end{tabular}
\end{table*}

\subsection{Summary and Discussion}

For a better perception, the illustrations of radial basis functions of unit disk-based orthogonal moments are summarized in Figure 3.

Furthermore, we reveal the mathematical properties of these methods in Table 2, including radial basis functions’ parameter, computational complexity, numerical stability, number of zeros, and distribution of zeros. The complexity of basis functions (considering only the definition-style computation) is graded as low, high, and very high based on whether the definition involves factorial/gamma terms, summation/series operations, and root-finding processes. Numerical stability is graded as poor, medium, and good depending on whether the function includes factorial/gamma terms and very high absolute-values (mainly unbounded). The number of zeros of radial kernels is related to the ability of moments for capturing image high-frequency information [17, 40]. Besides the quantity, radial kernel’s another key attribute is the distribution of zeros, because it is related to the description emphasis of the moments in image plane [17, 40]. When the essential discriminative information is distributed uniformly in the spatial-domain, unfair emphasis of the extracted moments on certain regions has been shown to have a negative impact on the discrimination quality, termed as \emph{information suppression problem} [21].

It can be seen in Table 2 that, among these unit disk-based orthogonal moments, almost no method has both low complexity, good stability, large number and unbiased zeros. This observation strongly motivates the design of the improvement strategies that address the above common shortcomings, which will be discussed in Section 4.

\begin{figure*}[!t]
	\centering
	\includegraphics[scale=0.3]{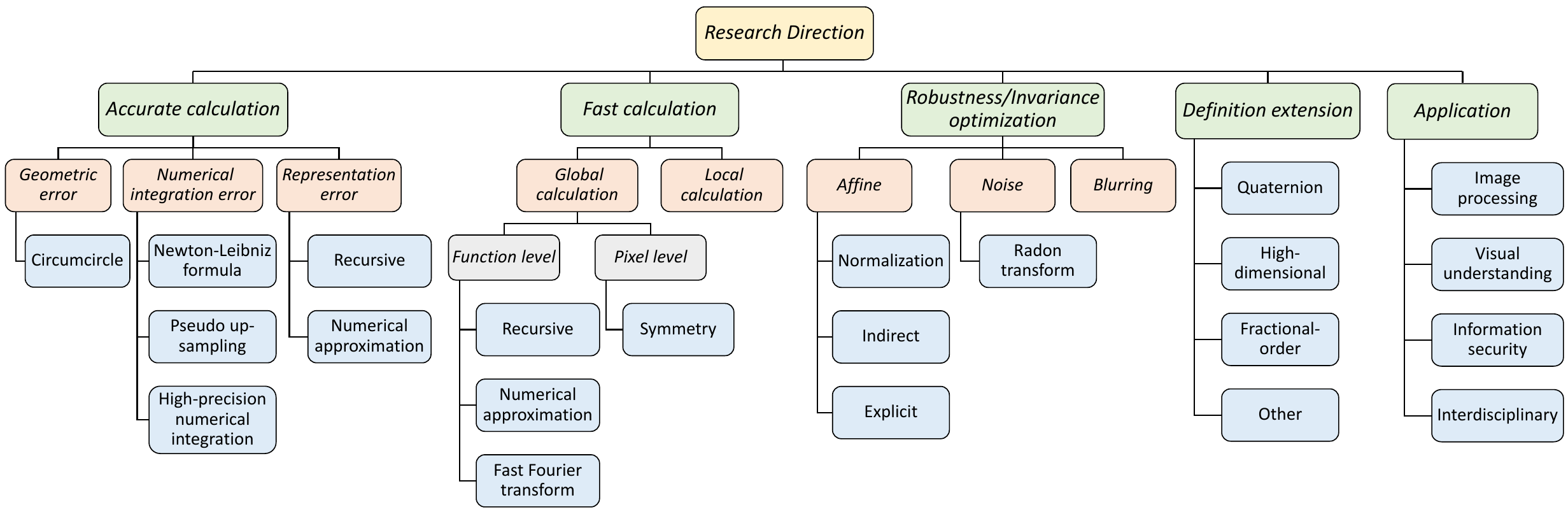}
	\centering
	\caption{The research directions of image moments.}
\end{figure*}

\section{Research Directions and Recent Advances}

In addition to defining new image moments, existing work mainly focuses on optimizing the classical moments listed in Sections 2 and 3. As shown in Figure 4, the current research directions generally include: accurate calculation, fast calculation, robustness/invariance optimization, definition extension, and application. This section will introduce the above directions along with the recent advances.

\subsection{Accurate Calculation}

As the mathematical background of Sections 4.1 and 4.2, the general procedure for calculating image moments is first given.

Going back to Equation (1), for computing the image moments $ \left< f,{V_{nm}} \right> $ of a digital image defined in the discrete Cartesian grid $\{ f(i,j):(i,j) \in {[1,2,...,N]^2}\} $, it is first necessary to unify the domains of $f(i,j)$  and ${V_{nm}}(x,y)$, i.e., to design a mapping between the two:
\begin{equation}
	(i,j) \to ({x_i},{y_j}),
\end{equation}
where a pixel region $[i - \frac{{\Delta i}}{2},i + \frac{{\Delta i}}{2}] \times [j - \frac{{\Delta j}}{2},j + \frac{{\Delta j}}{2}]$  centered at $(i,j)$  is mapped into a region $[{x_i} - \frac{{\Delta {x_i}}}{2},{x_i} + \frac{{\Delta {x_i}}}{2}] \times [{y_j} - \frac{{\Delta {y_j}}}{2},{y_j} + \frac{{\Delta {y_j}}}{2}]$  centered at $({x_i},{y_j})$. By the coordinate mapping $(i,j) \to ({x_i},{y_j})$, Equation (1) can be written down in discrete form as follows:
\begin{equation}
	\left< f,{V_{nm}}\right> \; = \;\sum\limits_{({x_i},{y_j}) \in D} {{h_{nm}}({x_i},{y_j})f(i,j)},
\end{equation}
where ${h_{nm}}({x_i},{y_j})$ is the integral value of the basis function ${V_{nm}}$  over the mapped pixel region $[{x_i} - \frac{{\Delta {x_i}}}{2},{x_i} + \frac{{\Delta {x_i}}}{2}] \times [{y_j} - \frac{{\Delta {y_j}}}{2},{y_j} + \frac{{\Delta {y_j}}}{2}]$, defined as:
\begin{equation}
	{h_{nm}}({x_i},{y_j}) = \int\limits_{{x_i} - \frac{{\Delta {x_i}}}{2}}^{{x_i} + \frac{{\Delta {x_i}}}{2}} {\int\limits_{{y_j} - \frac{{\Delta {y_j}}}{2}}^{{y_j} + \frac{{\Delta {y_j}}}{2}} {V_{nm}^*(x,y)dxdy} } .
\end{equation}

In general, the calculation of unit disk-based orthogonal moments suffers from \emph{geometric error}, \emph{numerical integration error}, and \emph{representation error} (mainly numerical instability). These errors will severely restrict the quality of image representation, especially when high-order moments are required to better describe the image. Hence, the accurate computation strategies of moments are vital for the applicability.

\subsubsection{Geometric Error}
Geometric error may occur when mapping the image domain into the basis function domain, i.e., Equation (8). Such errors are common in the calculation of the unit disk-based orthogonal moments, because digital images are generally defined over a square region in Cartesian coordinate system, rather than the unit disk.

As for mapping between the square region and the unit disk, there are naturally, as shown in Figure 5, the \emph{incircle} mapping [42]:
\begin{equation}
	\left\{ {\begin{array}{*{20}{c}}
			{i \to {x_i} = \frac{{2i - N}}{N} = i\Delta {x_i} - 1}\\
			{j \to {y_j} = \frac{{2j - N}}{N} = j\Delta {y_j} - 1}
	\end{array}} \right.,
\end{equation}
and the \emph{circumcircle} mapping [43]:
\begin{equation}
	\left\{ {\begin{array}{*{20}{c}}
			{i \to {x_i} = \frac{{2i - N}}{{\sqrt 2 N}} = i\Delta {x_i} - \frac{{\sqrt 2 }}{2}}\\
			{j \to {y_j} = \frac{{2j - N}}{{\sqrt 2 N}} = j\Delta {y_j} - \frac{{\sqrt 2 }}{2}}
	\end{array}} \right..
\end{equation}

\begin{figure}[!t]
	\centering
	\subfigure[Incircle]{\includegraphics[scale=0.3]{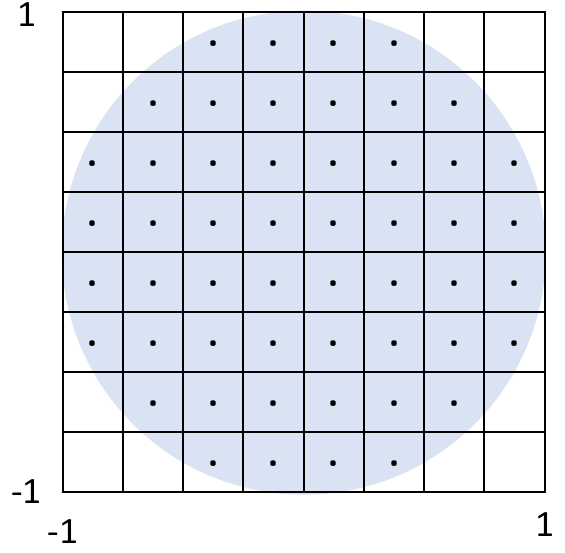}}
	\subfigure[Circumcircle]{\includegraphics[scale=0.3]{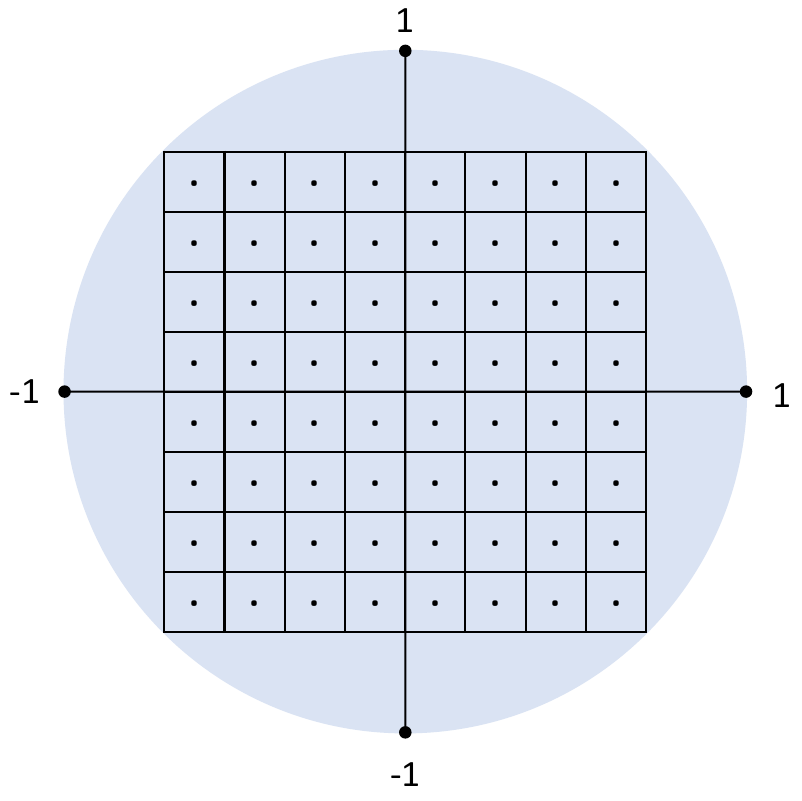}}
	\caption{The coordinate mapping between the square region of digital image and the unit disk.}
\end{figure}

When the incircle mapping is used, there exists $(i,j)$  such that $({x_i},{y_j}) \notin D$, i.e., the pixels that are mapped to the outside of the unit disk are not counted, and there exists  $(i,j)$  such that $[{x_i} - \frac{{\Delta {x_i}}}{2},{x_i} + \frac{{\Delta {x_i}}}{2}] \times [{y_j} - \frac{{\Delta {y_j}}}{2},{y_j} + \frac{{\Delta {y_j}}}{2}] - D \ne \emptyset $ , i.e., some mapped pixel regions partially intersect the unit disk. In both cases, geometric errors occur.

As for the circumcircle mapping, it is able to completely avoid such two cases, so there will be no geometric errors. However, this comes at the cost of representation capability [17], as the mapped regions containing image information occupy only $\frac{2}{\pi }$  of the entire unit disk.

\subsubsection{Numerical Integration Error}

Numerical integration error may occur when calculating the integral of the continuous basis functions, i.e., Equation (10). Here, for discrete moments, the basis functions ${V_{nm}}$  are generally constant on the interval $[{x_i} - \frac{{\Delta {x_i}}}{2},{x_i} + \frac{{\Delta {x_i}}}{2}] \times [{y_j} - \frac{{\Delta {y_j}}}{2},{y_j} + \frac{{\Delta {y_j}}}{2}]$. Thus, it is easy to check that ${h_{nm}}({x_i},{y_j}) = V_{nm}^*({x_i},{y_j})\Delta {x_i}\Delta {y_j}$, meaning discrete moments typically do not involve numerical integration errors [23]. As for continuous moments, since the basis functions  ${V_{nm}}$  are continuous over the interval  $[{x_i} - \frac{{\Delta {x_i}}}{2},{x_i} + \frac{{\Delta {x_i}}}{2}] \times [{y_j} - \frac{{\Delta {y_j}}}{2},{y_j} + \frac{{\Delta {y_j}}}{2}]$, solving ${h_{nm}}({x_i},{y_j})$ requires some integration tricks.

As the simplest case, there is an analytical solution to ${h_{nm}}({x_i},{y_j})$, i.e., Equation (10) can be solved directly by the \emph{Newton-Leibniz formula} [44, 45]. Such calculations also do not involve numerical integration errors.

As the more general case, considering the complication of the definition of many basis functions ${V_{nm}}$, it is often difficult to determine the analytical solution and some approximate algorithms are needed for achieving the numerical solution of ${h_{nm}}({x_i},{y_j})$. The most commonly used approximation algorithm is \emph{Zero-Order Approximation} (ZOA) [42], which imitates the calculation of discrete moments, as follows:

\begin{equation}
	{h_{nm}}({x_i},{y_j}) \simeq V_{nm}^*({x_i},{y_j})\Delta {x_i}\Delta {y_j}.
\end{equation}

Generally, the accuracy of the numerical integration method is inversely proportional to the interval area $\Delta {x_i}\Delta {y_j}$. Therefore, by further dividing a single pixel region into multiple smaller integration intervals (e.g., $3 \times 3$ sub-intervals), higher accuracy can be easily obtained. This strategy is often called \emph{pseudo up-sampling} [46, 47]. When the ZOA is used in these sub-intervals, the integral can be expressed as

\begin{equation}
	{h_{nm}}({x_i},{y_j}) \simeq \sum\limits_{(a,b)} {V_{nm}^*({u_a},{v_b})\Delta {u_a}\Delta {v_b}},
\end{equation}
where $({u_a},{v_b})$ is the sampling point with ${u_a} \in [{x_i} - \frac{{\Delta {x_i}}}{2},{x_i} + \frac{{\Delta {x_i}}}{2}]$ and ${v_b} \in [{y_j} - \frac{{\Delta {y_j}}}{2},{y_j} + \frac{{\Delta {y_j}}}{2}]$. In addition to this simple approach, other more complex \emph{high-precision numerical integration} strategies [48–50], such as the \emph{Gaussian quadrature rule} and \emph{Simpson's rule}, can also be used in the computation of Equation (10). Their general definition is
\begin{equation}
	{h_{nm}}({x_i},{y_j}) \simeq \sum\limits_{(a,b)} {{w_{ab}}V_{nm}^*({u_a},{v_b})\Delta {x_i}\Delta {y_j}},
\end{equation}
where ${w_{ab}}$ is the weight corresponding to the sampling point  $({u_a},{v_b})$.

It is worth noting that all above discussion in Section 4.1 relies on the Cartesian coordinate system, which we call Cartesian based calculation method. Correspondingly, there also exist a calculation method based on polar coordinate system for unit disk-based orthogonal moments, often referred as \emph{polar pixel tiling} [51, 52].

\begin{figure}[!t]
	\centerline{
		\includegraphics[scale=0.3]{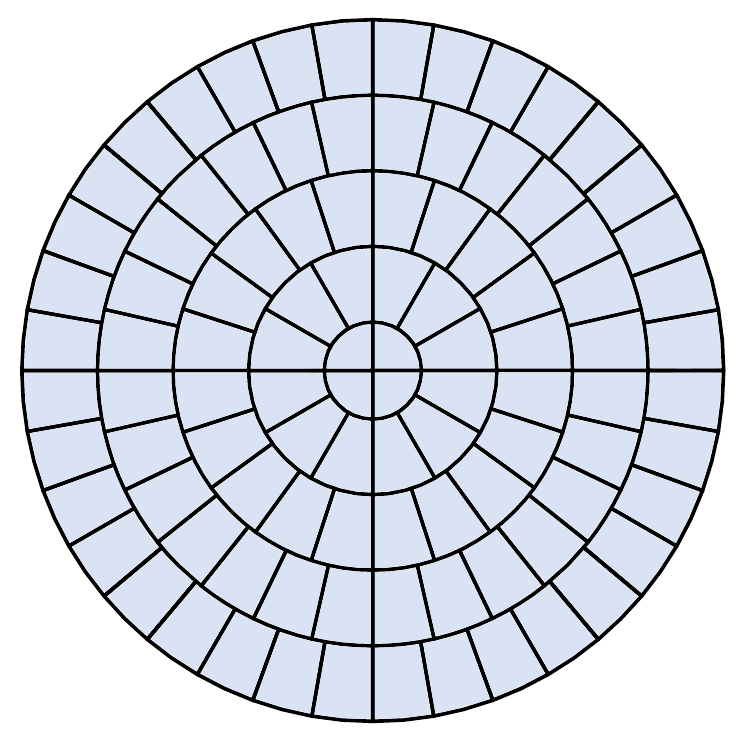}}
	\caption{The coordinate mapping in polar pixel tiling scheme.}
\end{figure}

It first resamples the digital image $f(i,j)$ to a discrete polar grid $({r_u},{\theta _{uv}})$, as shown in Figure 6, and then performs calculation in a manner similar to Equation (9) and Equation (10):
\begin{equation}
	\left< f,{V_{nm}} \right> \; = \;\sum\limits_{({r_u},{\theta _{uv}}) \in D} {{h_{nm}}({r_u},{\theta _{uv}})f({r_u},{\theta _{uv}})},
\end{equation}
with

\begin{equation}
	\begin{split}
		{h_{nm}}({r_u},{\theta _{uv}}) = \int\limits_{{r_u}}^{{r_{u + 1}}} {\int\limits_{{\theta _{uv}}}^{{\theta _{u(v + 1)}}} {V_{nm}^*(r,\theta )rdrd\theta } } 
		= \int\limits_{{r_u}}^{{r_{u + 1}}} {{R_n}(r)rdr} \int\limits_{{\theta _{uv}}}^{{\theta _{u(v + 1)}}} {\exp ( - \bm{j}m\theta )d\theta } .
	\end{split}
\end{equation}

Note that Equation (17) has a distinct advantage, i.e., the computation of ${h_{nm}}({r_u},{\theta _{uv}})$ can be separated into two independent parts [53]: 1) $\int\limits_{{r_u}}^{{r_{u + 1}}} {{R_n}(r)rdr}$  can be approximately integrated by pseudo up-sampling and Gaussian quadrature rule; 2)  $\int\limits_{{\theta _{uv}}}^{{\theta _{u(v + 1)}}} {\exp ( - \bm{j}m\theta )d\theta }$ can be exactly integrated by Newton-Leibniz formula, as follows:

\begin{equation}
	\begin{split}
		\int\limits_{{\theta _{uv}}}^{{\theta _{u(v + 1)}}} {\exp ( - \bm{j}m\theta )d\theta }
		= \left\{ {\begin{array}{*{20}{c}}
				{\frac{{\bm{j}[\exp ( - \bm{j}m{\theta _{u(v + 1)}}) - \exp ( - \bm{j}m{\theta _{uv}})]}}{m}}&{m \ne 0}\\
				{{\theta _{u(v + 1)}} - {\theta _{uv}}}&{m = 0}
		\end{array}} \right..
	\end{split}
\end{equation}

In addition to above advantage, the polar pixel tiling has similar properties to the Cartesian based calculation method in terms of geometric error and numerical integration error, and will not be repeated here.

\subsubsection{Representation Error}
Representation error is caused by the finite precision of the numerical computing systems, which occurs in all processes of the calculation, i.e., Equation (8), Equation (9), and Equation (10). The representation error can be further divided into \emph{overflow error}, \emph{underflow error}, and \emph{roundoff error} [17]. The common numerical instability is mainly attributable to roundoff error and overflow error.

The basis functions based on Jacobi polynomials contain factorial/gamma terms. When the order is large, the actual values of these coefficients may exceed the representation range of the numerical computing system, resulting in roundoff error or even overflow error. To avoid direct calculation of the factorial/gamma terms, \emph{recursive} strategies [54–56] and \emph{numerical approximation} algorithms [57, 58] are often used. The recursive method relies on the recursive relationship of the basis functions derived from $a! = a \cdot (a - 1)!$  and $\Gamma (a) = a \cdot \Gamma (a - 1)$, using several low-order basis functions to directly derive the high-order ones. This process does not involve the factorial/gamma of large number. In another path, the numerical approximation method achieves factorial-free calculations by resorting to a suitable approximation for factorials such as \emph{Stirling's formula}.

In addition to factorial/gamma terms, the basis functions of some orthogonal moments have very high (or even infinite) absolute-values at certain points, which may also exceed the representation range of the numerical computing system. For example, although the definitions of RHFM and EFM do not involve any factorial/gamma terms, their basis functions are infinite at the origin (see also Figure 3), which will cause roundoff error or even overflow error.

\subsubsection{Recent Advance of Accurate Calculation}
In the field of accurate calculation for unit disk-based orthogonal moments, Xin et al.’s polar pixel tiling [51] may be the most influential and representative work. Recent papers typically combine polar pixel tiling with other techniques, including circumcircle mapping, pseudo up-sampling, high-precision numerical integration, Newton-Leibniz formula, and recursive strategy, to provide the state-of-the-art performance.

In this way, Sáez-Landete [59] integrated the work of Camacho-Bello et al. [53] and Upneja et al. [56], giving an accurate calculation algorithm for the JFM. Hence, it also applies to other Jacobi polynomial based moments such as OFMM and CHFM. Due to the polar pixel tiling, the complicated radial parts can be evaluated using Gaussian quadrature, pseudo up-sampling, and recursive relationship, while the angular parts can be exactly evaluated by Newton-Leibniz formula. A somewhat similar method was proposed by Hosny et al. [60] for the harmonic function based PHT. The main difference is that, due to the simple definition of the radial basis function, the radial parts can also be integrated exactly by the Newton-Leibniz formula just like the angular parts.

These two recent works provide near-perfect accuracy for calculating Jacobi polynomial based moments and harmonic function based moments, respectively. To be critical, the only flaw may be the error introduced in the process of image resampling to discrete polar grid [51]. If the nonlinear interpolation methods are used, such as bicubic method, the interpolation error is negligible for image representation tasks. On the other hand, if a given task is sensitive to such errors, other more complex mathematical tools, such as pseudo-polar [61], are instructive for the design of accurate calculations [62].

\subsection{Fast Calculation}

In the implementation of orthogonal moments, the overall complexity may become excessively high when 1) a large number of moments is needed, 2) the image has high resolution, 3) many images need to be processed, or 4) a high-precision computation is required. Since these requirements are common in practical applications, the fast calculation of orthogonal moments is strongly demanded. Depending on the application scenario, optimization efforts for computational speed can be divided into two categories: \emph{global calculation} and \emph{local calculation}.

\begin{figure}[!t]
	\centerline{
		\includegraphics[scale=0.32]{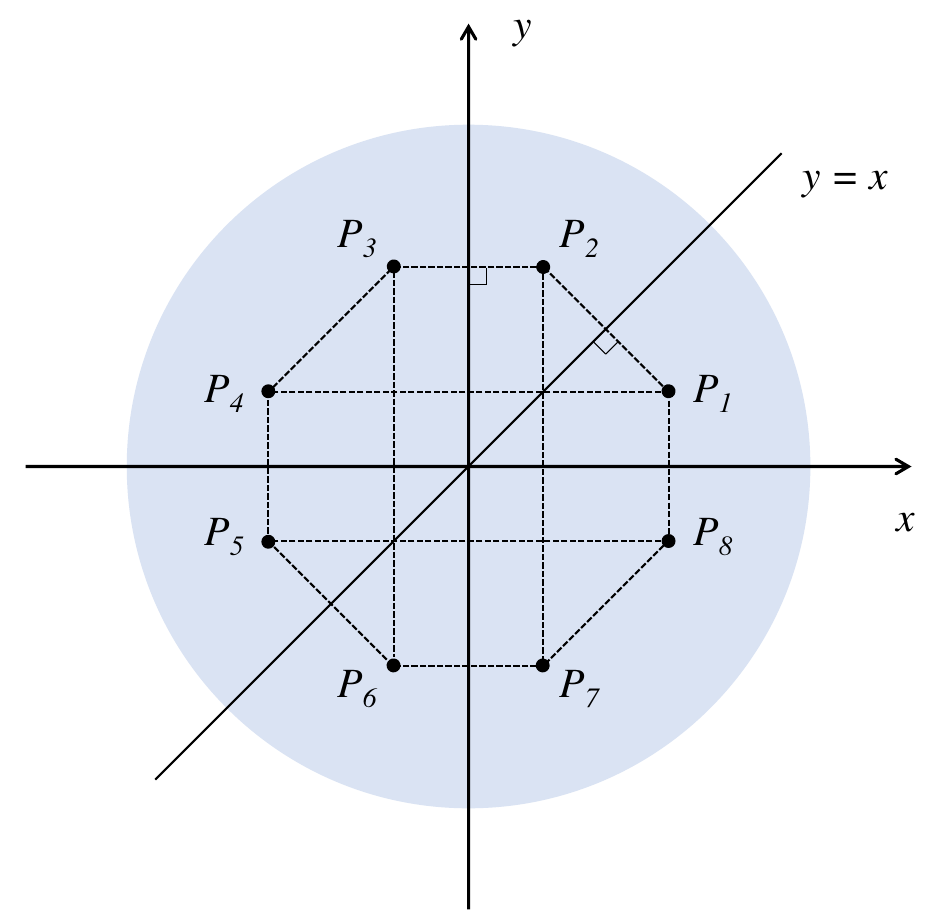}}
	\caption{Illustrations of symmetrical points in the unit disk.}
\end{figure}

\subsubsection{Global Calculation}
Global calculation is to calculate the image moments for the entire image. In this scenario, the number of moments and the resolution of the image are the main factors that affect the time complexity. Taking the simplest ZOA-based direct calculation as an example, the time cost of an  $n$-order image moment comes from:

\begin{itemize}
	\item For Equation (10), it is necessary to evaluate the values of the basis function ${V_{nm}}$ at $N \times N$ sampling points, i.e., $\{ V_{nm}^*({x_i},{y_j}):(i,j) \in {[1,2,...,N]^2}\} $. The time complexity is related to the definition of the basis functions. For example, the basis functions using Jacobi polynomials require
	$\mathcal{O}(n{N^2})$ additions for the summation. In contrast, the basis functions using harmonic functions do not involve such additions.
	\item For Equation (9), it is necessary to calculate the inner product of the basis function ${V_{nm}}$  and the digital image $f$  at  $N \times N$  sampling points, i.e., $\left< f(i,j),V_{nm}^*({x_i},{y_j})\Delta {x_i}\Delta {y_j} \right> $. The time complexity is  $\mathcal{O}({N^2})$ multiplications and  $\mathcal{O}({N^2})$ additions.
\end{itemize}

Note that the calculations listed above are only required for one moment. Let $K$ be some integer constant, if all the moments of orders in set $\{ (n,m):|n|,|m| \le K\} $  are computed, the total complexity will increase by a factor of $\mathcal{O}({K^2})$. Moreover, the computational cost of Equation (10) may rise sharply if the high-precision numerical integration strategy, such as Gaussian quadrature, is adopted. To reduce the complexity of Equation (9) and Equation (10), the existing methods are designed at the \emph{function level} and the \emph{pixel level}.

\begin{itemize}
	\item Function level: For the orthogonal moments based on Jacobi polynomials, if the recursive method and numerical approximation (see also Section 4.1.3) are used to evaluate the basis functions, the number of additions (from summation) and multiplications (from factorial) in the calculation of Equation (10) can be reduced [54–58]. For example, the recursive calculation of polynomial  ${R_n}(\sqrt {{x_i}^2 + {y_j}^2} )$ requires $\mathcal{O}({N^2})$  additions, less than the $\mathcal{O}(n{N^2})$  in direct computation. If $K$ polynomials ${R_n}(\sqrt {{x_i}^2 + {y_j}^2} )$  of orders $\{ 0,1,...,K\} $ are required, the direct method involves  $\mathcal{O}({K^2N^2})$ additions, while the recursive scheme only requires  $\mathcal{O}({KN^2})$  additions. For the orthogonal moments based on harmonic functions, in addition to adopting the similar recursive form [63–66], more effective strategy is to make use of its inherent relationship with Fourier transform [67–71]. Once the explicit relationship between the two is determined, the \emph{Fast Fourier Transform} (FFT) algorithm can be used to calculate the moments. Note that FFT has the ability to reduce the number of the most time-consuming multiplications in Equation (9) from $\mathcal{O}({N^2})$  to $\mathcal{O}({N\log N})$.
	\item Pixel level: The most common strategy in this path is to simplify the calculation by exploring the \emph{symmetry} and \emph{anti-symmetry} of the basis functions in the domain [72, 73]. More specifically, the basis function values at all  $N \times N$  sampling points can be completely derived by the basis function values at a few special sampling points, thus reducing the complexity of Equation (10) and Equation (9). In Figure 7, we give an illustrations of symmetrical points in the unit disk. For a point ${P_1}$ , there are seven symmetrical points $\{ {P_2},{P_3},{P_4},{P_5},{P_6},{P_7},{P_8}\} $  w.r.t. coordinate axes, origin, and $y=x$. Their Cartesian coordinates and polar coordinates are listed in Table 3. It can be seen that all these points have the same radial coordinate and related angular coordinate. Based on the mathematical properties of the complex exponential function, i.e., the trigonometric identities, such correlation of coordinates will be directly converted to the correlation of basis function values. Hence, this observation can lead to a reduction in computational complexity of Equation (10) and Equation (9) by approximately $1/8$. Considering that all unit disk-based orthogonal moments are based on angular basis functions using complex exponential functions, the above symmetry and anti-symmetry properties are maintained in all these methods. As a result, this fast calculation strategy is generic and easy to use in combination with other fast algorithms.
\end{itemize}

\begin{table}[!t]
	\caption{Cartesian Coordinates and Polar Coordinates of the Symmetrical Points}
	\centering
	\begin{tabular}{cccc}
		\toprule
		 Symmetrical Point &  Symmetrical Axis & Cartesian Coordinates & Polar Coordinates \\ \midrule
		${P_1}$ &  & $(x,y)$ & $(r,\theta )$ \\ 
		${P_2}$ & $y=x$ &  $(y,x)$ & $(r,\frac{\pi }{2} - \theta )$ \\
		${P_3}$ & $y=x$, $y$-axis &  $(-y,x)$ & $(r,\frac{\pi }{2} + \theta )$ \\ 
		${P_4}$ & $y$-axis & $(-x,y)$ & $(r,\pi  - \theta )$ \\ 
		${P_5}$ & origin & $(-x,-y)$ & $(r,\pi  + \theta )$ \\
		${P_6}$ & origin, $y=x$ & $(-y,-x)$ & $(r,\frac{{3\pi }}{2} - \theta )$ \\ 
		${P_7}$ & $y=x$, $x$-axis & $(y,-x)$ & $(r,\frac{{3\pi }}{2} + \theta )$ \\
		${P_8}$ & $x$-axis & $(x,-y)$ & $(r,2\pi  - \theta )$ \\
		\bottomrule
	\end{tabular}
\end{table}

\subsubsection{Local Calculation}
Local calculation is to calculate the image moments for a part of the image such as dense image blocks or interest regions of keypoints. In this application scenario, the resolutions of such image patches are generally small, which can be described well using few moments. As the main reason for pushing up the time complexity, the number of such patches is typically very large, e.g., this number for dense blocks can be of order $10^5$ or even $10^6$. Therefore, speed optimization methods for global computation often do not solve the local calculation problem well. At present, the problem of calculating local image moments has not been fully discussed. The most promising path is to find and compress redundant operations in the for-loop like processing of image patches. In this regard, \emph{useful properties of basis functions} (such as shift properties in [74, 75]) and \emph{special data structures} (such as complex-valued integral images in [76]) have been explored to improve efficiency.

\subsubsection{Recent Advance of Fast Calculation}

For the fast global calculation of Jacobi polynomial based orthogonal moments, state-of-the-art methods usually use recursion and symmetry in combination. A recent work in this way was proposed by Upneja et al [56] for JFM (see also [53, 59]), it requires 
$\mathcal{O}(\frac{1}{8}(K{N^2} + {K^2}{N^2}))$ additions and 
$\mathcal{O}(\frac{1}{8}{K^2}{N^2})$ multiplications for all the moments of orders in set $\{ (n,m) : |n|,|m| \le K\} $. Considering that multiplication is usually much more expensive than addition in modern computing systems, so the total elapsed time is mainly determined by the $\mathcal{O}(\frac{1}{8}{K^2}{N^2}) = \mathcal{O}({K^2}{N^2})$ multiplications. In this case, it is clear that when higher image resolution (i.e., $N$ increases) or more moments (i.e., $K$ increases) are required, the computational cost still increases in \emph{quadratic} manner. This is mainly due to the complicacy of the Jacobi polynomials, in other words, which leads to the lack of fast implementation with a complexity below the quadratic time.

For the fast global calculation of harmonic function based orthogonal moments, the state-of-the-art performance offered by the FFT based fast implementation. The earliest idea in this way can be traced back to 2014, when Ping et al. [70] introduced the FFT by the relationship between EFM and Fourier transform. Subsequently, similar ideas have been used for RHFM [68] and PCET [69]. An important work was recently proposed by Yang et al. [67], which generalized such FFT-based techniques to a generic version of harmonic function based orthogonal moments (will be seen in Section 4.4.3). This generic fast implementation, similar to its previous special forms, exhibits the multiplicative complexity of $\mathcal{O}({M^2}\log M) = \mathcal{O}({N^2}\log N)$, where $M$ is a sampling parameter and $M \propto N$. Note that, surprisingly, $K$ is \emph{no role} in the multiplicative complexity. This property means that when $K$ is slightly higher, so that  $K^2$ is greater than $\log N$, the elapsed time of this generic method will be significantly lower than the strategy based on recursion and symmetry such as [67]. In addition to the efficiency gain, this method has also been proven (both analytically and experimentally) to avoid numerical instability, while providing quite high calculation accuracy when $M$ is large.

For the fast local calculation, the recent work of Bera et al. [76] makes a crucial contribution. They reduced the ZM calculation of dense image blocks to \emph{constant} time, i.e., $\mathcal{O}(1)$ , by introducing an elegant data structure: complex-valued integral image. The further speed improvements can be achieved with the help of another structure: lookup table. While the use of integral image in fast implementation is not rare, e.g., SURF uses it to speed up SIFT [77], this is still the very first time such techniques have been used for orthogonal moments. Through the complicated derivation, we found that two mathematical tools played an important role in rewriting the definition of ZM towards integral images, namely the complex-plane representation of radial and angular coordinates and the binomial expansion. The above structures and tools introduced by Bera et al. may provide important insights to researchers in the field. Unfortunately, as discussed by the authors, this fast algorithm seems to be applicable only for ZM and not for other moments with better representation power (i.e., no information suppression) such as OFMM. If the similar constant-time implementation could be developed for all the unit disk-based orthogonal moments, this will greatly promote their application in real-time tasks.

\subsection{Robustness/Invariance Optimization}

As the mathematical background of Section 4.3, the general definitions of invariance, robustness, and discriminability in image representation are first given.

If there exists a function ${\mathcal{R}}$  such that the original image $f$ and its degraded version ${\mathcal{D}}(f)$, where ${\mathcal{D}}$  is the degradation operator, satisfy
\begin{equation}
	{\mathcal{R}}(f) \equiv {\mathcal{R}}({\mathcal{D}}(f)),
\end{equation}
for any $f$, the representation ${\mathcal{R}}$  is said to be \emph{invariant} to the degradation ${\mathcal{D}}$.

We consider function ${\mathcal{L}}:{\bf{X}} \times {\bf{X}} \to [0, + \infty )$  to be a \emph{distance} measure on a set ${\bf{X}}$, where for all ${\bf{x}},{\bf{y}},{\bf{z}} \in {\bf{X}}$, the following three axioms are satisfied:
\begin{itemize}
	\item \emph{identity of indiscernibles} – ${\mathcal{L}}({\bf{x}},{\bf{y}}) = 0 \Leftrightarrow {\bf{x}} = {\bf{y}}$;
	\item \emph{symmetry} – ${\mathcal{L}}({\bf{x}},{\bf{y}}) = {\mathcal{L}}({\bf{y}},{\bf{x}})$;
	\item \emph{subadditivity} or \emph{triangle inequality} – ${\mathcal{L}}({\bf{x}},{\bf{y}}) \le {\mathcal{L}}({\bf{x}},{\bf{z}}) + {\mathcal{L}}({\bf{z}},{\bf{y}})$;
\end{itemize}

Given a distance function ${\mathcal{L}}$, \emph{robustness} requires that the \emph{intra-class distance} of the representation ${\mathcal{R}}$:
\begin{equation}
	{\mathcal{L}}({\mathcal{R}}(f),{\mathcal{R}}({\mathcal{D}}(f))),
\end{equation}
should be sufficiently small. Conversely, assuming that image $g$ is semantically different from the image $f$, \emph{discriminability} requires that the \emph{inter-class distance} of the representation ${\mathcal{R}}$:
\begin{equation}
	{\mathcal{L}}({\mathcal{R}}(f),{\mathcal{R}}(g)),
\end{equation}
should be sufficiently large. It can be seen that the invariance implies the perfect robustness, i.e., ${\mathcal{L}}({\mathcal{R}}(f),{\mathcal{R}}({\mathcal{D}}(f))) = 0$  holds if and only if Equation (19) holds due to the first axiom of ${\mathcal{L}}$. Here, the moment-based image representation ${\mathcal{I}}$  is a special form of the generic representation ${\mathcal{R}}$, which depends on the given basis function ${V_{nm}}$ and can be written as
\begin{equation}
	{\mathcal{R}}(f) = {\mathcal{I}}(\{  < f,{V_{nm}} > \} ).
\end{equation}

It is almost impossible to devise a representation that maintains well invariance or robustness to all kinds of degradations. More precisely, the only possibility corresponds to a representation without any discriminability [8]. Thus, in practice, the design of representation  ${\mathcal{R}}$ generally relies on certain assumptions about the degradation ${\mathcal{D}}$. In general, invariance/robustness optimization methods are proposed against three types of attacks: \emph{affine transformation}, \emph{noise}, and \emph{blurring}.

\subsubsection{Affine Transformation}
It is a transformation of image-space coordinates, that is, the degradation operator ${\mathcal{D}}$ is a mapping from the pixel coordinates $(x,y)$ to the new coordinates $(x',y')$, ${\mathcal{D}}:(x,y) \to (x',y')$, is defined as:
\begin{equation}
	\left( {\begin{array}{*{20}{c}}
			{x'}\\
			{y'}\\
			1
	\end{array}} \right) = \left( {\begin{array}{*{20}{c}}
			{{a_{00}}}&{{a_{01}}}&{{t_x}}\\
			{{a_{10}}}&{{a_{11}}}&{{t_y}}\\
			0&0&1
	\end{array}} \right) \cdot \left( {\begin{array}{*{20}{c}}
			x\\
			y\\
			1
	\end{array}} \right),
\end{equation}
where the affine parameters $({a_{00}},{a_{01}},{a_{10}},{a_{11}},{t_x},{t_y})$ can encode rotation, scaling, translation, shearing, flipping, and all linear combinations of the above transformations. In the moment-based image representation, there are three methods to achieve invariance to the affine transformation  ${\mathcal{D}}$: \emph{normalization}, \emph{indirect}, and \emph{explicit} methods [8, 16, 78].
\begin{itemize}
	\item Normalization method [79, 80]:  In this approach, the degraded input image ${\mathcal{D}}(f)$ is converted to some \emph{reference} form ${f^{{\rm{ref}}}}$, so the representation of this normalized image,  ${\mathcal{R}}({f^{{\rm{ref}}}})$, will be invariant to the affine transformation ${\mathcal{D}}$. Obviously, the invariance of this approach comes from the independent correction technique. More specifically, the normalization can be considered as a function ${\mathcal{N}}$, which must hold ${\mathcal{N}}({\mathcal{D}}(f)) = {\mathcal{N}}(f) = {f^{{\rm{ref}}}}$ for any image $f$ and any admissible affine transformation ${\mathcal{D}}$. Note that the geometric transformation do not need to be actually implemented on the digital image, this normalization process is conceptual [8]. As can be seen, the normalization results will fundamentally affect the invariance of the representation. Over the years, the design of such a normalization approach has been pursued by many researchers in different tasks such as geometric correction [81, 82] and image registration [83, 84]. One of the simplest methods is based on geometric moments, which obtain translation, scale, and rotation invariance by evaluating/eliminating the \emph{centroid}, \emph{scaling factor}, and \emph{principal axis} of the image, all based on low-order geometric moments. For more details on this, we strongly encourage readers to see the books by Flusser et al. [8, 16], where they provide the complete analysis and definition. For other types of normalization methods, readers can refer to the paper by Zitova et al [84].
	\item Indirect method [85, 86]: It is well known that geometric moments are easy to derive explicit invariants for affine transformation ${\mathcal{D}}$, as show in [87, 8, 16]. Therefore, this approach express orthogonal moment invariants as a linear combination of the geometric moment invariants, based on the algebraic relation between the orthogonal moments and geometric ones. Obviously, the invariance of this approach comes from the geometric moment invariants. Here, the conversion relationship is universal, since polynomials are formed directly from linear combinations of sequences $\{ {x^0},{x^1},...,{x^n}\} $, and harmonic functions can be written in a similar form via the \emph{Taylor series}. Note that image normalization using geometric moments followed by orthogonal moments based description and the orthogonal moment invariants derived by a linear combination of geometric moment invariants play a very similar role in theory; in practice, different methods have different issues on numerical stability and computational complexity [8].
	\item Explicit method [88–90]: This approach seeks to derive invariants directly from the orthogonal moments. Mathematically, it tries to satisfy the identity ${\mathcal{I}}(\{ \left< f,{V_{nm}} \right> \} ) = {\mathcal{I}}(\{ \left< {\mathcal{D}}(f),{V_{nm}} \right>  \} )$ by designing ${V_{nm}}$ and ${\mathcal{I}}$. The invariance of this approach directly comes from the given moments. A very common method in this approach is to explicitly construct the rotation invariants of the circular moments. Duo to the angular basis function ${A_m}(\theta ) = \exp (\bm{j}m\theta )$ and \emph{Fourier Shift Theorem}, the circular moments defined in Equation (1) and Equation (4) of the rotated image ${f^{\rm{rot}}}(r,\theta ) = f(r,\theta  + \phi )$ are ${M'_{nm}} = \exp (\bm{j}m\phi ){M_{nm}}$. Hence, image rotation operation only affects the phase of circular moments. The traditional method of achieving phase cancellation is based on magnitude, i.e., $|{M'_{nm}}| = |\exp (\bm{j}m\phi ){M_{nm}}| = |{M_{nm}}|$. However, such simple methods discard too much image information: leaving only the magnitude, and the more important phase is ignored [91, 92]. For this, Flusser [93] proposed the complex-valued rotation invariants containing phase information. Let  $L \ge 1$ and ${n_i} \in \vmathbb{N}$, ${m_i, k_i} \in \vmathbb{Z}$, 
	$i = 1,...,L$, such that $\sum\nolimits_{i = 1}^L {{m_i}{k_i} = 0}$. Then, the complex-valued rotation invariants of circular moments can be defined as $inv = \prod\nolimits_{i = 1}^L {{{[{M_{{n_i}{m_i}}}]}^{{k_i}}}} $ [93]. In fact, the normalization method may be impractical for many applications, e.g. the dense image block representation. Moreover, the moments computed by normalization scheme may differ from the true moments of the standard image, owing to errors in the evaluation of the normalization parameters. As for the indirect method, a long time is allocated to compute the polynomial coefficients, hence requiring the recursive calculation of such coefficients. In contrast, due to its direct structure, the explicit method is more applicable in a variety of scenarios and can usually achieve higher accuracy and efficiency [62].
\end{itemize}

\subsubsection{Noise}
Different from the affine transformation, the noise attack acts on the intensity domain, that is, the degradation operation ${\mathcal{D}}$ is a mapping from the original intensity function $f(x,y)$ to the new intensity function $f'(x,y)$, ${\mathcal{D}}:f(x,y) \to f'(x,y)$. Mathematically, based on a specific noise function $\eta (x,y)$, the common additive noise is defined as:
\begin{equation}
	f'(x,y) = f(x,y) + \eta (x,y),
\end{equation}
and multiplicative noise is defined as:
\begin{equation}
	f'(x,y) = f(x,y) \times \eta (x,y).
\end{equation}

For enhancing the robustness to noise, it is difficult to adopt a similar normalization as in the affine invariant representation, because the inverse problems are often \emph{ill-conditioned} or \emph{ill-posed} [94, 95]. Currently, a common strategy in moment-based image representation is to convert the image to a new space with higher \emph{Signal-to-Noise Ratio} (SNR) via a specific transformation ${\mathcal{T}}$, e.g., \emph{Radon space} [96–98]. Based on this new space, ${V_{nm}}$ and ${\mathcal{I}}$ are designed for achieving a robust representation ${\mathcal{R}}$:
\begin{equation}
	{\mathcal{I}}(\{  \left< {\mathcal{T}}(f),{V_{nm}} \right> \} ) \simeq {\mathcal{I}}(\{  \left< {\mathcal{T}}({\mathcal{D}}(f)),{V_{nm}} \right> \} ).
\end{equation}

Image representation based on the Radon transform has the advantage of being robust to additive noise [99]. The Radon transform of an image function $f(x,y)$ is defined as:
\begin{equation}
	{{\mathop{\rm Rad}\nolimits} _f}(\rho ,\alpha ) = \int\limits_{ - \infty }^{ + \infty } {\int\limits_{ - \infty }^{ + \infty } {f(x,y){\delta _{\rho ,x\cos \alpha  + y\sin \alpha }}dxdy} },
\end{equation}
where ${\delta _{ij}}$ is the Kronecker delta function defined in Equation (3), $\rho  = x\cos \alpha  + y\sin \alpha $ is a straight line with the angle $\alpha$  (w.r.t. the $y$-axis) and the distance $\rho$ (w.r.t. the origin). The Radon transform of the noisy image defined in Equation (24) can be written as:
\begin{equation}
	\begin{split}
		{{\mathop{\rm Rad}\nolimits} _{f'}}(\rho ,\alpha )\; = \int\limits_{ - \infty }^{ + \infty } {\int\limits_{ - \infty }^{ + \infty } {[f(x,y) + \eta (x,y)]{\delta _{\rho ,x\cos \alpha  + y\sin \alpha }}dxdy} } 
		= {{\mathop{\rm Rad}\nolimits} _f}(\rho ,\alpha ) + {{\mathop{\rm Rad}\nolimits} _\eta }(\rho ,\alpha ).
	\end{split}
\end{equation}

In the continuous domain, the Radon transform of noise ${{\mathop{\rm Rad}\nolimits} _\eta }(\rho ,\alpha )$ is equal to the mean value of the noise, which is assumed to be zero. Thus, we have
\begin{equation}
	{{\mathop{\rm Rad}\nolimits} _{f'}}(\rho ,\alpha )\; = {{\mathop{\rm Rad}\nolimits} _f}(\rho ,\alpha ).
\end{equation}

This means that zero-mean additive noise has no effect on the Radon transform of the image. As for multiplicative noise as Equation (25), a potential strategy is to use logarithmic transformation as a pre-processing to separate the image and noise into two additive parts, i.e., $\log [f(x,y) \times \eta (x,y)] = \log [f(x,y)] + \log [\eta (x,y)]$. In practice, however, the perfect identity of Equation (29) does not hold because the image and noise are sampled and quantized. For the discrete case, the SNR of the image in Radon space is still significantly higher than that of the original image, meaning better robustness. Readers are referred to the papers of Hoang et al. [97] and Jafari-Khouzani et al. [99] for detailed theoretical analysis and experiments.

\subsubsection{Blurring}
Similar to the noise attack, the blurring also acts on the intensity domain, that is, ${\mathcal{D}}:f(x,y) \to f'(x,y)$. Mathematically, based on a specific \emph{Point-Spread Function} (PSF) $\mu (x,y)$, the observed blurred image can be described as a convolution:
\begin{equation}
	f'(x,y) = f(x,y) \otimes \mu (x,y).
\end{equation}

Therefore, blur invariance can also be called convolution invariance. For image blurring, “normalization” means in fact blind deconvolution, which is a strongly ill-conditioned or ill-posed problem [100, 101]. At present, the invariant representation of blurred images usually relies on specific image space transformation such as projection operation similar to Radon transform. The main idea can also be expressed by the Equation (26). In this regard, the works of Flusser et al. [102–105] are remarkable.

\subsubsection{Recent Advance of Robustness/Invariance Optimization}
The affine moment invariants of 2D and 3D image are the well-established tools [106–109]. In the books of Flusser et al. [8, 16], explicit affine invariants and normalization strategy using geometric/complex moments are derived in detail. Not only that, such invariants can easily be extended to orthogonal moments through the inherent conversion relationship. Hence, the work of Flusser et al. on affine invariants is very generic and remains one of the most landmark efforts in the field. More recently, the extension of invariants has attracted strong scientific interest [110]. 1) Extension to other \emph{transform groups}, such as the invariants under reflection [111], projective [112], Mobius [113], and chiral [114] transformations. 2) Extension to other \emph{data dimensions} or \emph{formats}: such as invariants of color image [115], curve [116], surface [117], vector field [118, 119], and tensor field [120–122]. 3) Extension to other \emph{moments}, such as the invariants of OFMM [123], JFM [124], Fourier-Mellin transform [125], radial Tchebichef moments [126], radial Legendre moments [127], and Gaussian-Hermite moments [128–131]. As a noteworthy work, Li et al. [132] deeply explored the structure of invariants. They proposed two fundamental \emph{Generating Functions} (GF), which can encode geometric moment invariants and further construct shape descriptors, just like DNA encodes proteins. They also found that Hu’s seven invariants can be further decomposed into a simpler set of \emph{Primitive Invariants} (PI), may meaning a new perspective of invariant study [133].

It is well known that the inverse problems of image noise and blurring are usually more difficult to solve than the inverse problems of affine transformations. This means that the normalization methods, i.e. denoising and deblurring, are very challenging tasks. Such approach is, generally, slow, and unstable due to the restoration artifacts [102]. For the brute-force path such as CNN, the lack of inherent invariance causes them to be very sensitive to noise and blurring operations not seen in the training. A recent work [134] confirmed this conclusion through extensive experiments. Most common strategy to alleviate this problem is data augmentation. However, it is very time and memory consuming, just “learning by rote”. Due to above facts, the moment-based invariant representation of noisy/blurred images is crucial to many practical applications. In this path, a main goal pursued by the researchers is to simplify the assumptions about degradation, which are the basis for the design of invariants. For example, by the additional constraints on PSF, they derived invariants to motion blur [135], axially symmetric blur in case of two axes [136], circularly symmetric blur [137, 138], arbitrary $N$-fold symmetric blur [103], circularly symmetric Gaussian blur [104], and general (anisotropic) Gaussian blur [102]. Recently, a promising invariant representation to blurring was proposed by Kostková et al [102]. The main contribution is the design of the invariants to general Gaussian blur, where the PSF is a Gaussian function with unknown parameters, i.e., the blur kernel may be arbitrary oriented, scaled, and elongated.

\subsection{Definition Extension}

Starting from different application scenarios and optimization goals, mathematical extension on the definitions of classical moment basis functions is also a popular research topic. In the following, we will introduce some common paths of definition extension.

\subsubsection{Quaternion}
Mathematically, the gray-level image function $f(x,y)$ can be defined as a mapping from 2D image plane to 1D intensity value, i.e., $f: \vmathbb{R}^2 \to \vmathbb{R}$; the color image function $\bm{f}(x,y) = \{ {f_R}(x,y),{f_G}(x,y),{f_B}(x,y)\} $  can be defined as a mapping from 2D image plane to 3D intensity value, i.e., $\bm{f}: \vmathbb{R}^2 \to \vmathbb{R}^3$. Here, ${f_R}(x,y)$,  ${f_G}(x,y)$, and  ${f_B}(x,y)$ are the Red, Green, and Blue components of color image in RGB model, respectively. Note that other three-channel color models are also applicable to the analysis in this section.

In Sections 2 and 3, the listed classical moments was directly designed for gray-level images but not for color images. When dealing with color image, there are two straightforward strategies [139, 140]: 1) graying method, calculating the moments of the gray-level version of color image; 2) channel-wise method, direct calculating the moments of each color channel. However, the graying method may lose some significant color information. In addition, the channel-wise method can hardly produce the most compact representation of a color image and ignores the correlation between different color channels.

For solving above problems, a common strategy is to define the color image $\bm{f}(x,y)$ as a mapping from 2D image plane to \emph{quaternion} intensity value ${f_R}(x,y)\bm{i} + {f_G}(x,y)\bm{j} + {f_B}(x,y)\bm{k}$, i.e., $\bm{f}: \vmathbb{R}^2 \to \vmathbb{H}$. Correspondingly, the basis function ${V_{nm}}$ is also extended from the real domain $\vmathbb{R}$ or the complex domain $\vmathbb{C}$ to the quaternion domain $\vmathbb{H}$, i.e., new basis functions ${\bm{V}_{nm}} \in \vmathbb{H}$, for achieving a well counterpart $ \left< \bm{f},{\bm{V}_{nm}} \right> $ of the original inner product $ \left< f,{V_{nm}} \right> $ [67, 141–142]. For more details on quaternion algebra and quaternion moments, we encourage readers to see the papers by Chen et al. [141, 142].

\subsubsection{High-dimensional}
Quaternion extension acts on the range of images and basis functions, while high-dimensional extension acts on their domain. In Sections 2 and 3, the listed classical moments was directly designed for 2D images. When dealing with image defined in high-dimensional space $\vmathbb{R}^d$ (mainly 3D image, $d=3$), the domain $D$ of the basis functions ${V_{nm}}$ should also be extended from $\vmathbb{R}^2$  to  $\vmathbb{R}^d$ [16].

For Cartesian moments, such extension is quite straightforward. Based on a 1D orthogonality polynomials set, we can generate $d$-dimensional orthogonal basis functions by using the same polynomials set for the each directions/variables [145].

For circular moments, extension to high-dimensional space is more difficult. In 3D case, a common choice is replacing the angular basis function ${A_m}(\theta ) = \exp (\bm{j}m\theta )$ with the \emph{spherical harmonic} ${Y_{ml}}(\theta ,\varphi )$ of degree $l \in \vmathbb{N}$ and order $m \in \vmathbb{Z}$ [146, 147]:
\begin{equation}
	{Y_{ml}}(\theta ,\varphi ) = \sqrt {\frac{{(2l + 1)(l - m)!}}{{4\pi (l + m)!}}} {L_{ml}}(\cos \theta )\exp (\bm{j}m\varphi ),
\end{equation}
where $\theta  \in [0,\pi )$ and $\varphi  \in [0,2\pi )$ are inclination and azimuth of the spherical coordinate system respectively, $|m| \le l$, and ${L_{ml}}$ represents the associated Legendre functions, which can be written explicitly as:
\begin{equation}
	{L_{ml}}(x) = {( - 1)^m} {2^l} {(1 - {x^2})^{\frac{m}{2}}} \sum\limits_{k = m}^l {\frac{{k!{x^{k - m}}}}{{(k - m)!}}} \binom{l}{k} \binom{\frac{l + k + 1}{2}}{l}.
\end{equation}

The spherical harmonic ${Y_{ml}}(\theta ,\varphi )$ satisfies the orthogonality condition:
\begin{equation}
	\begin{split}
		\left< {Y_{ml}},{Y_{m'l'}} \right> = \int\limits_0^{2\pi } {\int\limits_0^\pi  {{Y_{ml}}(\theta ,\varphi )} } Y_{m'l'}^*(\theta ,\varphi )\sin \theta d\theta d\varphi 
		= {\delta _{mm'}}{\delta _{ll'}}.
	\end{split}
\end{equation}

Therefore, the 3D orthogonal basis functions in spherical coordinates can be constructed by combining the spherical harmonic ${Y_{ml}}(\theta ,\varphi )$  and radial basis functions ${R_n}(r)$ with some slight modifications [148].

\subsubsection{Fractional-order}
In Sections 2 and 3, the listed classical moments was designed rely on integer-order domain $(n,m) \in \vmathbb{Z}^2$. Recently, an interesting idea on extension the order domain of the classical moments has emerged [149–151]. It introduces a fractional-order parameter $\alpha  \in \vmathbb{R}$ through certain suitable variable substitution, e.g., $r: = {r^\alpha } \in [0,1]$ (circular moments) or $x: = {x^\alpha },y: = {y^\alpha } \in [0,1]$ (Cartesian moments). By the substitution, order of the newly defined moments can be extended to real domain, e.g., $\alpha n  \in \vmathbb{R}$ (circular moments) or $(\alpha n,\alpha m) \in \vmathbb{R}^2$ (Cartesian moments).

It is worth noting that such kind of fractional-order moments is not only the mathematical extension of classical moments, but also has a distinctive \emph{time-frequency analysis} capability [152]. Specifically, the fractional-order moments are able to control the zero distributions of the basis functions by changing value of the fractional-order parameter. According to the research on information suppression problem (mentioned in Section 3.4), the distribution of zeros of the basis functions is a very important property because it is closely related to the description emphasis of the moments in the spatial-domain. As a result, the computed fractional-order moments are able to put emphasis on certain regions of an image, which are useful for solving information suppression issues and extracting image local features [149–151].

\begin{figure}[!t]
	\centerline{
		\includegraphics[scale=0.36]{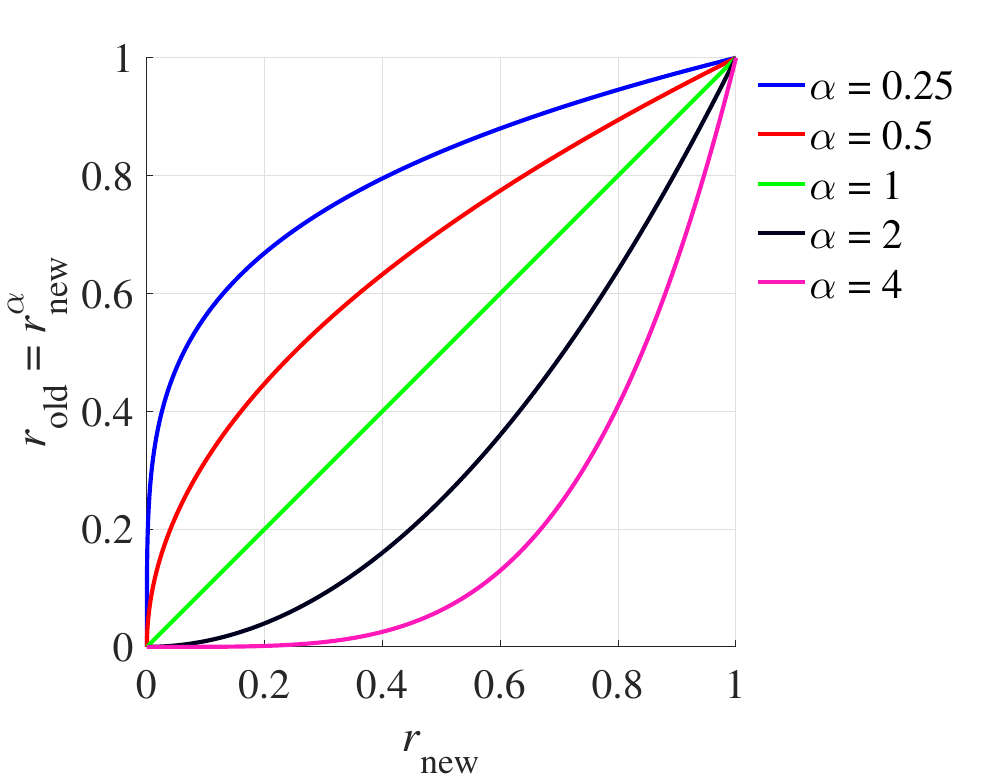}}
	\caption{Illustration of the variable substitution in fractional-order extension.}
\end{figure}

We will explain why fractional-order extension brings the time-frequency analysis capability. Taking the fractional-order circular moments as an example, where a new variable ${r_{{\rm{new}}}} \in [0,1]$ is used in the definition with ${r_{{\rm{old}}}} = {r_{{\rm{new}}}}^\alpha  \in [0,1]$  and $\alpha  \in \vmathbb{R}^+$. As illustrated in Figure 8, we can derive the following conclusions on the distribution of zeros and the description emphasis of extracted moments [152]:
\begin{itemize}
	\item When $\alpha  = 1$, the zeros of radial basis functions and description emphasis are the same as the corresponding integer-order version due to ${r_{{\rm{old}}}} = {r_{{\rm{new}}}}^\alpha  = {r_{{\rm{new}}}}$;
	\item When $\alpha <  1$, the zeros of radial basis functions are biased towards 0 due to ${r_{{\rm{old}}}} = {r_{{\rm{new}}}}^\alpha  > {r_{{\rm{new}}}}$, meaning more emphasis on the inner region of the image;
	\item When $\alpha >  1$, the zeros of radial basis functions are biased towards 1 due to ${r_{{\rm{old}}}} = {r_{{\rm{new}}}}^\alpha  < {r_{{\rm{new}}}}$, meaning more emphasis on the outer region of the image.
\end{itemize}

\subsubsection{Other}
In addition to the above common paths, several other extension strategies can be found in the literature.

Zhu [153] uses \emph{multivariate} orthogonal polynomials [154], which are the tensor product of two different orthogonal polynomials, for the definition of Cartesian moments. In contrast, the classical strategy is to use the tensor product of two same orthogonal polynomials. This new definition is more flexible and may have better performance in certain situations [155–158].

Wang et al. [159] proposed \emph{semi-orthogonal} moments to adjust the frequency-domain nature and spatial-domain description emphasis of the basis functions. It adopts a certain modulation function (may with parameter [160, 161]) to weight the basis functions. Such semi-orthogonal moments are reported to be powerful for time-frequency analysis, but orthogonality condition of their basis functions is not hold.

Zhu et al. [162, 163] adopted some \emph{generalized} polynomials to extend the definition of existing classical moments. The corresponding recursive calculation formulas are also explicitly given in the papers. Note that a newly introduced parameter in these methods exhibit a similar property to the fractional-order parameter (mentioned in Section 4.4.3), i.e., the ability to adjust the distribution of zeros. One possible implication is that there may be a mathematical connection between the two. Some similar papers on definition extension are [164–167].

\subsubsection{Recent Advance of Definition Extension}
Quaternion extension has become a very common strategy in moment-based color image processing [139, 168–172]. In addition, quaternion theory is increasingly used in related fields, including both hand-crafted [173–175] and learning-based representation [176–179]. We list below some recent advances in quaternion moments. Chen et al. [180] adopted quaternion algebra for representing RGB-D (RGB and depth) images, where the real part of quaternion number encodes the depth component. Yamni et al. [181] proposed a new category of moments for color stereo image representation, called octonion moments. Such kind of moments is a natural generalization of quaternion moments, rely on octonion algebra. We believe that the further research in this field will be inspired by the advance of hyper-complex algebra such as [182].

High-dimensional extension of image moments appeared very early [148] and attracted significant attention in the last decade [145, 158, 185–189]. A possible explanation for its popularity might be the rapid development of devices and technologies related to 3D images such as medical imaging [183] and computer graphics [184]. Recent work in this field mainly focuses on the new definitions [145, 158], invariants [188, 189], and accurate/fast calculations [185–187] of 3D image moments. Note that most of the main ideas for accurate/fast calculations of 2D image moments (mentioned in Sections 4.1 and 4.2) can be natural generalized to the 3D case. As for the 3D moment invariants, some promising works have been analyzed in Section 4.3.

Most recently, more attention of researchers has been drawn to the fractional-order extension, compared to the above two paths [190–192]. In this regard, the earliest work we find was presented by Hoang et al. [149, 150]. They extend the existing harmonic function based orthogonal moments (mentioned in Section 3.2) to a generic version, called Generic Polar Harmonic Transforms (GPHT), by fractional-order extension. A key contribution is that the time-frequency nature of GPHT and its potential in image representation were first discovered and analyzed. Xiao et al. [151] defined a general framework for the fractional-order extension of Jacobi polynomial based orthogonal moments in both circular and Cartesian case (mentioned in Sections 2 and 3.1). Most of the subsequent related papers [193–200] follow Hoang et al. [149, 150] and Xiao et al. [151], where main ideas of [198–200] are actually special cases of [149, 150]. Note that all the above related work focuses on solving the information suppression problem or extracting local image information. In a recent paper, Yang et al. [152] define a new set of generic fractional-order orthogonal moments, called Fractional-order Jacobi-Fourier Moments (FJFM), taking Jacobi polynomial-based classical and fractional-order orthogonal moments as special cases. More importantly, they found that all above methods have a common defect in the image global representation. Such global representation using a specific fractional parameter (determined experimentally) only brings a slight performance improvement compared to the classic moments, mainly due to the contradiction between the robustness and discriminability. Starting from this, Yang et al. [152] take a first step to improve the performance of global representation using the time-frequency property of fractional-order orthogonal moments.

\subsection{Application}

In addition to the theoretical works described above, the application research of orthogonal moments is also active. Such applications cover many familiar problems of image processing, computer vision, and information security, as well as some interdisciplinary frontiers. In this section, considering the focus of this survey, we aim to draw a high-level intuition for the applications of image moments without involving technical details.

\begin{itemize}
\item \emph{Image processing} – For the low-level vision tasks, image moments are popular choices. The application involves different contents of image processing: low-level feature detection and representation (e.g., detection of edge [201] and keypoint [202], representation of interest region [203], texture [55], and optical flow [204]), degraded image restoration and representation (e.g., denoising [205], deblurring [206], and superresolution [207]), image registration [208], image compression, coding and communication [209], and image quality assessment [210]. Similarly, moments have also been employed for the low-level vision tasks in video analysis [211] and computer graphics [212].
\item \emph{Visual understanding} – For the high-level vision tasks, image moments have been explored in different applications of computer vision. The related works are found in image classification [213], instance retrieval [214], object detection [215], and semantic segmentation [216]. Another major class of applications is pattern recognition, including behavior recognition [217], text recognition [218], biometric recognition [219], and sentiment recognition [220].
\item \emph{Information security} – Moments and moment invariants have shown significant impact on the research of information security, especially for the visual media. The popularity should be attributed to the robustness of moment-based representation, which is consistent with the \textit{two-player} nature of security research. Successful applications mainly comprise digital watermarking [221], steganography [222], perceptual hashing [223], and passive media forensics (e.g., copy-move [224] and splicing [180] detection).
\item \emph{Interdisciplinary} – Since visual information processing and understanding are common tasks in many disciplines, the moment-based representation can naturally be extended to these fields. Typical interdisciplinary applications cover medicine (e.g., medical imaging [225]), geography (e.g., remote sensing [226]), robotics (e.g., visual servoing [227]), physics (e.g., optics [228] and fluid mechanics [118]), chemistry (e.g., analytical chemistry [229]), biology (e.g., protein structure representation [230]), and materials science (e.g., atomic environment representation [231]).
\end{itemize}

From the above review, two observations should be mentioned: 1) many works in Sections 4.1, 4.2, 4.3, and 4.4 are generic in their nature and are found in above application areas; 2) moment-based representation plays a key role in a variety of scenarios that require high efficiency or strong robustness.

\section{Software Package and Experiment Results}
In this section, we will give an open-source software for a variety of widely-used orthogonal moments. Some accurate/fast calculation, robustness/invariance optimization, and definition extension strategies are also included in the package. This software is thus called \texttt{MomentToolbox}, which is available at \texttt{https://github.com/ShurenQi/MomentToolbox}.

On this unified base, we will evaluate the accuracy/complexity, representation capability, and robustness/invariance of these methods through moment calculation, image reconstruction, and pattern recognition experiments, respectively. It should be highlighted that all experiments are performed under Microsoft Windows environment on a PC with 2.90 GHz CPU and 8 GB RAM, and all the presented algorithms are implemented in Matlab R2021a.

\subsection{Moment Calculation: Accuracy and Complexity}
The accuracy and complexity are evaluated by using different methods to calculate the moments of a $128 \times 128$ image with a unity gray-level, i.e., $\{ {f_{{\rm{uni}}}}(x,y) = 1:(x,y) \in D\} $.

Going back to Equation (1) and Equation (4), we can easily derive the unit disk-based orthogonal moments of this image are
\begin{equation}
	\begin{split}
		\left< {f_{{\rm{uni}}}},{V_{nm}} \right> = \iint\limits_D {V_{nm}^*(x,y){f_{{\rm{uni}}}}(x,y)dxdy}
		= \left\{ {\begin{array}{*{20}{c}}
				0&{m \ne 0}\\
				{2\pi \int\limits_0^1 {R_n^*(r)rdr} }&{m = 0}
		\end{array}} \right..
	\end{split}
\end{equation}

The Equation (34) means that for any $(n,m)$ in $\{ (n,m):(n,m) \in \vmathbb{Z}^2,m \ne 0\} $, the theoretical value of the corresponding moment $\left< {f_{{\rm{uni}}}},{V_{nm}} \right>$ should be 0. But in practice, such identity usually does not hold due to certain error caused by the imperfection of the moment calculation method. For further analysis, please refer to [232, 233]. In fact, the errors listed in Section 4.1, i.e., geometric error, numerical integration error, and representation error, may all cause this phenomenon.

\begin{table*}[!t]
	\caption{Definition and Cardinality of Order Set ${\bf{S}}(K)$ for Different Orthogonal Moment Methods}
	\centering
	\begin{tabular}{ccc}
		\toprule
		Method & ${\bf{S}}(K)$ & $|{\bf{S}}(K)|$ \\ \midrule
		ZM [29] & $n - |m| = {\rm{even,}}\;|m| \le n \le K$ & $(K + 1)(K + 2)/2$ \\
		PZM [20] & $|m| \le n \le K$ & ${(K + 1)^2}$ \\
		\tabincell{c}{OFMM [33], CHFM [34], PJFM [35], JFM [36],\\ RHFM [38], PCT [40], BFM [41],\\ FJFM [152], GRHFM [150], GPCT [150]}  & $0 \le n,\;|m| \le K$ & $(K + 1)(2K + 1)$ \\
		EFM [39], PCET [40], GPCET [150] & $|n|,\;|m| \le K$ & ${(2K + 1)^2}$ \\
		PST [40], GPST [150] & $1 \le n \le K,\;|m| \le K$ & $K(2K + 1)$ \\ \bottomrule
	\end{tabular}
\end{table*}

Based on the above facts, the computational accuracy is evaluated by a newly defined simple measure, called Average Calculation Error (ACE):
\begin{equation}
	{\rm{ACE}} = \frac{{\sum\limits_{(n,m) \in \{ {\bf{S}}(K),\;m \ne 0\} } {|M_{nm}^{({f_{{\rm{uni}}}})}|} }}{{|{\bf{S}}(K)|}}
\end{equation}
where $M_{nm}^{({f_{{\rm{uni}}}})}$ is the unit disk-based orthogonal moment of image ${f_{{\rm{uni}}}}$ calculated by a specific method, and ${\bf{S}}(K)$ is the set of selected order $(n,m)$ based on a integer constant $K$. Here, the definition and cardinality of order set ${\bf{S}}(K)$  for a variety of unit disk-based orthogonal moments are listed in Table 4.

The computational complexity is evaluated by the elapsed time over all feasible orders in order set ${\bf{S}}(K)$, called Decomposition Time (DT):
\begin{equation}
	{\rm{DT}} = \sum\limits_{(n,m) \in {\bf{S}}(K)} {{\rm{time}}(M_{nm}^{({f_{{\rm{uni}}}})})}
\end{equation} 
where ${{\rm{time}}(M_{nm}^{({f_{{\rm{uni}}}})})}$ is the elapsed time for a moment $M_{nm}^{({f_{{\rm{uni}}}})}$. To make a fair comparison, all the methods operate in single-thread modality, using \texttt{-singleCompThread} option in Matlab.

The comparison methods include:
\begin{itemize}
	\item	Direct computation of classical Jacobi polynomial-based methods (ZM [29], PZM [20], OFMM [33], CHFM [34], PJFM [35], and JFM [36]);
	\item	Direct computation of classical harmonic function-based methods (RHFM [38], EFM [39], PCET [40], PCT [40], and PST [40]);
	\item	Direct computation of classical eigenfunction-based methods (BFM [41]);
	\item	Direct computation of fractional-order Jacobi polynomial-based methods (FJFM [152]);
	\item	Direct computation of fractional-order harmonic function-based methods (GRHFM [150], GPCET [150], GPCT [150], and GPST [150]);
	\item	Recursive computation of fractional-order Jacobi polynomial-based methods (FJFM [152]);
	\item	FFT-based computation of fractional-order harmonic function-based methods (GPCET [67]).
\end{itemize}

Here, JFM has parameters $(p,q)$, FJFM has parameters $(p,q,\alpha )$, and GRHFM/GPCET/GPCT/GPST has parameter $\alpha$. Note that FJFM and GPCET are actually the generic versions of the existing Jacobi polynomial-based moments [20, 29, 33–36, 151, 193–196] and Harmonic function-based moments [38–40, 150, 198–200, 234], respectively. In other words, the accurate/fast calculation algorithm for FJFM or GPCET can be directly used in its special case, by properly setting the values of the parameters. For further analysis, please refer to [152, 67].

\begin{figure*}[!t]
	\centering
	\subfigure[]{\includegraphics[scale=0.45]{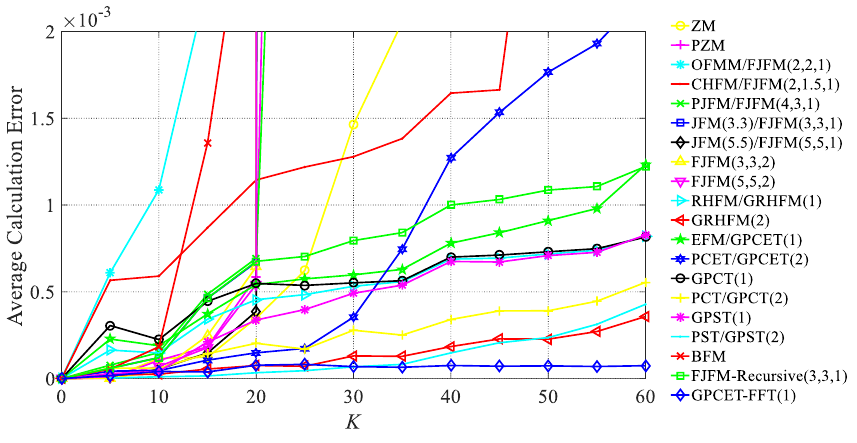}}
	\subfigure[]{\includegraphics[scale=0.45]{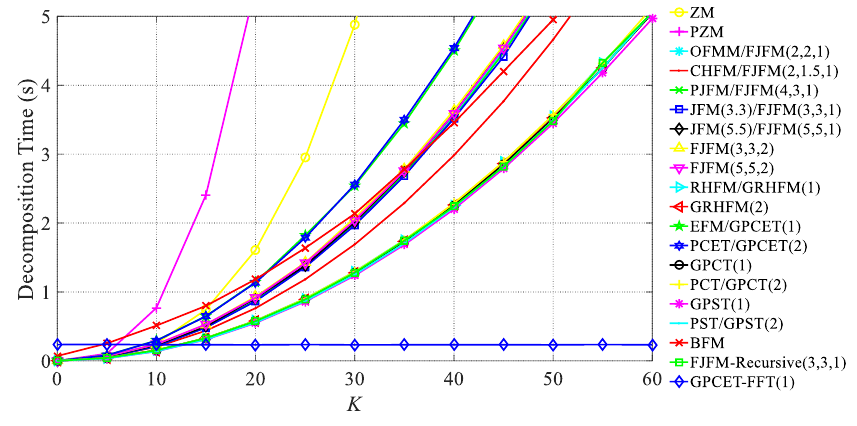}}
	\caption{Average calculation error (a) and decomposition time (b) for different orthogonal moment methods.}
\end{figure*}

Figure 9 provides the ACE and DT of all above comparison methods. It is observed from the figure that

\begin{itemize}
	\item Accuracy: 1) The error curves of Jacobi polynomial-based moments exhibit sudden upturns when $K$ is increased up to a certain point. Obviously, there exists a main problem of numerical instability (i.e., representation error) due to the factorial/gamma terms in the Jacobi polynomials. As can be expected, the calculation based on recursion has significantly better accuracy, no numerical instability is observed in Figure 9, because it does not involve the factorial/gamma of large numbers. 2) The main problem, in harmonic function-based moments, is numerical integration error. According to the sampling theorem, such error increases as $K$ increases, which is consistent with the phenomenon we observed in Figure 9. Note that the radial basis functions of some kinds of harmonic function-based moments are unbounded, e.g., RHFM and EFM, meaning potential representation errors. This is the main reason why their calculation accuracy is worse than similar methods. Thanks to the pseudo up-sampling and the polar domain definition, the FFT-based calculation has almost constant integration error and avoids numerical instability caused by unboundedness. 3) As for eigenfunction-based moments, although we use a fast and stable Matlab built-in function \texttt{besselj} to implement the radial basis functions, its error curves still rise sharply. In fact the error is between Jacobi polynomial-based moments and harmonic function-based moments for larger $K$ (no numerical instability), which is not shown in Figure 9.

	\item Complexity: 1) In general, the complexity of the Jacobi polynomial-based moments is relatively high, especially when $K$ is large, due to the complicated factorial/gamma terms and long summations in radial basis functions. With the recursive strategy, this part of the computation can be greatly reduced, exhibiting a complexity similar to that of harmonic function-based moments. 2) The calculation cost of the harmonic function-based moments mainly comes from the inner product due to the simplicity of the radial basis functions, which is less than the Jacobi polynomial-based moments. Note that, for the same $K$, the number of moments of EFM and PCET is twice that of other harmonic function-based moments (see also Table 4). Thus, the calculation time of EFM and PCET is higher than the similar methods, about twice, as expected. If the FFT-based calculation is used, the computation of the inner product will be greatly reduced, and its complexity is independent of $K$. 3) In Figure 9, eigenfunction-based moments show similar complexity to Jacobi polynomial-based moments. However, if the calculation is done directly from the definition without using the fast \texttt{besselj}, its complexity will be significantly higher due to gamma terms, infinite series, and root-finding operations.
\end{itemize}

The above experimental evidence supports our theoretical analysis and observations in Sections 3, 4.1 and 4.2.

\textbf{Remark:} In the direct calculation scenario, harmonic function-based moments have better computational complexity and accuracy, compared with Jacobi polynomial-based moments and eigenfunction-based moments. The easy-to-implement recursive strategy can effectively overcome the numerical stability issue of Jacobi polynomial-based moments, also with better efficiency, it should be promoted in real-world applications. As a promising research path, the harmonic function-based moments calculated by FFT show superior performance in terms of both complexity and accuracy.

\begin{figure*}[!t]
	\centering
	\subfigure[ZM]{
		\includegraphics[scale=0.14]{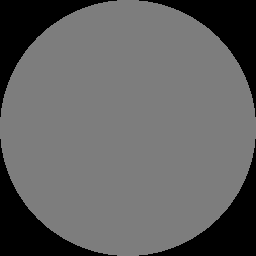}
		\includegraphics[scale=0.14]{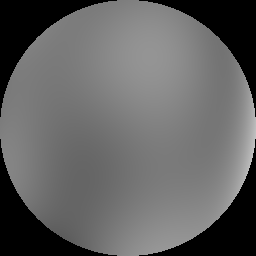}
		\includegraphics[scale=0.14]{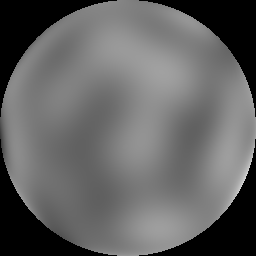}
		\includegraphics[scale=0.14]{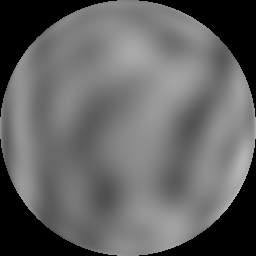}
		\includegraphics[scale=0.14]{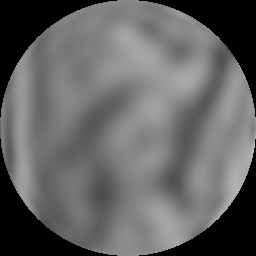}
	}
	\subfigure[JFM(3.3)/FJFM(3,3,1)]{
		\includegraphics[scale=0.14]{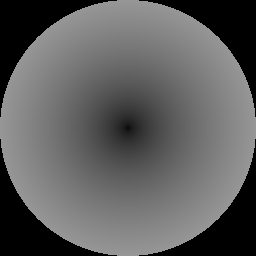}
		\includegraphics[scale=0.14]{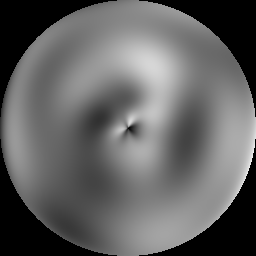}
		\includegraphics[scale=0.14]{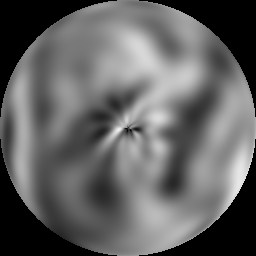}
		\includegraphics[scale=0.14]{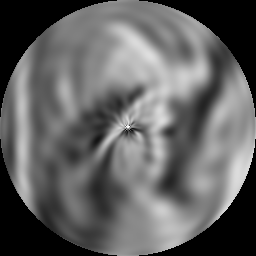}
		\includegraphics[scale=0.14]{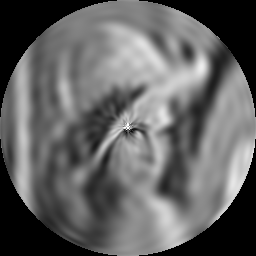}
	}\\
	\subfigure[FJFM(3,3,2)]{
		\includegraphics[scale=0.14]{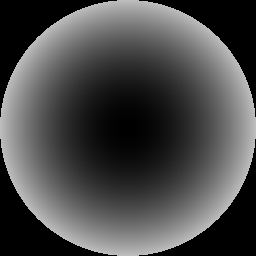}
		\includegraphics[scale=0.14]{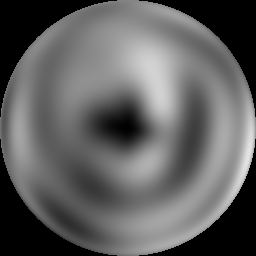}
		\includegraphics[scale=0.14]{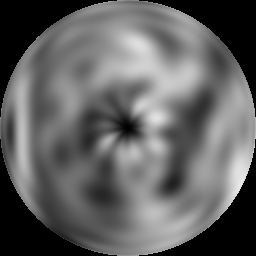}
		\includegraphics[scale=0.14]{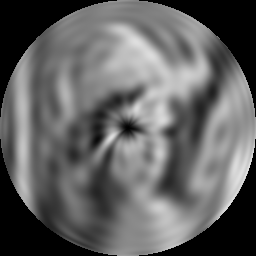}
		\includegraphics[scale=0.14]{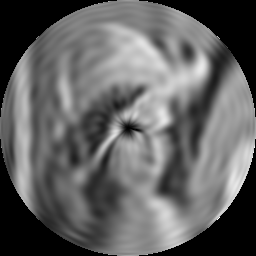}
	}
	\subfigure[RHFM/GRHFM(1)]{
		\includegraphics[scale=0.14]{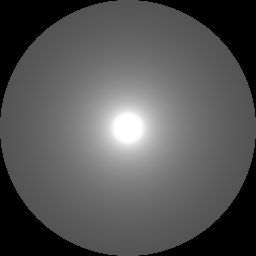}
		\includegraphics[scale=0.14]{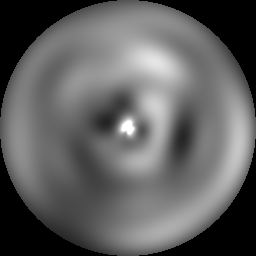}
		\includegraphics[scale=0.14]{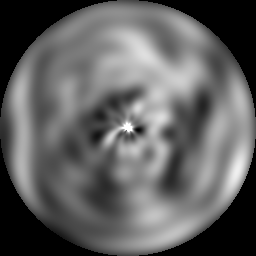}
		\includegraphics[scale=0.14]{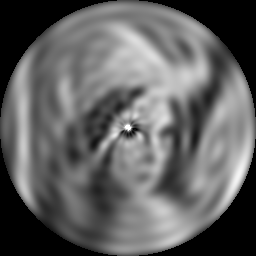}
		\includegraphics[scale=0.14]{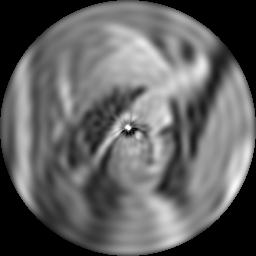}
	}\\
	\subfigure[GRHFM(2)]{
		\includegraphics[scale=0.14]{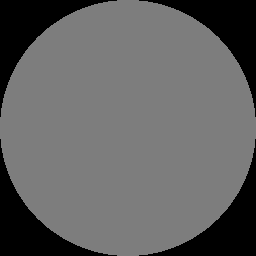}
		\includegraphics[scale=0.14]{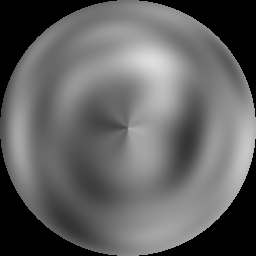}
		\includegraphics[scale=0.14]{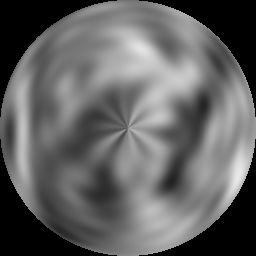}
		\includegraphics[scale=0.14]{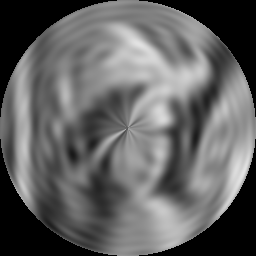}
		\includegraphics[scale=0.14]{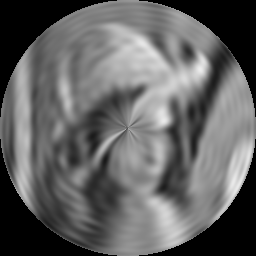}
	}
	\subfigure[EFM/GPCET(1)]{
		\includegraphics[scale=0.14]{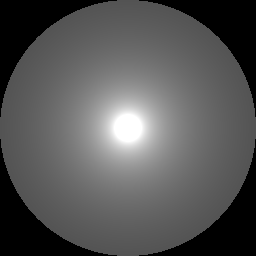}
		\includegraphics[scale=0.14]{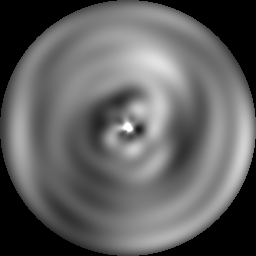}
		\includegraphics[scale=0.14]{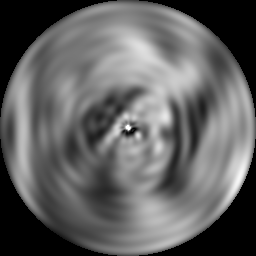}
		\includegraphics[scale=0.14]{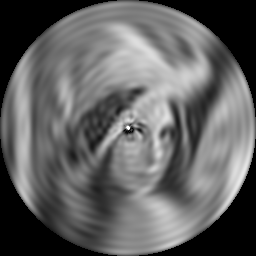}
		\includegraphics[scale=0.14]{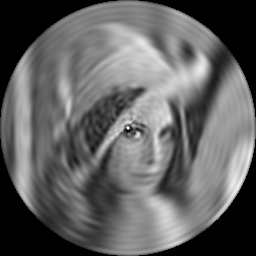}
	}\\
	\subfigure[PCET/GPCET(2)]{
		\includegraphics[scale=0.14]{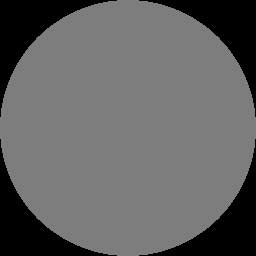}
		\includegraphics[scale=0.14]{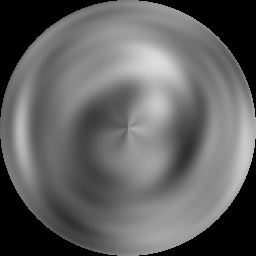}
		\includegraphics[scale=0.14]{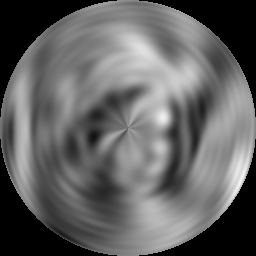}
		\includegraphics[scale=0.14]{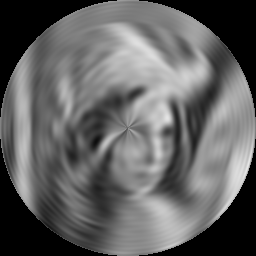}
		\includegraphics[scale=0.14]{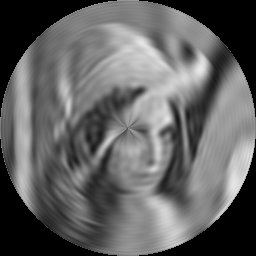}
	}
	\subfigure[BFM]{
		\includegraphics[scale=0.14]{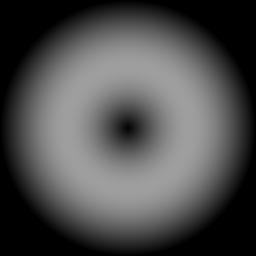}
		\includegraphics[scale=0.14]{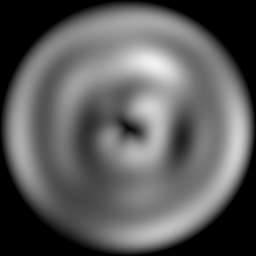}
		\includegraphics[scale=0.14]{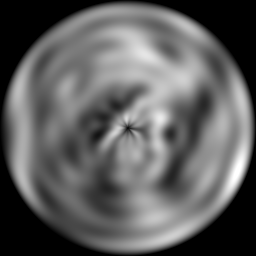}
		\includegraphics[scale=0.14]{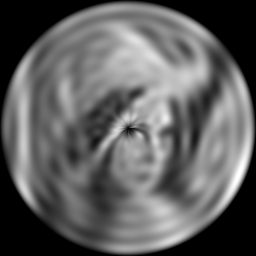}
		\includegraphics[scale=0.14]{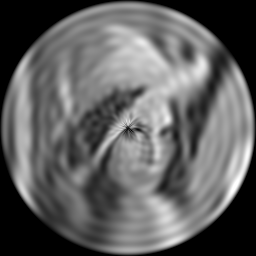}
	}\\
	\centering
	\caption{Some samples of the reconstructed images by different orthogonal moment methods with $K = \{ 0,5,...,20\}$ (from left to right).}
\end{figure*}

\subsection{Image Reconstruction: Representation Capability}

The representation capability is evaluated by using comparison methods to reconstruct the gray image “Lena” with size $256 \times 256$.

Going back to Equation (1) and Equation (2). Thanks to the orthogonality of ${V_{nm}}$, the image reconstruction from orthogonal moments $\{ {M_{nm}} =  \left< f,{V_{nm}} \right> \}$ can be performed easily as:
\begin{equation}
	\hat f(x,y) = \sum\limits_{(n,m) \in {\bf{S}}(K)} {{M_{nm}}{V_{nm}}} (x,y)
\end{equation} 
where $\hat f$ is the reconstructed version of original image $f$, and ${\bf{S}}(K) \in \vmathbb{Z}^2$ is defined in Table 4. The reconstruction error is evaluated by the Mean-Square Reconstruction Error (MSRE) [20]:
\begin{equation}
	{\rm{MSRE}} = \frac{{\int\limits_{ - \infty }^{ + \infty } {\int\limits_{ - \infty }^{ + \infty } {{{[f(x,y) - \hat f(x,y)]}^2}dxdy} } }}{{\int\limits_{ - \infty }^{ + \infty } {\int\limits_{ - \infty }^{ + \infty } {{{[f(x,y)]}^2}dxdy} } }},
\end{equation}
and the well-known Structural SIMilarity (SSIM) [235] is used as a fidelity criteria:
\begin{equation}
	{\rm{SSIM}} = \frac{{(2{\mu _f}{\mu _{\hat f}} + {C_1})(2{\sigma _{f,\hat f}} + {C_2})}}{{(\mu _f^2 + \mu _{\hat f}^2 + {C_1})(\sigma _f^2 + \sigma _{\hat f}^2 + {C_2})}},
\end{equation}
where ${\mu _ \bullet }$, ${\sigma _ \bullet }$, and ${\sigma _{ \bullet , \bullet }}$ represent the mean value, standard deviation, and covariance, respectively; constants ${C_1}$ and ${C_2}$ are set to ${C_1} = {(0.01 \times 255)^2}$ and ${C_2} = {(0.03 \times 255)^2}$.

The comparison methods include:
\begin{itemize}
	\item Classical Jacobi polynomial-based methods (ZM [29], PZM [20], OFMM [33], CHFM [34], PJFM [35], and JFM [36]);
	\item Classical harmonic function-based methods (RHFM [38], EFM [39], and PCET [40]);
	\item Classical eigenfunction-based methods (BFM [41]);
	\item Fractional-order Jacobi polynomial-based methods (FJFM [152]);
	\item Fractional-order harmonic function-based methods (GPCET [150]).
\end{itemize}

For better reflect the true capability in representing image functions, more accurate calculation based on recursion or FFT is adopted instead of direct method. In other words, the influence of calculation errors in the corresponding reconstruction results will be suppressed. Specifically, ZM and PZM are implemented by $q$-recursive method [43] and $p$-recursive method [73], respectively; the generic recursive calculation in [152] and the generic FFT-based calculation in [67] are used for other Jacobi polynomial-based moments and harmonic function-based moments, respectively. As for BFM, the evaluation of radial basis functions is facilitated by stable Matlab built-in function \texttt{besselj}.

\begin{figure*}[!t]
	\centering
	\subfigure[]{\includegraphics[scale=0.45]{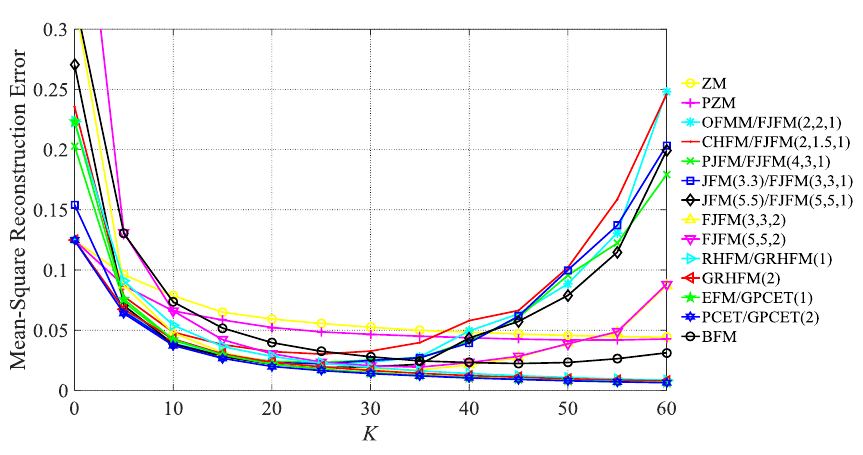}}
	\subfigure[]{\includegraphics[scale=0.45]{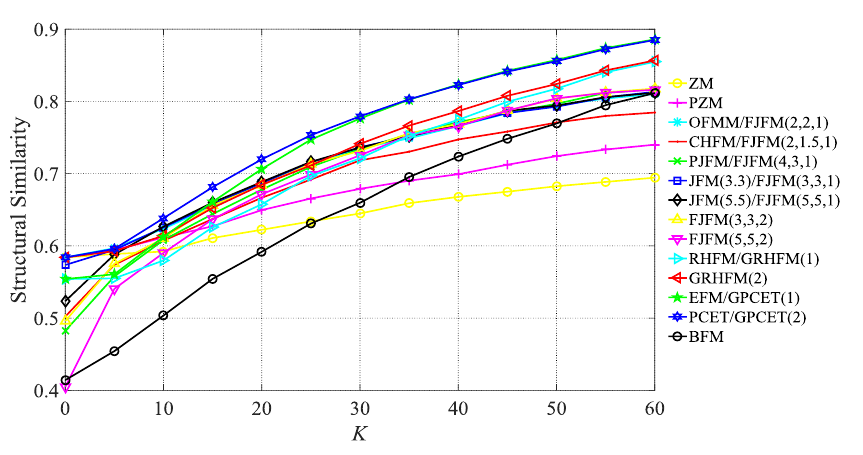}}
	\caption{Mean-square reconstruction error (a) and structural similarity (b) for different orthogonal moment methods.}
\end{figure*}

Figure 10 shows some samples of the reconstructed images for $K = \{ 0,5,...,20\}$, while the MSRE and SSIM values of all above comparison methods are given in Figure 11. It is observed from these results that
\begin{itemize}
	\item It is intuitive that more moments can better reconstruct/represent the image function. From the visual result (Figure 10) and MSRE/SSIM value (Figure 11) of reconstructed image, in general, the larger the $K$ value the better the reconstruction quality. As some exceptions, the MSRE curves of some methods (e.g., OFMM) start to rise when $K$ is increased up to a certain point, which is mainly due to the representation error at $r \simeq 1$ (see also Figure 3).
	\item Another important observation is that, as shown in Figure 10, the low-order moment mainly reflects the overall grayscale/intensity information of image, while the high-order moment reflects the more complex texture/shape information. On the other hand, from a robustness perspective, we can infer that the higher-order moments (corresponding to the high-frequency components of image) are less robust to noise, blur, and sampling/quantization effect.
	\item It should be noted that the reconstruction/representation quality is closely related to the number of zeros of basis functions. In the cases of Jacobi polynomial-based moments, the reconstructed images by ZM and PZM exhibit worse MSRE/SSIM values compared to similar methods due to their fewer zeros of basis functions (see also Table 2). The same reason can explain why the MSRE/SSIM values of RHFM and GRHFM are worse than other harmonic function-based moments (see also Table 2).
	\item As for the fractional-order moments (i.e., FJFM, GRHFM, and GPCET), the reconstruction quality is relatively better at $\alpha  \simeq 1$, which should be attributed to the uniform distributions of radial kernel’s zeros. Furthermore, they are hard to reconstruct the image’s inner/outer part when $\alpha  > 1$ / $\alpha  < 1$; on the contrary, the inner/outer part is able to be reconstructed with fewer moments when $\alpha  < 1$ / $\alpha  > 1$. For the sake of brevity, only the case of $\alpha  > 1$ is shown in Figure 10, readers can use \texttt{MomentToolbox} to verify other cases.
\end{itemize}

The above experimental evidence supports our theoretical analysis and observations in Sections 3 and 4.4.

\textbf{Remark:}  In the family of classical Jacobi polynomial-based methods, except for ZM and PZM, the reconstruction/representation qualities of all other methods are similar, due to their similar mathematical properties. In fact, ZM and PZM require higher-order moments to obtain a close performance to other methods, but possibly at the cost of reduced robustness to signal corruptions. In contrast, the harmonic function-based moments, especially EFM, PCET, and GPCET, are generally better than other types of methods in the clean image representation. It is worth noting that the time-frequency analysis capability of fractional-order moments may useful for certain scenarios such as solving information suppression issues and extracting image local features.

\subsection{Pattern Recognition: Robustness and Invariance}

The robustness and invariance are evaluated by using comparison methods to recognize the degraded images with different levels of affine transformations and signal corruptions.

\begin{figure}[!t]
	\centering
	\subfigure{\includegraphics[scale=0.3]{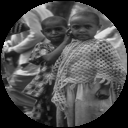}}
	\subfigure{\includegraphics[scale=0.3]{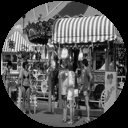}}
	\subfigure{\includegraphics[scale=0.3]{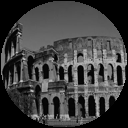}}
	\subfigure{\includegraphics[scale=0.3]{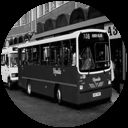}}
	\subfigure{\includegraphics[scale=0.3]{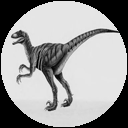}}
	\\
	\subfigure{\includegraphics[scale=0.3]{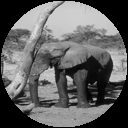}}
	\subfigure{\includegraphics[scale=0.3]{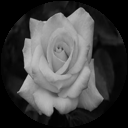}}
	\subfigure{\includegraphics[scale=0.3]{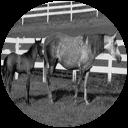}}
	\subfigure{\includegraphics[scale=0.3]{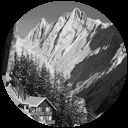}}
	\subfigure{\includegraphics[scale=0.3]{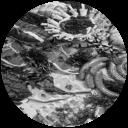}}
	\caption{Some samples of the training images.}
\end{figure}

\begin{figure}[!t]
	\centering
	\subfigure{\includegraphics[scale=0.3]{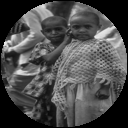}}
	\subfigure{\includegraphics[scale=0.3]{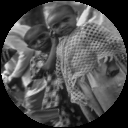}}
	\subfigure{\includegraphics[scale=0.3]{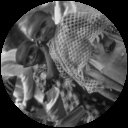}}
	\subfigure{\includegraphics[scale=0.3]{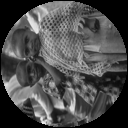}}
	\subfigure{\includegraphics[scale=0.3]{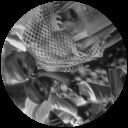}}
	\\
	\subfigure{\includegraphics[scale=0.3]{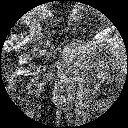}}
	\subfigure{\includegraphics[scale=0.3]{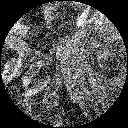}}
	\subfigure{\includegraphics[scale=0.3]{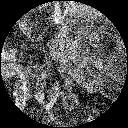}}
	\subfigure{\includegraphics[scale=0.3]{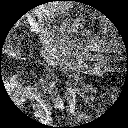}}
	\subfigure{\includegraphics[scale=0.3]{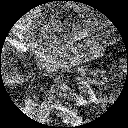}}
	\\
	\subfigure{\includegraphics[scale=0.3]{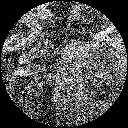}}
	\subfigure{\includegraphics[scale=0.3]{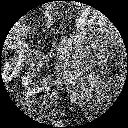}}
	\subfigure{\includegraphics[scale=0.3]{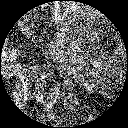}}
	\subfigure{\includegraphics[scale=0.3]{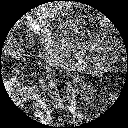}}
	\subfigure{\includegraphics[scale=0.3]{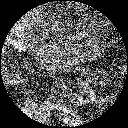}}
	\caption{Some samples of the testing images with rotation angles $\{ 0^\circ , 30^\circ , ... , 120^\circ \} $ (from left to right) and noise variances $\{ 0, 0.05, 0.1\} $ (from top to bottom).}
\end{figure}

To generate the training set, we selected 100 images from the COREL dataset [236] and resized them to $128 \times 128$. The calculated feature vectors of these training images are considered to be the ground-truth for comparing with the testing images. Figure 12 shows several images in the training set. To generate the testing set, each image from the training set is rotated at angles $\{ 0^\circ, 10^\circ, ... , 350^\circ \} $, reflecting the arbitrariness of the orientation. Also, the scale and position are random perturbed by pixels $\{10, 12, ... , 18 \} $ and $\{0, 1, ... , 4 \} $, respectively, aiming to simulate the imperfect segmentation of interest region in the visual system. Then, each image is destroyed by white Gaussian noise with variances $\{ 0, 0.05, ..., 0.3\} $. Thus, the testing images are with both affine transformations and signal corruptions. Through the above operations, a training image corresponds to $36 \times 7 = 252$ testing images, resulting in a total of $252 \times 100 = 25200$ images in the testing set. For better visualization, Figure 13 shows some selected images from the testing set.

For achieving classification, we use the moment-based feature vector to describe all images in the training and testing sets, which is defined as:
\begin{equation}
	{\bf{v}}(f) = \{ | \left< f,{V_{nm}} \right> |:(n,m) \in {\bf{S}}(K)\} ,
\end{equation}
where ${\bf{S}}(K) \in \vmathbb{Z}^2$ is defined in Table 4. Going back to Equation (4), mathematically, it is easy to verify that Equation (40) must satisfy the rotation invariance when ${V_{nm}}$ is of the form ${R_n}(r)\exp (\bm{j}m\theta )$ (see also Section 4.3.1). In addition, the most naive minimum distance classifier (based on Euclidean distance) is used to better reflect the robustness and invariance of the representation itself. The classification accuracy is evaluated by the Correct Classification Percentages (CCP) [41]:
\begin{equation}
	{\rm{CCP}} = \frac{{\# \;{\rm{correctly \; classified \; samples}}}}{{\# \;{\rm{total \; samples \; in \; the \; dataset}}}}.
\end{equation}

The comparison methods include:
\begin{itemize}
	\item Classical Jacobi polynomial-based methods (ZM [29], PZM [20], OFMM [33], CHFM [34], PJFM [35], and JFM [36]);
	\item Classical harmonic function-based methods (RHFM [38], EFM [39], PCET [40], PCT [40], and PST [40]);
	\item Classical eigenfunction-based methods (BFM [41]);
	\item Fractional-order Jacobi polynomial-based methods (FJFM [152]);
	\item Fractional-order harmonic function-based methods (GRHFM [150], GPCET [150], GPCT [150], and GPST [150]);
	\item Two lightweight learning methods, PCANet [4] and Compact Binary Face Descriptor (CBFD) [237], are inspired by the dimensionality reduction-based descriptor PCA and texture-based descriptor LBP, respectively;
	\item Two deep learning methods, GoogeLeNet [238] and ResNet-50 [239], are well known in numerous vision applications.
	
\end{itemize}

\begin{table*}[!t]
	\centering
	\caption{Correct Classification Percentages (\%) for Different Orthogonal Moment Methods}
	\begin{tabular}{ccccccccccccccccc}
		\toprule
		\multirow{2}[4]{*}{Method} & \multicolumn{1}{c}{\multirow{2}[4]{*}{\textit{K}}} & \multicolumn{7}{c}{Gaussian white noise (variance)} \\
		\cmidrule{3-9}    \multicolumn{1}{c}{} &       & 0     & 0.05  & 0.1   & 0.15  & 0.2   & 0.25  & 0.3 \\
		\midrule
		ZM    & 10    & \cellcolor[rgb]{ .984,  .894,  .835} 89.03 & \cellcolor[rgb]{ .984,  .894,  .835} 68.78 & \cellcolor[rgb]{ .984,  .894,  .835} 42 & \cellcolor[rgb]{ .984,  .894,  .835} 26.67 & \cellcolor[rgb]{ .984,  .894,  .835} 17.47 & \cellcolor[rgb]{ .984,  .894,  .835} 12.31 & \cellcolor[rgb]{ .984,  .894,  .835} 9.06 \\
		PZM   & 10    & \cellcolor[rgb]{ .984,  .894,  .835} 92.17 & \cellcolor[rgb]{ .984,  .894,  .835} 76.33 & \cellcolor[rgb]{ .984,  .894,  .835} 45.36 & \cellcolor[rgb]{ .984,  .894,  .835} 30.44 & \cellcolor[rgb]{ .984,  .894,  .835} 18.14 & \cellcolor[rgb]{ .984,  .894,  .835} 12.94 & 9.47 \\
		OFMM/FJFM(2,2,1) & 10    & 97.25 & 90    & 72.42 & 47.11 & 26.53 & 15.36 & \cellcolor[rgb]{ .984,  .894,  .835} 8.78 \\
		CHFM/FJFM(2,1.5,1) & 10    & 98.56 & 93.86 & 78.03 & \cellcolor[rgb]{ .871,  .918,  .965} 60.94 & \cellcolor[rgb]{ .871,  .918,  .965} \textbf{40.22} & \cellcolor[rgb]{ .871,  .918,  .965} \textbf{25.42} & \cellcolor[rgb]{ .871,  .918,  .965} 16.5 \\
		PJFM/FJFM(4,3,1) & 10    & 97.94 & 92.64 & 79.08 & 58.28 & 37.5  & 22.67 & 13.81 \\
		JFM(3.3)/FJFM(3,3,1) & 10    & 97.92 & 93.81 & 77.19 & 55.94 & 35.44 & 22.39 & 13.89 \\
		JFM(5.5)/FJFM(5,5,1) & 10    & 97.75 & 93.53 & 78.86 & 59.67 & 38.56 & \cellcolor[rgb]{ .871,  .918,  .965} 24.92 & \cellcolor[rgb]{ .871,  .918,  .965} \textbf{16.86} \\
		FJFM(3,3,2) & 10    & 97.36 & 93.22 & 76.92 & 56.83 & 38.11 & 24.25 & \cellcolor[rgb]{ .871,  .918,  .965} 16.5 \\
		FJFM(5,5,2) & 10    & \cellcolor[rgb]{ .984,  .894,  .835} 96.58 & 91.47 & 75.64 & 56.22 & 37.53 & 22.44 & 15.22 \\
		RHFM/GRHFM(1) & 10    & 98.17 & 94.22 & 77.81 & 57.72 & 35.33 & 21.94 & 13.61 \\
		GRHFM(2) & 10    & 97.42 & 89.03 & \cellcolor[rgb]{ .984,  .894,  .835} 68.17 & \cellcolor[rgb]{ .984,  .894,  .835} 43.17 & 25.78 & \cellcolor[rgb]{ .984,  .894,  .835} 14.78 & 9.72 \\
		EFM/GPCET(1) & 10    & \cellcolor[rgb]{ .871,  .918,  .965} \textbf{99.56} & \cellcolor[rgb]{ .871,  .918,  .965} \textbf{97.06} & \cellcolor[rgb]{ .871,  .918,  .965} \textbf{86.47} & \cellcolor[rgb]{ .871,  .918,  .965} \textbf{61.78} & 37.33 & 21.94 & 13.44 \\
		PCET/GPCET(2) & 10    & \cellcolor[rgb]{ .871,  .918,  .965} 98.75 & 92.97 & 79    & 52.53 & 29.67 & 16.92 & 10.5 \\
		GPCT(1) & 10    & 98.64 & \cellcolor[rgb]{ .871,  .918,  .965} 95.25 & \cellcolor[rgb]{ .871,  .918,  .965} 80.39 & 60.22 & \cellcolor[rgb]{ .871,  .918,  .965} 39.33 & \cellcolor[rgb]{ .871,  .918,  .965} 24.75 & \cellcolor[rgb]{ .871,  .918,  .965} 16.53 \\
		PCT/GPCT(2) & 10    & 97.17 & \cellcolor[rgb]{ .984,  .894,  .835} 88.44 & 70.14 & 43.81 & \cellcolor[rgb]{ .984,  .894,  .835} 25.08 & 15.44 & 9.22 \\
		GPST(1) & 10    & \cellcolor[rgb]{ .871,  .918,  .965} 98.89 & \cellcolor[rgb]{ .871,  .918,  .965} 95.72 & \cellcolor[rgb]{ .871,  .918,  .965} 79.61 & \cellcolor[rgb]{ .871,  .918,  .965} 60.5 & \cellcolor[rgb]{ .871,  .918,  .965} 38.72 & 23.53 & 13.75 \\
		PST/GPST(2) & 10    & 96.94 & 89.69 & 70.47 & 46.5  & 25.56 & 15.11 & \cellcolor[rgb]{ .984,  .894,  .835} 9.06 \\
		BFM   & 10    & 98.39 & 94.28 & 79.33 & 60.06 & 37.47 & 21.53 & 13.28 \\
		\midrule
		\multirow{2}[4]{*}{Method} & \multirow{2}[4]{*}{\textit{K}} & \multicolumn{7}{c}{Gaussian white noise (variance)} \\
		\cmidrule{3-9}          &       & 0     & 0.05  & 0.1   & 0.15  & 0.2   & 0.25  & 0.3 \\
		\midrule
		ZM    & 20    & \cellcolor[rgb]{ .984,  .894,  .835} 89.97 & \cellcolor[rgb]{ .984,  .894,  .835} 70.86 & \cellcolor[rgb]{ .984,  .894,  .835} 43.03 & \cellcolor[rgb]{ .984,  .894,  .835} 27.36 & \cellcolor[rgb]{ .984,  .894,  .835} 16.5 & \cellcolor[rgb]{ .984,  .894,  .835} 12.28 & \cellcolor[rgb]{ .984,  .894,  .835} 9.19 \\
		PZM   & 20    & \cellcolor[rgb]{ .984,  .894,  .835} 92.67 & \cellcolor[rgb]{ .984,  .894,  .835} 76.53 & \cellcolor[rgb]{ .984,  .894,  .835} 46.36 & \cellcolor[rgb]{ .984,  .894,  .835} 30.58 & \cellcolor[rgb]{ .984,  .894,  .835} 18.44 & \cellcolor[rgb]{ .984,  .894,  .835} 12.92 & \cellcolor[rgb]{ .984,  .894,  .835} 9.03 \\
		OFMM/FJFM(2,2,1) & 20    & 97.78 & 90.78 & 74.08 & 49.47 & 29.97 & \cellcolor[rgb]{ .984,  .894,  .835} 16.61 & \cellcolor[rgb]{ .984,  .894,  .835} 10 \\
		CHFM/FJFM(2,1.5,1) & 20    & 98.5  & 94.36 & 78.83 & 61.39 & 41.36 & 26.53 & 17.61 \\
		PJFM/FJFM(4,3,1) & 20    & 98.17 & 93.75 & 80.28 & 60.97 & 40.78 & 23.53 & 15.69 \\
		JFM(3.3)/FJFM(3,3,1) & 20    & 98.19 & 94.22 & 78.44 & 57.22 & 37.53 & 22.69 & 14.42 \\
		JFM(5.5)/FJFM(5,5,1) & 20    & 97.94 & 94.11 & 79.83 & 60.94 & 41.31 & \cellcolor[rgb]{ .871,  .918,  .965} 26.69 & \cellcolor[rgb]{ .871,  .918,  .965} \textbf{18.11} \\
		FJFM(3,3,2) & 20    & 97.67 & 93.83 & 78.78 & 59.44 & 41.06 & 26.47 & 17.58 \\
		FJFM(5,5,2) & 20    & \cellcolor[rgb]{ .984,  .894,  .835} 97.08 & 92.89 & 77.33 & 60.86 & 40.94 & \cellcolor[rgb]{ .871,  .918,  .965} \textbf{26.92} & \cellcolor[rgb]{ .871,  .918,  .965} 17.89 \\
		RHFM/GRHFM(1) & 20    & 98.44 & 94.97 & 79.78 & 60.67 & 39.08 & 24.67 & 14.86 \\
		GRHFM(2) & 20    & 97.81 & 90.78 & \cellcolor[rgb]{ .984,  .894,  .835} 70.06 & \cellcolor[rgb]{ .984,  .894,  .835} 45.89 & 27.64 & 16.78 & 10.39 \\
		EFM/GPCET(1) & 20    & \cellcolor[rgb]{ .871,  .918,  .965} \textbf{99.47} & \cellcolor[rgb]{ .871,  .918,  .965} \textbf{97.86} & \cellcolor[rgb]{ .871,  .918,  .965} \textbf{87.97} & \cellcolor[rgb]{ .871,  .918,  .965} \textbf{66.83} & \cellcolor[rgb]{ .871,  .918,  .965} \textbf{42.47} & 25.56 & 16.33 \\
		PCET/GPCET(2) & 20    & \cellcolor[rgb]{ .871,  .918,  .965} 99.11 & 94.56 & 80.42 & 55.42 & 33.19 & 21.61 & 14.42 \\
		GPCT(1) & 20    & 98.5  & \cellcolor[rgb]{ .871,  .918,  .965} 95.17 & \cellcolor[rgb]{ .871,  .918,  .965} 81.83 & \cellcolor[rgb]{ .871,  .918,  .965} 62 & \cellcolor[rgb]{ .871,  .918,  .965} 42.11 & \cellcolor[rgb]{ .871,  .918,  .965} 26.64 & \cellcolor[rgb]{ .871,  .918,  .965} 17.94 \\
		PCT/GPCT(2) & 20    & 97.53 & \cellcolor[rgb]{ .984,  .894,  .835} 89.72 & 70.89 & 46.06 & \cellcolor[rgb]{ .984,  .894,  .835} 27.36 & 16.75 & 10.31 \\
		GPST(1) & 20    & \cellcolor[rgb]{ .871,  .918,  .965} 98.89 & \cellcolor[rgb]{ .871,  .918,  .965} 96.25 & 81    & \cellcolor[rgb]{ .871,  .918,  .965} 62.14 & \cellcolor[rgb]{ .871,  .918,  .965} 41.58 & 24.67 & 15.42 \\
		PST/GPST(2) & 20    & 97.28 & 91.17 & 73.11 & 50.36 & 28.81 & 17.92 & 10.53 \\
		BFM   & 20    & 98.5  & 94.69 & \cellcolor[rgb]{ .871,  .918,  .965} 81.08 & 61.44 & 40.75 & 23.75 & 14.75 \\
		\bottomrule
\end{tabular}%
\end{table*}%

Note that, to further prove the effectiveness of orthogonal moments in image representation, we will compare them with learning-based methods. For PCANet and CBFD, we use the common parameter settings in the original papers. The resolution of input image in PCANet is reduced to $64 \times 64$ to avoid excessive memory usage. The GoogeLeNet and ResNet-50 are retrained to fit the image sets in this paper, based on a transfer learning strategy. For more details, please refer to the code of \texttt{MomentToolbox} and original papers [4, 237–239].

\begin{table*}[!t]
	\centering
	\caption{Correct Classification Percentages (\%) for Learning Methods and Orthogonal Moment Methods}
	\begin{tabular}{cccccccc}
		\toprule
		\multirow{2}[4]{*}{Method} & \multicolumn{7}{c}{Gaussian white noise (variance)} \\
		\cmidrule{2-8}    \multicolumn{1}{c}{} & 0     & 0.05  & 0.1   & 0.15  & 0.2   & 0.25  & 0.3 \\
		\midrule
		PCANet (without rotation versions) & 19.28 & 7.58  & 4.64  & 3.03  & 2.08  & 2.03  & 1.56 \\
		PCANet (with rotation versions) & \textbf{99.97} & 14.25 & 4.33  & 2.19  & 1.69  & 1.25  & 1.19 \\
		CBFD (without rotation versions) & 22.58 & 3.72  & 2.08  & 1.22  & 1.19  & 1.14  & 1.08 \\
		CBFD (with rotation versions) & 98.92 & 3.89  & 2.5   & 1.75  & 1.53  & 1.64  & 1.11 \\
		GoogLeNet (with rotation versions) & 96.17 & 6.28  & 2.5   & 1.39  & 1.22  & 0.89  & 1.11 \\
		ResNet-50 (with rotation versions) & 99.33 & 3.03  & 1.61  & 1     & 1.17  & 0.78  & 1.03 \\
		Orthogonal moments (average results) & 97.42 & \textbf{91.47} & \textbf{74.62} & \textbf{54.39} & \textbf{35.05} & \textbf{21.83} & \textbf{14.14} \\
		\bottomrule
	\end{tabular}
\end{table*}%

The CCP values of above orthogonal moment methods for $K = \{ 10,20\}$ are presented in Table 5. For better visualization, at each noise variance, relatively high/low CCP values are indicated by blue/red shading, and the highest CCP value is bolded to highlight. In addition, Table 6 shows the comparison with learning-based methods, where the average values of the CCP for $K = 20$ in Table 5 are used to represent the performance of orthogonal moments. From these presented results, it can be observed that:
\begin{itemize}
	\item All the orthogonal moment methods listed in Table 5 have achieved high classification accuracy on clean images, due to their theoretical invariance for rotation and certain robustness to segmentation error. It is intuitive that the classification accuracy is negatively correlated with the noise variance and positively correlated with $K$. Therefore, the degree to which the CCP value is affected by the variance or $K$ reflects the robustness to noise.
	\item We observe that the classification performance is closely related to the number/distribution of zeros of basis functions. As shown in Table 5, ZM and PZM have lower performance than that of other methods. The major explanation is their fewer and biased zeros. Also, in general, the peak performance of fractional-order moments (i.e., GRHFM, GPCET, GPCT, GPST, and FJFM) is at $\alpha  \simeq 1$ due to their uniform zeros.
	\item On the whole, at the corresponding values of variance and $K$, EFM/GPCET(1), GPCT(1), GPST(1), CHFM/FJFM(2,1.5,1), and JFM(5.5)/FJFM(5,5,1) have higher classification accuracy, especially EFM/GPCET(1). This means that they have better representation capability and robustness in the global description scenario.
	\item In Table 6, PCANet and CBFD are first trained without rotation versions, i.e., directly using the same training set as orthogonal moment methods. Due to the lack of rotation invariance, the performance of PCANet and CBFD in this scenario is poor, even for the clean images.
	\item In the learning-based methods, the common idea to deal with geometric transformation is data augmentation, which means including the corresponding transformed versions in the training set. Accordingly, PCANet and CBFD are then trained with rotation versions. Note that this increased the training time by nearly 36 times. As shown in Table 6, although the PCANet and CBFD achieve performance improvements (especially on clean images), the accuracy decreases sharply as the noise variance increases and does not differ much from the training without the rotation versions.
	\item Due to the inherent nature, GoogeLeNet and ResNet-50 are difficult to train with very few samples, i.e., using the same set as orthogonal moment methods. Therefore, the rotation versions are used in the training, with a considerable time/space cost. From the results, one can see that the similar phenomena also occur in GoogeLeNet and ResNet-50. For the clean testing images, the classification accuracy is satisfactory, mainly due to the fact that such images have similar settings to the training data. However, the models are still sensitive to noise, the accuracy drops sharply on noisy images.
	\item Such common phenomena prove that orthogonal moment methods exhibit certain advantages for recognizing the image variants under geometric transformations and signal corruptions, compared to the listed learning methods. The invariance and independence of the moment-based descriptor draw a distinction between other methods, leading to potential benefits on small-scale robust recognition problems.
\end{itemize}

The above experimental evidence supports our theoretical analysis and observations in Sections 1, 3, 4.3, and 4.4.

\textbf{Remark:} It is worth noting that the time-frequency nature of the fractional-order orthogonal moments, in addition to the application on solving information suppression issues and extracting image local features, also has the potential to improve the global image representation. For this, readers can refer to the preliminary work in [152]. Due to space limitations and implementation difficulties, the experiment did not consider other more complex geometric transformations and the corresponding invariant generation strategies. In such a pattern recognition problem that requires very strong robustness, the performance gap between orthogonal moments and learning methods may be even greater.

\section{Concluding Remarks and Future Directions}
Robust and discriminative image representation is a long-lasting battle in the computer vision and pattern recognition. In this paper, we have presented a comprehensive survey on the orthogonal moment methods in image representation.

Starting from the review on the basic theories and classical methods in the area, we abstracted several basic conditions that an efficient moment-based representation should satisfy: the invariance/robustness to affine transformations and signal corruptions, the discriminability to a large number of patterns, and the reasonable computational complexity and accuracy. Based on these observations, this paper aimed to analyze the motivation and successful experiences behind the recent advances in fast/accurate calculation, robustness/invariance optimization, definition extension, and application. Note that such overall theoretical analysis of the state-of-the-art research progress is mostly ignored in previous studies.

In addition to the above theoretical contributions, we also provided extensive open-source implementations and experimental evaluations at the practical level. For the first time, a software package called \texttt{MomentToolbox} is available for the image representation community, covering classical methods and improvement strategies in the field of image moments. With this software, this paper has evaluated the accuracy/complexity, representation capability, and robustness/invariance of the widely-used methods through moment calculation, image reconstruction, and pattern recognition experiments, respectively. To the best of our knowledge, such overall performance statistics of the state-of-the-art methods have not been given until this work.

As can be seen from this survey, over a period of nearly six decades, the widespread studies on this field have resulted in a great amount of achievements. Despite its long history, moment-based representation appears to be still in development. This is not so surprising, considering the fundamental role of image representation, i.e., the performance of computer vision and pattern recognition methods is heavily dependent on the choice of data representation. With this premise, it is important trying to identify the most promising areas for future research.
\begin{itemize}
	\item \emph{Moment invariant theory in bag-of-visual-words model}. As the most competitive representation model in the pre-CNN era, the Bag-of-Visual-Words (BoVW) model is a hand-crafted algorithm in which local features are extracted, encoded, and summarized into global image representation. One of the main difficulties faced by the BoVW model is the unsatisfactory robustness of the representation [240, 241]. Obviously, the core of the improvement is to expand the invariance of local features (including the descriptor and detector). Therefore, the moment-based local descriptor/detector with good invariance to geometric transformations and signal corruptions is promising in alleviating this problem. Specifically, we noted a number of potential efforts [76, 92, 202, 203, 242–247].
	\item \emph{Moment invariant theory in deep-learning model}. As one of the most important representation method in deep learning, CNN serves as a hierarchical model for large-scale visual tasks. The large number of neurons contained in the network allows CNN to fit any complicated data distribution, meaning a strong discriminability. For this reason, CNN has received widespread attention recently. However, its problems are also commonly reported such as high time/space complexity and difficulty in achieving satisfactory robustness [134, 240]. In this respect, the mathematical constraints of invariance and independence behind the moment-based image representation are useful for solving these problems. We believe that such exploration of introducing knowledge into data-driven algorithms is promising. Specifically, we noted a number of potential efforts [248–255].
	\item \emph{Moment invariant theory in real-world application}. In fact, the application of moments and moment invariants is still an active field. This is because different application scenarios have different requirements for the accuracy, complexity, invariance/robustness, and discriminability of image representation. In other words, the special optimization of moment-based image representation considering background knowledge is quite necessary in practice. Specifically, we noted a number of potential efforts [74, 76, 92, 118, 173, 205, 244, 245, 256].
\end{itemize}


\begin{thebibliography}{256}
	
	\bibitem{ref1}	C. E. Shannon, “A mathematical theory of communication,” \emph{Bell Syst. Tech. J.}, vol. 27, no. 3, pp. 379-423. 1984.
	\bibitem{ref2}	L. Zheng, Y. Yang, and Q. Tian, “SIFT meets CNN: A decade survey of instance retrieval,” \emph{IEEE Trans. Pattern Anal. Mach. Intell.}, vol. 40, no. 5, pp. 1224-1244, May 2018.
	\bibitem{ref3}	J. Gu, Z. Wang, J. Kuen, L. Ma, A. Shahroudy, B. Shuai, T. Liu, X. Wang, G. Wang, J. Cai, and T. Chen, “Recent advances in convolutional neural networks,” \emph{Pattern Recognit.}, vol. 77, pp. 354-377, May 2018.
	\bibitem{ref4}	T. H. Chan, K. Jia, S. Gao, J. Lu, Z. Zeng, and Y. Ma, “PCANet: A simple deep learning baseline for image classification?,” \emph{IEEE Trans. Image Process.}, vol. 24, no. 12, pp. 5017-5032, Dec. 2015.
	\bibitem{ref5}	K. He, X. Zhang, S. Ren, and J. Sun, “Spatial pyramid pooling in deep convolutional networks for visual recognition,” \emph{IEEE Trans. Pattern Anal. Mach. Intell.}, vol. 37, no. 9, pp. 1904-1916, Sep. 2015.
	\bibitem{ref6}	V. Balntas, K. Lenc, A. Vedaldi, T. Tuytelaars, J. Matas, and K. Mikolajczyk, “H-Patches: A benchmark and evaluation of handcrafted and learned local descriptors,” \emph{IEEE Trans. Pattern Anal. Mach. Intell.}, vol. 42, no. 11, pp. 2825-2841, Nov. 2020.
	\bibitem{ref7}	K. Mikolajczyk and C. Schmid, “A performance evaluation of local descriptors,” \emph{IEEE Trans. Pattern Anal. Mach. Intell.}, vol. 27, no. 10, pp. 1615-1630, Oct. 2005.
	\bibitem{ref8}	J. Flusser, B. Zitova, and T. Suk, \emph{Moments and Moment Invariants in Pattern Recognition}. John Wiley \& Sons, 2009.
	\bibitem{ref9}	J. Yuan, Y. Wu, and M. Yang, “Discovery of collocation patterns: from visual words to visual phrases.” in \emph{Proc. IEEE Conf. Comput. Vis. Pattern Recognit.}, Jun. 2007, pp. 1-8.
	\bibitem{ref10}	M. K. Hu, “Visual pattern recognition by moment invariants,” \emph{IRE Trans. Inf. Theory}, vol. 8, no. 2, pp. 179-187, Feb. 1962.
	\bibitem{ref11}	R. Mukundan and K. Ramakrishnan, \emph{Moment Functions in Image Analysis: Theory and Applications}. World Scientific, 1998.
	\bibitem{ref12}	M. Pawlak, \emph{Image Analysis by Moments: Reconstruction and Computational Aspects}. Oficyna Wydawnicza Politechniki Wrocławskiej, 2006.
	\bibitem{ref13}	H. Shu, L. Luo, and J. L. Coatrieux, “Moment-based approaches in imaging part 1: Basic features,” \emph{IEEE Eng. Med. Biol.}, vol. 26, no. 5, pp. 70-74, Sep-Oct. 2007.
	\bibitem{ref14}	H. Shu, L. Luo, and J. L. Coatrieux, “Moment-based approaches in imaging part 2: Invariance,” \emph{IEEE Eng. Med. Biol.}, vol. 27, no. 1, pp. 81-83, Jan-Feb. 2008.
	\bibitem{ref15}	H. Shu, L. Luo, and J. L. Coatrieux, “Moment-based approaches in imaging part 3: Computational considerations,” \emph{IEEE Eng. Med. Biol.}, vol. 27, no. 3, pp. 89-91, May-Jun. 2008.
	\bibitem{ref16}	J. Flusser, T. Suk, and B. Zitová, \emph{2D and 3D Image Analysis by Moments}. John Wiley \& Sons, 2016.
	\bibitem{ref17}	T. V. Hoang, “Image representations for pattern recognition,” Ph.D. dissertation, Dept. Comput. Sci., Nancy 2 Univ., Nancy, France, 2011.
	\bibitem{ref18}	G. A. Papakostas, “Over 50 years of image moments and moment invariants,” in \emph{Moments and Moment Invariants – Theory and Applications}, G. A. Papakostas Ed., Science Gate, 2014, pp. 3-32.
	\bibitem{ref19}	P. Kaur, H. S. Pannu, and A. K. Malhi, “Comprehensive study of continuous orthogonal moments-a systematic review,” \emph{ACM Comput. Surv.}, vol. 52, no. 4, Sep. 2019.
	\bibitem{ref20}	C. H. Teh and R. T. Chin, “On image analysis by the methods of moments,” \emph{IEEE Trans. Pattern Anal. Mach. Intell.}, vol. 10, no. 4, pp. 496-513, Jul. 1988.
	\bibitem{ref21}	Y. S. Abu-Mostafa and D. Psaltis, “Recognitive aspects of moment invariants,” \emph{IEEE Trans. Pattern Anal. Mach. Intell.}, vol. 6, no. 6, pp. 698-706, Jun. 1984.
	\bibitem{ref22}	D. Zhang and G. Lu, “Shape-based image retrieval using generic Fourier descriptor,” \emph{Signal Process.-Image Commun.}, vol. 17, no. 10, pp. 825-848, Nov. 2002.
	\bibitem{ref23}	R. Mukundan, S. H. Ong, and P. A. Lee, “Image analysis by Tchebichef moments,” \emph{IEEE Trans. Image Process.}, vol. 10, no. 9, pp. 1357-1364, Sep. 2001.
	\bibitem{ref24}	P. T. Yap, R. Paramesran, and S. H. Ong, “Image analysis by Krawtchouk moments,” \emph{IEEE Trans. Image Process.}, vol. 12, no. 11, pp. 1367-1377, Nov. 2003.
	\bibitem{ref25}	P. T. Yap, R. Paramesran, and S. H. Ong, “Image analysis using Hahn moments,” \emph{IEEE Trans. Pattern Anal. Mach. Intell.}, vol. 29, no. 11, pp. 2057-2062, Nov. 2007.
	\bibitem{ref26}	H. Zhu, H. Shu, J. Zhou, L. Luo, and J. L. Coatrieux, “Image analysis by discrete orthogonal dual Hahn moments,” \emph{Pattern Recognit. Lett.}, vol. 28, no. 13, pp. 1688-1704, Oct. 2007.
	\bibitem{ref27}	H. Zhu, H. Shu, J. Liang, L. Luo, and J. L. Coatrieux, “Image analysis by discrete orthogonal Racah moments,” \emph{Signal Process.}, vol. 87, no. 4, pp. 687-708, Apr. 2007.
	\bibitem{ref28}	A. B. Bhatia and E. Wolf, “On the circle polynomials of Zernike and related orthogonal sets.” \emph{Math. Proc. Camb. Philos. Soc.}, Jan. 1954, pp. 40-48.
	\bibitem{ref29}	M. R. Teague, “Image analysis via the general theory of moments,” \emph{J. Opt. Soc. Am. A}, vol. 70, no. 8, pp. 920-30, Aug. 1980.
	\bibitem{ref30}	J. Shen, Orthogonal Gaussian-Hermite moments for image characterization, in \emph{Proc. SPIE Intell. Rob. Comput. Vis.}, Sep. 1997, pp. 224-233.
	\bibitem{ref31}	K. M. Hosny, “Image representation using accurate orthogonal Gegenbauer moments,” \emph{Pattern Recognit. Lett.}, vol. 32, no. 6, pp. 795-804, Apr. 2011.
	\bibitem{ref32}	H. Zhu, “Image representation using separable two-dimensional continuous and discrete orthogonal moments,” \emph{Pattern Recognit.}, vol. 45, no. 4, pp. 1540-1558, Apr. 2012.
	\bibitem{ref33}	Y. Sheng and L. Shen, “Orthogonal Fourier-Mellin moments for invariant pattern recognition,” \emph{J. Opt. Soc. Am. A}, vol. 11, no. 6, pp. 1748-57, Jun. 1994.
	\bibitem{ref34}	Z. Ping, R. Wu, and Y. Sheng, “Image description with Chebyshev-Fourier moments,” \emph{J. Opt. Soc. Am. A}, vol. 19, no. 9, pp. 1748-1754, Sep. 2002.
	\bibitem{ref35}	G. Amu, S. Hasi, X. Yang, and Z. Ping, “Image analysis by pseudo-Jacobi (p=4, q=3)-Fourier moments,” \emph{Appl. Optics}, vol. 43, no. 10, pp. 2093-2101, Apr. 2004.
	\bibitem{ref36}	Z. Ping, H. Ren, J. Zou, Y. Sheng, and W. Bo, “Generic orthogonal moments: Jacobi-Fourier moments for invariant image description,” \emph{Pattern Recognit.}, vol. 40, no. 4, pp. 1245-1254, Apr. 2007.
	\bibitem{ref37}	T. V. Hoang and S. Tabbone, “Errata and comments on “Generic orthogonal moments: Jacobi-Fourier moments for invariant image description”,” \emph{Pattern Recognit.}, vol. 46, no. 11, pp. 3148-3155, Nov. 2013.
	\bibitem{ref38}	H. Ren, Z. Ping, W. Bo, W. Wu, and Y. Sheng, “Multidistortion-invariant image recognition with radial harmonic Fourier moments,” \emph{J. Opt. Soc. Am. A}, vol. 20, no. 4, pp. 631-637, Apr. 2003.
	\bibitem{ref39}	H. Hu, Y. Zhang, C. Shao, and Q. Ju, “Orthogonal moments based on exponent functions: Exponent-Fourier moments,” \emph{Pattern Recognit.}, vol. 47, no. 8, pp. 2596-2606, Aug. 2014.
	\bibitem{ref40}	P. T. Yap, X. Jiang, and A. C. Kot, “Two-dimensional polar harmonic transforms for invariant image representation,” \emph{IEEE Trans. Pattern Anal. Mach. Intell.}, vol. 32, no. 7, pp. 1259-1270, Jul. 2010.
	\bibitem{ref41}	B. Xiao, J. Ma, and X. Wang, “Image analysis by Bessel-Fourier moments,” \emph{Pattern Recognit.}, vol. 43, no. 8, pp. 2620-2629, Aug. 2010.
	\bibitem{ref42}	S. Liao and M. Pawlak, “On the accuracy of Zernike moments for image analysis,” \emph{IEEE Trans. Pattern Anal. Mach. Intell.}, vol. 20, no. 12, pp. 1358-1364, Dec. 1998.
	\bibitem{ref43}	C. W. Chong, P. Raveendran, and R. Mukundan, “A comparative analysis of algorithms for fast computation of Zernike moments,” \emph{Pattern Recognit.}, vol. 36, no. 3, pp. 731-742, Mar. 2003.
	\bibitem{ref44}	C. Y. Wee and R. Paramesran, “On the computational aspects of Zernike moments,” \emph{Image Vis. Comput.}, vol. 25, no. 6, pp. 967-980, Jun. 2007.
	\bibitem{ref45}	C. Singh and E. Walia, “Algorithms for fast computation of Zernike moments and their numerical stability,” \emph{Image Vis. Comput.}, vol. 29, no. 4, pp. 251-259, Mar. 2011.
	\bibitem{ref46}	X. Wang and S. Liao, “Image reconstruction from orthogonal Fourier-Mellin moments,” in \emph{Proc. Int. Conf. Image Analysis and Recognit.}, 2013, pp. 687-694.
	\bibitem{ref47}	M. Nwali and S. Liao, “A new fast algorithm to compute continuous moments defined in a rectangular region,” \emph{Pattern Recognit.}, vol. 89, pp. 151-160, May 2019.
	\bibitem{ref48}	A. Prata and W. V. Rusch, “Algorithm for computation of Zernike polynomials expansion coefficients,” \emph{Appl. Optics}, vol. 28, no. 4, pp. 749-54, Feb. 1989.
	\bibitem{ref49}	C. Singh and R. Upneja, “Accurate calculation of high order pseudo-Zernike moments and their numerical stability,” \emph{Digit. Signal Prog.}, vol. 27, pp. 95-106, Apr. 2014.
	\bibitem{ref50}	C. Singh and R. Upneja, “Accurate computation of orthogonal Fourier-Mellin moments,” \emph{J. Math. Imaging Vis.}, vol. 44, no. 3, pp. 411-431, Nov. 2012.
	\bibitem{ref51}	Y. Xin, M. Pawlak, and S. Liao, “Accurate computation of Zernike moments in polar coordinates,” \emph{IEEE Trans. Image Process.}, vol. 16, no. 2, pp. 581-587, Feb. 2007.
	\bibitem{ref52}	K. M. Hosny, M. A. Shouman, and H. M. A. Salam, “Fast computation of orthogonal Fourier-Mellin moments in polar coordinates,” \emph{J. Real-Time Image Process.}, vol. 6, no. 2, pp. 73-80, Jun. 2011.
	\bibitem{ref53}	C. Camacho-Bello, C. Toxqui-Quitl, A. Padilla-Vivanco, and J. J. Baez-Rojas, “High-precision and fast computation of Jacobi-Fourier moments for image description,” \emph{J. Opt. Soc. Am. A}, vol. 31, no. 1, pp. 124-134, Jan. 2014.
	\bibitem{ref54}	R. Mukundan and K. R. Ramakrishnan, “Fast computation of Legendre and Zernike moments,” \emph{Pattern Recognit.}, vol. 28, no. 9, pp. 1433-42, Sep. 1995.
	\bibitem{ref55}	C. Di Ruberto, L. Putzu, and G. Rodriguez, “Fast and accurate computation of orthogonal moments for texture analysis,” \emph{Pattern Recognit.}, vol. 83, pp. 498-510, Nov. 2018.
	\bibitem{ref56}	R. Upneja and C. Singh, “Fast computation of Jacobi-Fourier moments for invariant image recognition,” \emph{Pattern Recognit.}, vol. 48, no. 5, pp. 1836-1843, May 2015.
	\bibitem{ref57}	G. A. Papakostas, Y. S. Boutalis, D. A. Karras, and B. G. Mertzios, “A new class of Zernike moments for computer vision applications,” \emph{Inf. Sci.}, vol. 177, no. 13, pp. 2802-2819, Jul. 2007.
	\bibitem{ref58}	G. A. Papakostas, Y. S. Boutalis, D. A. Karras, and B. G. Mertzios, “Modified factorial-free direct methods for Zernike and pseudo-Zernike moment computation,” \emph{IEEE Trans. Instrum. Meas.}, vol. 58, no. 7, pp. 2121-2131, Jul. 2009.
	\bibitem{ref59}	J. Saez-Landete, “Comments on “Fast computation of Jacobi-Fourier moments for invariant image recognition”,” \emph{Pattern Recognit.}, vol. 67, pp. 16-22, Jul. 2017.
	\bibitem{ref60}	K. M. Hosny, and M. M. Darwish, “A kernel-based method for fast and accurate computation of PHT in polar coordinates,” \emph{J. Real-Time Image Process.}, vol. 16, no. 4, pp. 1235-1247, Aug. 2019.
	\bibitem{ref61}	A. Averbuch, R. R. Coifman, D. L. Donoho, M. Elad, and M. Israeli, “Fast and accurate polar Fourier transform,” \emph{Appl. Comput. Harmon. Anal.}, vol. 21, no. 2, pp. 145-167, Sep. 2006.
	\bibitem{ref62}	H. Yang, S. Qi, C. Wang, S. Yang, and X. Wang, “Image analysis by log-polar Exponent-Fourier moments,” \emph{Pattern Recognit.}, vol. 101, May 2020.
	\bibitem{ref63}	T. V. Hoang and S. Tabbone, “Fast generic polar harmonic transforms,” \emph{IEEE Trans. Image Process.}, vol. 23, no. 7, pp. 2961-2971, Jul. 2014.
	\bibitem{ref64}	C. Singh and S. K. Ranade, “A high capacity image adaptive watermarking scheme with radial harmonic Fourier moments,” \emph{Digit. Signal Prog.}, vol. 23, no. 5, pp. 1470-1482, Sep. 2013.
	\bibitem{ref65}	T. V. Hoang and S. Tabbone, “Fast computation of orthogonal polar harmonic transforms,” in \emph{Proc. Int. Conf. Pattern Recognit.}, 2012, pp. 3160-3163.
	\bibitem{ref66}	C. Singh and A. Kaur, “Fast computation of polar harmonic transforms,” \emph{J. Real-Time Image Process.}, vol. 10, no. 1, pp. 59-66, Mar. 2015.
	\bibitem{ref67}	H. Yang, S. Qi, P. Niu, and X. Wang, “Color image zero-watermarking based on fast quaternion generic polar complex exponential transform,” \emph{Signal Process.-Image Commun.}, vol. 82, Mar. 2020.
	\bibitem{ref68}	C. Wang, X. Wang, and Z. Xia, “Geometrically invariant image watermarking based on fast Radial Harmonic Fourier Moments,” \emph{Signal Process.-Image Commun.}, vol. 45, pp. 10-23, Jul. 2016.
	\bibitem{ref69}	X. Wang, C. Wang, H. Yang, and P. Niu, “Robust and effective multiple copy-move forgeries detection and localization,” \emph{ Pattern Anal. Appl. }, Feb. 2021.
	\bibitem{ref70}	Z. Ping, Y. Jiang, S. Zhou, and Y. Wu, “FFT algorithm of complex exponent moments and its application in image recognition,” in \emph{Proc. SPIE Int. Conf. Digit. Image Process.}, Apr. 2014, pp. 4177-4180.
	\bibitem{ref71}	S. P. Singh and S. Urooj, “Accurate and fast computation of Exponent-Fourier moment,” \emph{Arab. J. Sci. Eng.}, vol. 42, no. 8, pp. 3299-3306, Aug. 2017.
	\bibitem{ref72}	S. K. Hwang and W. Y. Kim, “A novel approach to the fast computation of Zernike moments,” \emph{Pattern Recognit.}, vol. 39, no. 11, pp. 2065-2076, Nov. 2006.
	\bibitem{ref73}	M. S. Al-Rawi, “Fast computation of pseudo Zernike moments,” \emph{J. Real-Time Image Process.}, vol. 5, no. 1, pp. 3-10, Mar. 2010.
	\bibitem{ref74}	B. Chen, G. Coatrieux, J. Wu, Z. Dong, J. L. Coatrieux, and H. Shu, “Fast computation of sliding discrete Tchebichef moments and its application in duplicated regions detection,” \emph{IEEE Trans. Signal Process.}, vol. 63, no. 20, pp. 5424-5436, Oct. 2015.
	\bibitem{ref75}	J. Martinez and F. Thomas, “Efficient computation of local geometric moments,” \emph{IEEE Trans. Image Process.}, vol. 11, no. 9, pp. 1102-1111, Sep. 2002.
	\bibitem{ref76}	A. Bera, P. Klesk, and D. Sychel, “Constant-time calculation of Zernike moments for detection with rotational invariance,” \emph{IEEE Trans. Pattern Anal. Mach. Intell.}, vol. 41, no. 3, pp. 537-551, Mar. 2019.
	\bibitem{ref77}	H. Bay, T. Tuytelaars, and L. Van Gool, “SURF: Speeded up robust features,” in \emph{ Proc. Eur. Conf. Comput. Vision}, May 2006, pp. 404-417.
	\bibitem{ref78}	R. Benouini, I. Batioua, K. Zenkouar, A. Zahi, H. El Fadili, and H. Qjidaa, “Fast and accurate computation of Racah moment invariants for image classification,” \emph{Pattern Recognit.}, vol. 91, pp. 100-110, Jul. 2019.
	\bibitem{ref79}	S. Pei and C. Lin, “Image normalization for pattern recognition,” \emph{Image Vis. Comput.}, vol. 13, no. 10, pp. 711-723. 1995.
	\bibitem{ref80}	S. Dinggang and H. H. S. Ip, “Generalized affine invariant image normalization,” \emph{IEEE Trans. Pattern Anal. Mach. Intell.}, vol. 19, no. 5, pp. 431-40, May 1997.
	\bibitem{ref81}	S. Dongseok, J. K. Pollard, and J. P. Muller, “Accurate geometric correction of ATSR images,” \emph{IEEE Trans. Geosci. Remote Sensing}, vol. 35, no. 4, pp. 997-1006, Jul. 1997.
	\bibitem{ref82}	C. Wang, X. Wang, C. Zhang, and Z. Xia, “Geometric correction based color image watermarking using fuzzy least squares support vector machine and Bessel K form distribution,” \emph{Signal Process.}, vol. 134, pp. 197-208, May 2017.
	\bibitem{ref83}	L. G. Brown, “A survey of image registration techniques,” \emph{ACM Comput. Surv.}, vol. 24, no. 4, pp. 325-376. 1992.
	\bibitem{ref84}	B. Zitova and J. Flusser, “Image registration methods: a survey,” \emph{Image Vis. Comput.}, vol. 21, no. 11, pp. 977-1000. 2003.
	\bibitem{ref85}	C. W. Chong, P. Raveendran, and R. Mukundan, “Translation invariants of Zernike moments,” \emph{Pattern Recognit.}, vol. 36, no. 8, pp. 1765-1773, Aug. 2003.
	\bibitem{ref86}	E. G. Karakasis, G. A. Papakostas, D. E. Koulouriotis, and V. D. Tourassis, “Generalized dual Hahn moment invariants,” \emph{Pattern Recognit.}, vol. 46, no. 7, pp. 1998-2014, Jul. 2013.
	\bibitem{ref87}	J. Flusser and T. Suk, “Pattern recognition by affine moment invariants,” \emph{Pattern Recognit.}, vol. 26, no. 1, pp. 167-74, Jan. 1993.
	\bibitem{ref88}	C. W. Chong, P. Raveendran, and R. Mukundan, “Translation and scale invariants of Legendre moments,” \emph{Pattern Recognit.}, vol. 37, no. 1, pp. 119-129, Jan. 2004.
	\bibitem{ref89}	S. Belkasim, E. Hassan, and T. Obeidi, “Explicit invariance of Cartesian Zernike moments,” \emph{Pattern Recognit. Lett.}, vol. 28, no. 15, pp. 1969-1980, Nov. 2007.
	\bibitem{ref90}	H. Zhu, H. Shu, T. Xia, L. Luo, and J. L. Coatrieux, “Translation and scale invariants of Tchebichef moments,” \emph{Pattern Recognit.}, vol. 40, no. 9, pp. 2530-2542, Sep. 2007.
	\bibitem{ref91}	A. V. Oppenheim and J. S. Lim, “The importance of phase in signals,” \emph{Proc. IEEE}, vol. 69, no. 5, pp. 529-41, May 1981.
	\bibitem{ref92}	J. Revaud, G. Lavoue, and A. Baskurt, “Improving Zernike moments comparison for optimal similarity and rotation angle retrieval,” \emph{IEEE Trans. Pattern Anal. Mach. Intell.}, vol. 31, no. 4, pp. 627-636, Apr. 2009.
	\bibitem{ref93}	J. Flusser, “On the independence of rotation moment invariants,” \emph{Pattern Recognit.}, vol. 33, no. 9, pp. 1405-1410, Sep. 2000.
	\bibitem{ref94}	L. Shao, R. Yan, X. Li, and Y. Liu, “From heuristic optimization to dictionary learning: A review and comprehensive comparison of image denoising algorithms,” \emph{IEEE T. Cybern.}, vol. 44, no. 7, pp. 1001-1013, Jul. 2014.
	\bibitem{ref95}	K. Zhang, W. Zuo, Y. Chen, D. Meng, and L. Zhang, “Beyond a gaussian denoiser: Residual learning of deep CNN for image denoising,” \emph{IEEE Trans. Image Process.}, vol. 26, no. 7, pp. 3142-3155, Jul. 2017.
	\bibitem{ref96}	B. Xiao, J. Cui, H. Qin, W. Li, and G. Wang, “Moments and moment invariants in the Radon space,” \emph{Pattern Recognit.}, vol. 48, no. 9, pp. 2772-2784, Sep. 2015.
	\bibitem{ref97}	T. V. Hoang and S. Tabbone, “Invariant pattern recognition using the RFM descriptor,” \emph{Pattern Recognit.}, vol. 45, no. 1, pp. 271-284, Jan. 2012.
	\bibitem{ref98}	Q. Miao, J. Liu, W. Li, J. Shi, and Y. Wang, “Three novel invariant moments based on radon and polar harmonic transforms,” \emph{Opt. Commun.}, vol. 285, no. 6, pp. 1044-1048, Mar. 2012.
	\bibitem{ref99}	K. Jafari-Khouzani and H. Soltanian-Zadeh, “Rotation-invariant multiresolution texture analysis using Radon and wavelet transforms,” \emph{IEEE Trans. Image Process.}, vol. 14, no. 6, pp. 783-795, Jun. 2005.
	\bibitem{ref100}	D. Kundur and D. Hatzinakos, “Blind image deconvolution,” \emph{IEEE Signal Process. Mag.}, vol. 13, no. 3, pp. 43-64, May 1996.
	\bibitem{ref101}	P. Campisi and K. Egiazarian, \emph{Blind Image Deconvolution: Theory and Applications}. CRC press, 2017.
	\bibitem{ref102}	J. Kostkova, J. Flusser, M. Lebl, and M. Pedone, “Handling Gaussian blur without deconvolution,” \emph{Pattern Recognit.}, vol. 103, Jul. 2020.
	\bibitem{ref103}	J. Flusser, T. Suk, J. Boldys, and B. Zitova, “Projection operators and moment invariants to image blurring,” \emph{IEEE Trans. Pattern Anal. Mach. Intell.}, vol. 37, no. 4, pp. 786-802, Apr. 2015.
	\bibitem{ref104}	J. Flusser, S. Farokhi, C. Hoschl, T. Suk, B. Zitova, and M. Pedone, “Recognition of images degraded by Gaussian blur,” \emph{IEEE Trans. Image Process.}, vol. 25, no. 2, pp. 790-806, Feb. 2016.
	\bibitem{ref105}	M. Pedone, J. Flusser, and J. Heikkila, “Registration of images with N-Fold dihedral blur,” \emph{IEEE Trans. Image Process.}, vol. 24, no. 3, pp. 1036-1045, Mar. 2015.
	\bibitem{ref106}	T. Suk and J. Flusser, “Tensor method for constructing 3D moment invariants,” in \emph{Proc. Int. Conf. Comput. Anal. Images Patterns}, Aug. 2011, pp. 212-219.
	\bibitem{ref107}	T. Suk and J. Flusser, “Graph method for generating affine moment invariants,” in \emph{Proc. Int. Conf. Pattern Recognit.}, Aug. 2004, pp. 192-195.
	\bibitem{ref108}	C. H. Lo and H. S. Don, “3-D moment forms: their construction and application to object identification and positioning,” \emph{IEEE Trans. Pattern Anal. Mach. Intell.}, vol. 11, no. 10, pp. 1053-64, Oct. 1989.
	\bibitem{ref109}	G. Mamistvalov, “N-dimensional moment invariants and conceptual mathematical theory of recognition n-dimensional solids,” \emph{IEEE Trans. Pattern Anal. Mach. Intell.}, vol. 20, no. 8, pp. 819-831, Aug. 1998.
	\bibitem{ref110}	D. Xu and H. Li, “Geometric moment invariants,” \emph{Pattern Recognit.}, vol. 41, no. 1, pp. 240-249, Jan. 2008.
	\bibitem{ref111}	E. Li and H. Li. (2017) “Reflection invariant and symmetry detection.” [Online]. Available: https://arxiv.org/abs/1705.10768
	\bibitem{ref112}	E. Li, H. Mo, D. Xu, and H. Li, “Image projective invariants,” \emph{IEEE Trans. Pattern Anal. Mach. Intell.}, vol. 41, no. 5, pp. 1144-1157, May 2019.
	\bibitem{ref113}	Z. He, M. Hanlin, H. You, L. Qi, and L. Hua, “Differential and integral invariants under Mobius transformation, ” in \emph{Proc. Chinese Conf. Pattern Recognit. Comput. Vis.}, Nov. 2018, pp.280-291.
	\bibitem{ref114}	H. Zhang, H. Mo, Y. Hao, Q. Li, S. Li, and H. Li, “Fast and efficient calculations of structural invariants of chirality,” \emph{Pattern Recognit. Lett.}, vol. 128, pp. 270-277, Dec. 2019.
	\bibitem{ref115}	H. You, M. Hanlin, L. Qi, Z. He, and L. Hua. (2019) “Dual affine moment invariants.” [Online]. Available: https://arxiv.org/abs/1911.08233
	\bibitem{ref116}	D. Xu and H. Li, “3-D curve moment invariants for curve recognition,” in \emph{Proc. Intell. Comput. Signal Process. Pattern Recognit.}, 2006, pp. 572-577.
	\bibitem{ref117}	D. Xu and H. Li, “3-D surface moment invariants,” in \emph{Proc. Int. Conf. Pattern Recognit.}, Aug. 2006, pp. 173-176.
	\bibitem{ref118}	J. Kostkova, T. Suk, and J. Flusser, “Affine invariants of vector fields,” \emph{IEEE Trans. Pattern Anal. Mach. Intell.}, Nov. 2019.
	\bibitem{ref119}	J. Kostkova, T. Suk, and J. Flusser, “Affine moment invariants of vector fields,” in \emph{Proc. Int. Conf. Image Process.}, Oct. 2018, pp. 1338-1342.
	\bibitem{ref120}	M. Langbein and H. Hagen, “A generalization of moment invariants on 2D vector fields to tensor fields of arbitrary order and dimension,” in \emph{Proc. Int. Symp. Vis. Comput.}, Nov. 2009, pp. 1151-1160.
	\bibitem{ref121}	R. Bujack, J. Kasten, I. Hotz, G. Scheuermann, and E. Hitzer, “Moment invariants for 3D flow fields via normalization,” in \emph{Proc. IEEE Pacific Vis. Symp.}, Apr. 2015, pp. 9-16.
	\bibitem{ref122}	R. Bujack and H. Hagen, “Moment invariants for multi-dimensional data,” in \emph{Modelling, Analysis, and Visualization of Anisotropy}, E. Ozerslan, T. Schultz, and I. Hotz, Eds. Mathematica and Visualization, 2017, pp. 43-64.
	\bibitem{ref123}	H. Zhang, H. Shu, P. Haigron, B. Li, and L. Luo, “Construction of a complete set of orthogonal Fourier-Mellin moment invariants for pattern recognition applications,” \emph{Image Vis. Comput.}, vol. 28, no. 1, pp. 38-44, Jan. 2010.
	\bibitem{ref124}	B. Xiao and G. Wang, “Generic radial orthogonal moment invariants for invariant image recognition,” \emph{J. Vis. Commun. Image Represent.}, vol. 24, no. 7, pp. 1002-1008, Oct. 2013.
	\bibitem{ref125}	J. Yang, Z. Lu, Y. Y. Tang, Z. Yuan, and Y. Chen, “Quasi Fourier-Mellin transform for affine invariant features,” \emph{IEEE Trans. Image Process.}, vol. 29, pp. 4114-4129, Jan. 2020.
	\bibitem{ref126}	B. Xiao, J. Ma, and J. Cui, “Radial Tchebichef moment invariants for image recognition,” \emph{J. Vis. Commun. Image Represent.}, vol. 23, no. 2, pp. 381-386, Feb. 2012.
	\bibitem{ref127}	B. Xiao, G. Wang, and W. Li, “Radial shifted Legendre moments for image analysis and invariant image recognition,” \emph{Image Vis. Comput.}, vol. 32, no. 12, pp. 994-1006, Dec. 2014.
	\bibitem{ref128}	B. Yang, J. Flusser, and T. Suk, “Design of high-order rotation invariants from Gaussian-Hermite moments,” \emph{Signal Process.}, vol. 113, pp. 61-67, Aug. 2015.
	\bibitem{ref129}	B. Yang, J. Kostkova, J. Flusser, and T. Suk, “Scale invariants from Gaussian-Hermite moments,” \emph{Signal Process.}, vol. 132, pp. 77-84, Mar. 2017.
	\bibitem{ref130}	B. Yang, J. Flusser, and J. Kautsky, “Rotation of 2D orthogonal polynomials,” \emph{Pattern Recognit. Lett.}, vol. 102, pp. 44-49, Jan. 2018.
	\bibitem{ref131}	B. Yang, T. Suk, J. Flusser, Z. Shi, and X. Chen, “Rotation invariants from Gaussian-Hermite moments of color images,” \emph{Signal Process.}, vol. 143, pp. 282-291, Feb. 2018.
	\bibitem{ref132}	E. Li, Y. Huang, D. Xu, and H. Li. (2017) “Shape DNA: Basic generating functions for geometric moment invariants.” [Online]. Available: https://arxiv.org/abs/1703.02242
	\bibitem{ref133}	E. Li and H. Li. (2017) “Isomorphism between differential and moment invariants under affine transform.” [Online]. Available: https://arxiv.org/abs/1705.08264
	\bibitem{ref134}	Y. Pei, Y. Huang, Q. Zou, X. Zhang, and S. Wang, “Effects of image degradation and degradation removal to CNN-based image classification,” \emph{IEEE Trans. Pattern Anal. Mach. Intell.}, Nov. 2019.
	\bibitem{ref135}	J. Flusser, T. Suk, and S. Saic, “Recognition of images degraded by linear motion blur without restoration,” in \emph{Theoretical Foundations of Computer Vision}, W. Kropatsch, R. Klette, F. Solina, and R. Albrecht, Eds. Computing Supplement, 1996, pp. 37-51.
	\bibitem{ref136}	J. Flusser, T. Suk, and S. Saic, “Image features invariant with respect to blur,” \emph{Pattern Recognit.}, vol. 28, no. 11, pp. 1723-32, Nov. 1995.
	\bibitem{ref137}	J. Flusser and B. Zitova, “Invariants to convolution with circularly symmetric PSF,” in \emph{Proc. Int. Conf. Pattern Recognit.}, vol. 2, pp. 11-14, Aug. 2004.
	\bibitem{ref138}	B. Chen, H. Shu, H. Zhang, G. Coatrieux, L. Luo, and J. L. Coatrieux, “Combined invariants to similarity transformation and to blur using orthogonal Zernike moments,” \emph{IEEE Trans. Image Process.}, vol. 20, no. 2, pp. 345-360, Feb. 2011.
	\bibitem{ref139}	E. G. Karakasis, G. A. Papakostas, D. E. Koulouriotis, and V. D. Tourassis, “A unified methodology for computing accurate quaternion color moments and moment invariants,” \emph{IEEE Trans. Image Process.}, vol. 23, no. 2, pp. 596-611, Feb. 2014.
	\bibitem{ref140}	K. M. Hosny and M. M. Darwish, “New set of multi-channel orthogonal moments for color image representation and recognition,” \emph{Pattern Recognit.}, vol. 88, pp. 153-173, Apr. 2019.
	\bibitem{ref141}	B. Chen, H. Shu, H. Zhang, G. Chen, C. Toumoulin, J. L. Dillenseger, and L. Luo, “Quaternion Zernike moments and their invariants for color image analysis and object recognition,” \emph{Signal Process.}, vol. 92, no. 2, pp. 308-318, Feb. 2012.
	\bibitem{ref142}	B. Chen, H. Shu, G. Coatrieux, G. Chen, X. Sun, and J. L. Coatrieux, “Color image analysis by quaternion-type moments,” \emph{J. Math. Imaging Vis.}, vol. 51, no. 1, pp. 124-144, Jan. 2015.
	\bibitem{ref143}	Y. Li, “Quaternion polar harmonic transforms for color images,” \emph{IEEE Signal Process. Lett.}, vol. 20, no. 8, pp. 803-806, Aug. 2013.
	\bibitem{ref144}	L. Guo and M. Zhu, “Quaternion Fourier-Mellin moments for color images,” \emph{Pattern Recognit.}, vol. 44, no. 2, pp. 187-195, Feb. 2011.
	\bibitem{ref145}	I. Batioua, R. Benouini, K. Zenkouar, S. Najah, H. E. Fadili, and H. Qjidaa, “3D image representation using separable discrete orthogonal moments,” \emph{Procedia Comput. Sci.}, vol. 148, pp. 389-398, 2019.
	\bibitem{ref146}	M. Kazhdan, T. Funkhouser, and S. Rusinkiewicz, “Rotation invariant spherical harmonic representation of 3D shape descriptors,” in \emph{Proc. Symp. Geom. Process.}, Jun. 2003, pp. 156-164.
	\bibitem{ref147}	M. Novotni and R. Klein, “Shape retrieval using 3D Zernike descriptors,” \emph{Comput.-Aided Des.}, vol. 36, no. 11, pp. 1047-1062, Sep. 2004.
	\bibitem{ref148}	N. Canterakis, “3D Zernike moments and Zernike affine invariants for 3D image analysis and recognition.” In \emph{Proc. Scandinavian Conf. Image Analysis }, 1999, pp. 85-93.
	\bibitem{ref149}	T. V. Hoang and S. Tabbone, “Generic polar harmonic transforms for invariant image description,” in \emph{Proc. Int. Conf. Image Process.}, 2011, pp. 829-832.
	\bibitem{ref150}	T. V. Hoang and S. Tabbone, “Generic polar harmonic transforms for invariant image representation,” \emph{Image Vis. Comput.}, vol. 32, no. 8, pp. 497-509, Aug. 2014.
	\bibitem{ref151}	B. Xiao, L. Li, Y. Li, W. Li, and G. Wang, “Image analysis by fractional-order orthogonal moments,” \emph{Inf. Sci.}, vol. 382, pp. 135-149, Mar. 2017.
	\bibitem{ref152}	H. Yang, S. Qi, J. Tian, P. Niu, X. Wang, “Robust and discriminative image representation: Fractional-order Jacobi-Fourier moments,” \emph{Pattern Recognit.}, vol. 115, Jul. 2021.
	\bibitem{ref153}	H. Zhu, “Image representation using separable two-dimensional continuous and discrete orthogonal moments,” \emph{Pattern Recognit.}, vol. 45, no. 4, pp. 1540-1558, Apr. 2012.
	\bibitem{ref154}	C. F. Dunkl and Y. Xu, \emph{Orthogonal Polynomials of Several Variables}. Cambridge University Press, 2014.
	\bibitem{ref155}	I. Batioua, R. Benouini, K. Zenkouar, and H. El Fadili, “Image analysis using new set of separable two-dimensional discrete orthogonal moments based on Racah polynomials,” \emph{EURASIP J. Image Video Process.}, Mar. 2017.
	\bibitem{ref156}	A. Hmimid, M. Sayyouri, and H. Qjidaa, “Fast computation of separable two-dimensional discrete invariant moments for image classification,” \emph{Pattern Recognit.}, vol. 48, no. 2, pp. 509-521, Feb. 2015.
	\bibitem{ref157}	E. D. Tsougenis, G. A. Papakostas, and D. E. Koulouriotis, “Image watermarking via separable moments,” \emph{Multimed. Tools Appl.}, vol. 74, no. 11, pp. 3985-4012, Jun. 2015.
	\bibitem{ref158}	I. Batioua, R. Benouini, K. Zenkouar, A. Zahi, and E. F. Hakim, “3D image analysis by separable discrete orthogonal moments based on Krawtchouk and Tchebichef polynomials,” \emph{Pattern Recognit.}, vol. 71, pp. 264-277, Nov. 2017.
	\bibitem{ref159}	X. Wang, T. Yang, and F. Guo, “Image analysis by circularly semi-orthogonal moments,” \emph{Pattern Recognit.}, vol. 49, pp. 226-236, Jan. 2016.
	\bibitem{ref160}	B. He, J. Cui, B. Xiao, and Y. Peng, “General semi-orthogonal moments with parameter modulation,” \emph{J. Image Graph.}, vol. 24, no. 10, pp. 1711-1727, 2019.
	\bibitem{ref161}	B. He, J. Cui, B. Xiao, and Y. Peng, “Image analysis using modified Exponent-Fourier moments,” \emph{EURASIP J. Image Video Process.}, vol. 72, Jul. 2019.
	\bibitem{ref162}	H. Zhu, Y. Yang, X. Zhu, Z. Gui, and H. Shu, “General form for obtaining unit disc-based generalized orthogonal moments,” \emph{IEEE Trans. Image Process.}, vol. 23, no. 12, pp. 5455-5469, Dec. 2014.
	\bibitem{ref163}	H. Zhu, Y. Yang, Z. Gui, Y. Zhu, and Z. Chen, “Image analysis by generalized Chebyshev-Fourier and generalized pseudo-Jacobi-Fourier moments,” \emph{Pattern Recognit.}, vol. 51, pp. 1-11, Mar. 2016.
	\bibitem{ref164}	T. Xia, H. Zhu, H. Shu, P. Haigron, and L. Luo, “Image description with generalized pseudo-Zernike moments,” \emph{J. Opt. Soc. Am. A}, vol. 24, no. 1, pp. 50-59, Jan. 2007.
	\bibitem{ref165}	M. Qi, B. Li, and H. Sun, “Image representation by harmonic transforms with parameters in SL(2, R),” \emph{J. Vis. Commun. Image Represent.}, vol. 35, pp. 184-192, Feb. 2016.
	\bibitem{ref166}	B. He and J. Cui, “Weighted spherical Bessel-Fourier image moments,” \emph{Cluster Comput.}, vol. 22, pp. 12985-12996, Sep. 2019.
	\bibitem{ref167}	X. Liu, Y. Wu, Z. Shao, and J. Wu, “The modified generic polar harmonic transforms for image representation,” \emph{Pattern Anal. Appl.}, vol. 23, no. 2, pp. 785-795, May 2020.
	\bibitem{ref168}	T. Yang, J. Ma, Y. Miao, X. Wang, B. Xiao, B. He, and Q. Meng, “Quaternion weighted spherical Bessel-Fourier moment and its invariant for color image reconstruction and object recognition,” \emph{Inf. Sci.}, vol. 505, pp. 388-405, Dec. 2019.
	\bibitem{ref169}	C. Wang, X. Wang, Y. Li, Z. Xia, and C. Zhang, “Quaternion polar polar harmonic Fourier moments for color images,” \emph{Inf. Sci.}, vol. 450, pp. 141-156, Jun. 2018.
	\bibitem{ref170}	X. Wang, W. Li, H. Yang, P. Niu, and Y. Li, “Invariant quaternion radial harmonic Fourier moments for color image retrieval,” \emph{Opt. Laser Technol.}, vol. 66, pp. 78-88, Mar. 2015.
	\bibitem{ref171}	X. Wang, P. Niu, H. Yang, C. Wang, and A. Wang, “A new robust color image watermarking using local quaternion exponent moments,” \emph{Inf. Sci.}, vol. 277, pp. 731-754, Sep. 2014.
	\bibitem{ref172}	Z. Shao, H. Shu, J. Wu, B. Chen, and J. L. Coatrieux, “Quaternion Bessel-Fourier moments and their invariant descriptors for object reconstruction and recognition,” \emph{Pattern Recognit.}, vol. 47, no. 2, pp. 603-611, Feb. 2014.
	\bibitem{ref173}	J. Wang, T. Li, X. Luo, Y. Q. Shi, and S. K. Jha, “Identifying computer generated images based on quaternion central moments in color quaternion wavelet domain,” \emph{IEEE Trans. Circuits Syst. Video Technol.}, vol. 29, no. 9, pp. 2775-2785, Sep. 2019.
	\bibitem{ref174}	S. Said, N. Le Bihan, and S. J. Sangwine, “Fast complexified quaternion Fourier transform,” \emph{IEEE Trans. Signal Process.}, vol. 56, no. 4, pp. 1522-1531, Apr. 2008.
	\bibitem{ref175}	T. Bulow and G. Sommer, “Hypercomplex signals-a novel extension of the analytic signal to the multidimensional case,” \emph{IEEE Trans. Signal Process.}, vol. 49, no. 11, pp. 2844-2852, Nov. 2001.
	\bibitem{ref176}	Y. Chen, X. Xiao, and Y. Zhou, “Low-rank quaternion approximation for color image processing,” \emph{IEEE Trans. Image Process.}, vol. 29, pp. 1426-1439, Sep. 2020.
	\bibitem{ref177}	S. Zhang, Y. Tay, L. Yao, and Q. Liu, “Quaternion knowledge graph embeddings,” in \emph{Proc. Adv. Neural Inf. Process. Syst.}, pp. 2731-2741, 2019.
	\bibitem{ref178}	Y. Liu, Y. Zheng, J. Lu, J. Cao, and L. Rutkowski, “Constrained quaternion-variable convex optimization: a quaternion-valued recurrent neural network approach,” \emph{IEEE Trans. Neural Netw. Learn. Syst.}, vol. 31, no. 3, pp. 1022-1035, Mar. 2020.
	\bibitem{ref179}	X. Zhu, Y. Xu, H. Xu, and C. Chen, “Quaternion convolutional neural networks,” in \emph{ Proc. Eur. Conf. Comput. Vision}, pp. 645-661, 2018.
	\bibitem{ref180}	B. Chen, X. Qi, X. Sun, and Y. Q. Shi, “Quaternion pseudo-Zernike moments combining both of RGB information and depth information for color image splicing detection,” \emph{J. Vis. Commun. Image Represent.}, vol. 49, pp. 283-290, Nov. 2017.
	\bibitem{ref181}	M. Yamni, H. Karmouni, M. Sayyouri, H. Qjidaa, and J. Flusser, “Novel octonion moments for color stereo image analysis,” \emph{Digit. Signal Prog.}, vol. 108, Jan. 2021.
	\bibitem{ref182}	D. Xu and D. P. Mandic, “The theory of quaternion matrix derivatives,” \emph{IEEE Trans. Signal Process.}, vol. 63, no. 6, pp. 1543-1556, Mar. 2015.
	\bibitem{ref183}	P. Suetens, \emph{Fundamentals of Medical Imaging}. Cambridge University Press, 2017.
	\bibitem{ref184}	S. Marschner and P. Shirley, \emph{Fundamentals of computer graphics}. CRC Press, 2018.
	\bibitem{ref185}	J. M. Pozo, M. C. Villa-Uriol, and A. F. Frangi, “Efficient 3D geometric and Zernike moments computation from unstructured surface meshes,” \emph{IEEE Trans. Pattern Anal. Mach. Intell.}, vol. 33, no. 3, pp. 471-484, Mar. 2011.
	\bibitem{ref186}	H. Karmouni, T. Jahid, M. Sayyouri, R. El Alami, and H. Qjidaa, “Fast 3D image reconstruction by cuboids and 3D Charlier's moments,” \emph{J. Real-Time Image Process.}, vol. 17, no. 4, pp. 949-965, Aug. 2020.
	\bibitem{ref187}	A. Daoui, M. Yamni, O. El Ogri, H. Karmouni, M. Sayyouri, and H. Qjidaa, “New algorithm for large-sized 2D and 3D image reconstruction using higher-order Hahn moments,” \emph{Circuits Syst. Signal Process.}, vol. 39, no. 9, pp. 4552-4577, Sep. 2020.
	\bibitem{ref188}	R. Benouini, I. Batioua, K. Zenkouar, S. Najah, and H. Qjidaa, “Efficient 3D object classification by using direct Krawtchouk moment invariants,” \emph{Multimed. Tools Appl.}, vol. 77, no. 20, pp. 27517-27542, Oct. 2018.
	\bibitem{ref189}	I. Batioua, R. Benouini, and K. Zenkouar, “Image recognition using new set of separable three-dimensional discrete orthogonal moment invariants,” \emph{Multimed. Tools Appl.}, vol. 79, no. 19-20, pp. 13217-13245, May 2020.
	\bibitem{ref190}	B. Xiao, J. Luo, X. Bi, W. Li, and B. Chen, “Fractional discrete Tchebyshev moments and their applications in image encryption and watermarking,” \emph{Inf. Sci.}, vol. 516, pp. 545-559, Apr. 2020.
	\bibitem{ref191}	M. Yamni, A. Daoui, O. El Ogri, H. Karmouni, M. Sayyouri, H. Qjidaa, and J. Flusser, “Fractional Charlier moments for image reconstruction and image watermarking,” \emph{Signal Process.}, vol. 171, Jun. 2020.
	\bibitem{ref192}	B. Chen, M. Yu, Q. Su, and L. Li, “Fractional quaternion cosine transform and its application in color image copy-move forgery detection,” \emph{Multimed. Tools Appl.}, vol. 78, no. 7, pp. 8057-8073, Apr. 2019.
	\bibitem{ref193}	R. Benouini, I. Batioua, K. Zenkouar, A. Zahi, S. Najah, and H. Qjidaa, “Fractional-order orthogonal Chebyshev moments and moment invariants for image representation and pattern recognition,” \emph{Pattern Recognit.}, vol. 86, pp. 332-343, Feb. 2019.
	\bibitem{ref194}	M. Hosny, M. M. Darwish, and T. Aboelenen, “New fractional-order Legendre-Fourier moments for pattern recognition applications,” \emph{Pattern Recognit.}, vol. 103, Jul. 2020.
	\bibitem{ref195}	H. Zhang, Z. Li, and Y. Liu, “Fractional orthogonal Fourier-Mellin moments for pattern recognition,” in \emph{Proc. Chinese Conf. Pattern Recognit.}, 2016, pp. 766-778.
	\bibitem{ref196}	B. Chen, M. Yu, Q. Su, H. J. Shim, and Y. Q. Shi, “Fractional quaternion Zernike moments for robust color image copy-move forgery detection,” \emph{IEEE Access}, vol. 6, pp. 56637-56646, Sep. 2018.
	\bibitem{ref197}	M. Hosny, M. M. Darwish, and T. Aboelenen, “Novel fractional-order generic Jacobi-Fourier moments for image analysis,” \emph{Signal Process.}, vol. 172, Jul. 2020.
	\bibitem{ref198}	M. Hosny, M. M. Darwish, and T. Aboelenen, “Novel fractional-order polar harmonic transforms for gray-scale and color image analysis,” \emph{J. Frankl. Inst.-Eng. Appl. Math.}, vol. 357, no. 4, pp. 2533-2560, Mar. 2020.
	\bibitem{ref199}	M. Hosny, M. M. Darwish, and M. M. Eltoukhy, “Novel multi-channel fractional-order radial harmonic Fourier moments for color image analysis,” \emph{IEEE Access}, vol. 8, pp. 40732-40743, Feb. 2020.
	\bibitem{ref200}	M. Hosny, M. Abd Elaziz, and M. M. Darwish, “Color face recognition using novel fractional-order multi-channel exponent moments,” \emph{Neural Comput. Appl.}. Nov. 2020.
	\bibitem{ref201}	Y. Qu, C. Cui, S. Chen, and J. Li, “A fast subpixel edge detection method using Sobel-Zernike moments operator,” \emph{Image Vis. Comput.}, vol. 23, no. 1, pp. 11-17, Jan. 2005.
	\bibitem{ref202}	A. Iscen, G. Tolias, P. H. Gosselin, and H. Jegou, “A comparison of dense region detectors for image search and fine-grained classification,” \emph{IEEE Trans. Image Process.}, vol. 24, no. 8, pp. 2369-2381, Aug. 2015.
	\bibitem{ref203}	Z. Chen and S. K. Sun, “A Zernike moment phase-based descriptor for local image representation and matching,” \emph{IEEE Trans. Image Process.}, vol. 19, no. 1, pp. 205-219, Jan. 2010.
	\bibitem{ref204}	M. Schlemmer, M. Heringer, F. Morr, I. Hotz, M. H. Bertram, C. Garth, W. Kollmann, B. Hamann, and H. Hagen, “Moment invariants for the analysis of 2D flow fields,” \emph{IEEE Trans. Vis. Comput. Graph.}, vol. 13, no. 6, pp. 1743-1750, Nov. 2007.
	\bibitem{ref205}	A. Kumar, M. O. Ahmad, and M. N. S. Swamy, “Tchebichef and adaptive steerable-based total variation model for image denoising,” \emph{IEEE Trans. Image Process.}, vol. 28, no. 6, pp. 2921-2935, Jun. 2019.
	\bibitem{ref206}	A. Kumar, “Deblurring of motion blurred images using histogram of oriented gradients and geometric moments,” \emph{Signal Process.-Image Commun.}, vol. 55, pp. 55-65, Jul. 2017.
	\bibitem{ref207}	X. Gao, Q. Wang, X. Li, D. Tao, and K. Zhang, “Zernike-moment-based image super resolution,” \emph{IEEE Trans. Image Process.}, vol. 20, no. 10, pp. 2738-2747, Oct. 2011.
	\bibitem{ref208}	Z. Yang and F. S. Cohen, “Cross-weighted moments and affine invariants for image registration and matching,” \emph{IEEE Trans. Pattern Anal. Mach. Intell.}, vol. 21, no. 8, pp. 804-814, Aug. 1999.
	\bibitem{ref209}	Xiao, G. Lu, Y. Zhang, W. Li, and G. Wang, “Lossless image compression based on integer discrete Tchebichef transform,” \emph{Neurocomputing}, vol. 214, pp. 587-593, Nov. 2016.
	\bibitem{ref210}	L. Li, W. Lin, X. Wang, G. Yang, K. Bahrami, and A. C. Kot, “No-reference image blur assessment based on discrete orthogonal moments,” \emph{IEEE T. Cybern.}, vol. 46, no. 1, pp. 39-50, Jan. 2016.
	\bibitem{ref211}	M. Liang, J. Du, L. Li, Z. Xue, X. Wang, F. Kou, and X. Wang, “Video super-resolution reconstruction based on deep learning and spatio-temporal feature self-similarity,” \emph{IEEE Trans. Knowl. Data Eng.}, Oct. 2020.
	\bibitem{ref212}	M. Bronstein, M. M. Bronstein, L. J. Guibas, and M. Ovsjanikov, “Shape Google: Geometric words and expressions for invariant shape retrieval,” \emph{ACM Trans. Graph.}, vol. 30, no. 1, Jan. 2011.
	\bibitem{ref213}	F. Murtagh and J. L. Starck, “Wavelet and curvelet moments for image classification: Application to aggregate mixture grading,” \emph{Pattern Recognit. Lett.}, vol. 29, no. 10, pp. 1557-1564, Jul. 2008.
	\bibitem{ref214}	G. Paschos, I. Radev, and N. Prabakar, “Image content-based retrieval using chromaticity moments,” \emph{IEEE Trans. Knowl. Data Eng.}, vol. 15, no. 5, pp. 1069-1072, Sep. 2003.
	\bibitem{ref215}	L. Zhao and L. S. Davis, "Closely coupled object detection and segmentation," in \emph{Proc. IEEE Int. Conf. Comput. Vis.}, 2005, pp. 454-461. 
	\bibitem{ref216}	X. Wang, Z. Wu, L. Chen, H. Zheng, and H. Yang, “Pixel classification based color image segmentation using quaternion exponent moments,” \emph{Neural Netw.}, vol. 74, pp. 1-13, Feb. 2016.
	\bibitem{ref217}	D. Ioannidis, D. Tzovaras, I. G. Damousis, S. Argyropoulos, and K. Moustakas, “Gait recognition using compact feature extraction transforms and depth information,” \emph{IEEE Trans. Inf. Forensic Secur.}, vol. 2, no. 3, pp. 623-630, Sep. 2007.
	\bibitem{ref218}	N. Vincent and J. M. Ogier, “Shall deep learning be the mandatory future of document analysis problems?,” \emph{Pattern Recognit.}, vol. 86, pp. 281-289, Feb. 2019.
	\bibitem{ref219}	M. Liu and P. T. Yap, “Invariant representation of orientation fields for fingerprint indexing,” \emph{Pattern Recognit.}, vol. 45, no. 7, pp. 2532-2542, Jul. 2012.
	\bibitem{ref220}	S. M. Lajevardi, and Z. M. Hussain, “Higher order orthogonal moments for invariant facial expression recognition,” \emph{Digit. Signal Prog.}, vol. 20, no. 6, pp. 1771-1779, Dec. 2010.
	\bibitem{ref221}	D. Tsougenis, G. A. Papakostas, D. E. Koulouriotis, and V. D. Tourassis, “Performance evaluation of moment-based watermarking methods: A review,” \emph{J. Syst. Softw.}, vol. 85, no. 8, pp. 1864-1884, Aug. 2012.
	\bibitem{ref222}	Y. Zhang, X. Luo, Y. Guo, C. Qin, and F. Liu, “Zernike moment-based spatial image steganography resisting scaling attack and statistic detection,” \emph{IEEE Access}, vol. 7, pp. 24282-24289, Feb. 2019.
	\bibitem{ref223}	Y. Zhao, S. Wang, X. Zhang, and H. Yao, “Robust hashing for image authentication using Zernike moments and local features,” \emph{IEEE Trans. Inf. Forensic Secur.}, vol. 8, no. 1, pp. 55-63, Jan. 2013.
	\bibitem{ref224}	D. Cozzolino, G. Poggi, and L. Verdoliva, “Efficient dense-field copy-move forgery detection,” \emph{IEEE Trans. Inf. Forensic Secur.}, vol. 10, no. 11, pp. 2284-2297, Nov. 2015.
	\bibitem{ref225}	X. Dai, H. Shu, L. Luo, G. Han, and J. L. Coatrieux, “Reconstruction of tomographic images from limited range projections using discrete Radon transform and Tchebichef moments,” \emph{Pattern Recognit.}, vol. 43, no. 3, pp. 1152-1164, Mar. 2010.
	\bibitem{ref226}	X. Dai and S. Khorram, “A feature-based image registration algorithm using improved chain-code representation combined with invariant moments,” \emph{IEEE Trans. Geosci. Remote Sensing}, vol. 37, no. 5, pp. 2351-2362, Sep. 1999.
	\bibitem{ref227}	F. Chaumette, “Image moments: A general and useful set of features for visual servoing,” \emph{IEEE Trans. Robot. Autom.}, vol. 20, no. 4, pp. 713-723, Aug. 2004.
	\bibitem{ref228}	D. Casasent and D. Psaltis, “New optical transforms for pattern recognition,” \emph{Proc. IEEE}, vol. 65, no. 1, pp. 77-84, Jan. 1977.
	\bibitem{ref229}	L. Zhai, B. Li, J. Chen, X. Wang, M. Xu, J. Liu, and S. Lu, “Chemical image moments and their applications,” \emph{Trac-Trends Anal. Chem.}, vol. 103, pp. 119-125, Jun. 2018.
	\bibitem{ref230}	A. Sit, W. H. Shin, and D. Kihara, “Three-dimensional Krawtchouk descriptors for protein local surface shape comparison,” \emph{Pattern Recognit.}, vol. 93, pp. 534-545, Sep. 2019.
	\bibitem{ref231}	M. Uhrin. (2021) “Through the eyes of a descriptor: Constructing complete, invertible, descriptions of atomic environments.” [Online]. Available: https://arxiv.org/abs/2104.09319
	\bibitem{ref232}	Y. Xin, S. Liao, and M. Pawlak, “Circularly orthogonal moments for geometrically robust image watermarking,” \emph{Pattern Recognit.}, vol. 40, no. 12, pp. 3740-3752, Dec. 2007.
	\bibitem{ref233}	L. Li, S. Li, A. Abraham, and J. S. Pan, “Geometrically invariant image watermarking using polar harmonic transforms,” \emph{Inf. Sci.}, vol. 199, pp. 1-19, Sep. 2012.
	\bibitem{ref234}	C. Wang, X. Wang, Z. Xia, B. Ma, and Y. Q. Shi, “Image description with polar harmonic Fourier moments,” \emph{IEEE Trans. Circuits Syst. Video Technol.}, vol. 30, no. 12, pp. 4440-4452, Dec. 2020.
	\bibitem{ref235}	Z. Wang, A. C. Bovik, H. R. Sheikh, and E. P. Simoncelli, “Image quality assessment: From error visibility to structural similarity,” \emph{IEEE Trans. Image Process.}, vol. 13, no. 4, pp. 600-612, Apr. 2004.
	\bibitem{ref236}	Z. Wang, J. Li, and G. Wiederhold, “SIMPLIcity: Semantics-sensitive integrated matching for picture libraries,” \emph{IEEE Trans. Pattern Anal. Mach. Intell.}, vol. 23, no. 9, pp. 947-963, Sep. 2001.
	\bibitem{ref237}	J. Lu, V. E. Liong, X. Zhou, and J. Zhou, “Learning compact binary face descriptor for face recognition,” \emph{IEEE Trans. Pattern Anal. Mach. Intell.}, vol. 37, no. 10, pp. 2041-2056, Oct. 2015.
	\bibitem{ref238}	C. Szegedy, W. Liu, Y. Jia, P. Sermanet, S. Reed, D. Anguelov, D. Erhan, and A. Rabinovich, “Going deeper with convolutions,” in \emph{Proc. IEEE Conf. Comput. Vis. Pattern Recognit.}, Jun. 2015, pp. 1–9.
	\bibitem{ref239}	K. He, X. Zhang, S. Ren and J. Sun, "Deep residual learning for image recognition", in \emph{Proc. IEEE Conf. Comput. Vis. Pattern Recognit.}, pp. 770-778, Jun. 2016.
	\bibitem{ref240}	X. Zhang, C. Liu, and C. Suen, “Towards robust pattern recognition: a review,” \emph{Proc. IEEE}, vol. 108, no. 6, pp. 894-922, Jun. 2020.
	\bibitem{ref241}	Y. Wang, Y. Y. Tang, L. Li, H. Chen, and J. Pan, “Atomic representation-based classification: theory, algorithm, and applications,” \emph{IEEE Trans. Pattern Anal. Mach. Intell.}, vol. 41, no. 1, pp. 6-19, Jan. 2019.
	\bibitem{ref242}	Y. Wang, M. Shi, S. You, and C. Xu, “DCT inspired feature transform for image retrieval and reconstruction,” \emph{IEEE Trans. Image Process.}, vol. 25, no. 9, pp. 4406-4420, Sep. 2016.
	\bibitem{ref243}	T. Zhao and T. Blu, “The Fourier-Argand representation: an optimal basis of steerable patterns,” \emph{IEEE Trans. Image Process.}, vol. 29, pp. 6357-6371, May 2020.
	\bibitem{ref244}	J. Luis Silvan-Cardenas, and A. Salazar-Garibay, “Local geometric deformations in the DHT domain with applications,” \emph{IEEE Trans. Image Process.}, vol. 28, no. 4, pp. 1980-1992, Apr. 2019.
	\bibitem{ref245}	A. Sit and D. Kihara, “Comparison of image patches using local moment invariants,” \emph{IEEE Trans. Image Process.}, vol. 23, no. 5, pp. 2369-2379, May 2014.
	\bibitem{ref246}	W. Tan and A. Kumar, “Accurate iris recognition at a distance using stabilized iris encoding and Zernike moments phase features,” \emph{IEEE Trans. Image Process.}, vol. 23, no. 9, pp. 3962-3974, Sep. 2014.
	\bibitem{ref247}	A. Kar, S. Pramanik, A. Chakraborty, D. Bhattacharjee, E. S. L. Ho, and H. P. H. Shum, “LMZMPM: Local modified Zernike moment per-unit mass for robust human face recognition,” \emph{IEEE Trans. Inf. Forensic Secur.}, vol. 16, pp. 495-509, Aug. 2021.
	\bibitem{ref248}	Y. Hao, Q. Li, H. Mo, H. Zhang, and H. Li, “AMI-Net: Convolution neural networks with affine moment invariants,” \emph{IEEE Signal Process. Lett.}, vol. 25, no. 7, pp. 1064-1068, Jul. 2018.
	\bibitem{ref249}	J. Wu, S. Qiu, Y. Kong, Y. Chen, L. Senhadji, and H. Shu, “MomentsNet: A simple learning-free method for binary image recognition,” in \emph{Proc. Int. Conf. Image Process.}, 2017, pp. 2667-2671.
	\bibitem{ref250}	Y. Duan, J. Lu, J. Feng, and J. Zhou, “Learning rotation-invariant local binary descriptor,” \emph{IEEE Trans. Image Process.}, vol. 26, no. 8, pp. 3636-3651, Aug. 2017.
	\bibitem{ref251}	L. Xie, J. Wang, W. Lin, B. Zhang, and Q. Tian, “Towards reversal-invariant image representation,” \emph{Int. J. Comput. Vis.}, vol. 123, no. 2, pp. 226-250, Jun. 2017.
	\bibitem{ref252}	L. Sifre and S. Mallat, “Rotation, scaling and deformation invariant scattering for texture discrimination,” in \emph{Proc. IEEE Conf. Comput. Vis. Pattern Recognit.}, 2013, pp. 1233-1240.
	\bibitem{ref253}	M. Jaderberg, K. Simonyan, A. Zisserman, and K. Kavukcuoglu. (2015) “Spatial transformer networks.” [Online]. Available: https://arxiv.org/abs/1506.02025
	\bibitem{ref254}	Y. Wang, C. Xu, C. Xu, and D. Tao, “Packing convolutional neural networks in the frequency domain,” \emph{IEEE Trans. Pattern Anal. Mach. Intell.}, vol. 41, no. 10, pp. 2495-2510, Oct. 2019.
	\bibitem{ref255}	A. Agarwal, R. Singh, M. Vatsa, and N. K. Ratha, “Image transformation based defense against adversarial perturbation on deep learning models,” \emph{IEEE Trans. Dependable Secur. Comput.}. 2020.
	\bibitem{ref256}	T. Yang, J. Ma, Y. Miao, X. Liu, X. Wang, and Q. Meng, “PLCOM: Privacy-preserving outsourcing computation of Legendre circularly orthogonal moment over encrypted image data,” \emph{Inf. Sci.}, vol. 505, pp. 198-214, Dec. 2019.
	
\end{thebibliography}

\end{document}